\documentclass[lettersize,journal]{IEEEtran}

\usepackage{graphicx}  
\usepackage{float}  
\usepackage{amsmath,amsfonts}
\usepackage{array}
\usepackage{subfig}
\usepackage{textcomp}
\usepackage{stfloats}
\usepackage{url}
\usepackage{verbatim}
\usepackage{graphicx}
\usepackage{cite}
\usepackage{xspace}
\usepackage{algorithmic}
\usepackage{algorithm}
\usepackage{caption}
\usepackage{enumerate}
\usepackage{enumitem}
\usepackage{hyperref}
\usepackage{multirow, makecell}
\usepackage{nccmath}
\usepackage{ragged2e}
\usepackage{threeparttable} 
\usepackage{tabularx}
\usepackage{xspace}
\usepackage{color}
\usepackage{multirow}
\usepackage{xcolor}
\usepackage{tabularx}
\usepackage{cellspace}
\usepackage{makecell}
\definecolor{xxrdarkgreen}{RGB}{0,100,0} 

\newcommand{\ie}{\emph{i.e.},\xspace}

\newcommand{\eg}{\emph{e.g.},\xspace}

\newcommand{\etal}{\emph{et al.}\xspace}

\newcommand\figref[1]{Fig.~\ref{#1}}

\newcommand\tabref[1]{Tab.~\ref{#1}}

\newcommand\secref[1]{Sec.~\ref{#1}}

\ifodd 1

\else

\fi
\begin{document}

\title{Adaptive and Resource-efficient Agentic AI Systems for Mobile and Embedded Devices: A Survey}

\author{
Sicong Liu,~\IEEEmembership{Member,~IEEE,}
Weiye Wu, 
Xiangrui Xu,
Teng Li, 
Bowen Pang,
Bin Guo,~\IEEEmembership{Senior Member,~IEEE,}
Zhiwen Yu,~\IEEEmembership{Senior Member,~IEEE}
}
\markboth{Journal of \LaTeX\ Class Files,~Vol.~00, No.~0, April~2023}{}

\maketitle

\begin{abstract}
Large foundation models (FMs) such as LLMs, VLMs, diffusion models, and MLLMs have shifted AI from fragmented, task-specific models toward versatile cognitive systems. 
In parallel, the AI agents has been refreshed by FMs as their cognitive core, enabling autonomy, perception, planning, and self-reflection in dynamic environments. 
Together, these shifts open opportunities for agentic AI on mobile and edge platforms, where real-world applications demand low-latency, energy-efficient, and adaptive intelligence. 
However, current surveys mainly focus on static FM optimization or generic agents, overlooking mobile-specific challenges of resource constraints, runtime adaptability, and diverse conditions. 
This article fills that gap by providing the first systematic survey on adaptive and resource-efficient agentic AI systems on mobile/edge devices. 
We propose a novel taxonomy covering elastic FM inference, test-time adaptation, dynamic multimodal integration, and application-driven optimization, and we outline open issues and evaluation methodologies to inspire future research at the intersection of FMs, agents, and mobile/edge intelligence.
We believe this survey can help readers to understand the connections between enabling technologies while promoting further discussions.
\end{abstract}

\begin{IEEEkeywords}
Adaptive and resource-efficient, Agentic AI, Elastic FM Inference, Test-time FM adaptation
\end{IEEEkeywords}

\section{Introduction}
Large foundation models (FMs), including large language models (LLMs) such as GPTs and LLaMA, vision–language models (VLMs) such as CLIP and BLIP-2, diffusion models, and multimodal large models (MLLMs) such as GPT-4o and Gemini, have demonstrated remarkable general-purpose machine learning capabilities.
This marks a fundamental \textit{shift}, \ie fragmented, task-specific machine learning models (\eg CNNs~\cite{krizhevsky2017imagenet}, Transformers~\cite{2020vit}) are converging toward versatile FMs with high-level cognition, capable of scalability, multimodal reasoning, and contextual adaptation.

Meanwhile, the concept of the AI agent was first formalized in the 1990s as autonomous systems defined by the sensing–action loop~\cite{russell1995modern}.
Early agents, however, built on symbolic AI or \textit{small-scale} machine learning, lacked the reasoning power and adaptability required for long-term autonomy.
With the advent of FMs as the \textit{cognitive core}, a second \textit{shift} has emerged, \ie AI agents can surpass traditional rule-based behaviors and achieve greater autonomy and generalization, thereby enabling perception, planning, action, and self-reflection in dynamic environments.

This dual shift is further accelerated by the growing demands of real-world mobile applications, such as autonomous driving (\eg Google Waymo~\cite{teoh2017rage}), robotics (\eg Meta FAIR~\cite{hu2023toward}), and mobile task automation (\eg Apple Intelligence~\cite{gunter2024apple}).
Such applications require long-term complex operation, real-time or low-latency interaction, and robust adaptation to diverse and dynamic environments. 
For example, self-driving cars must process multimodal sensor streams (\eg camera, LiDAR) in vehicle systems for navigation and obstacle avoidance, while mobile task automation must respond immediately to evolving user contexts.
Moreover, growing privacy concerns further highlight the importance of on-device and edge execution, where sensitive data can be processed locally without reliance on the cloud~\cite{delgado2022survey}.

\begin{figure}[t]
    \centering
    \includegraphics[width=0.4\textwidth]{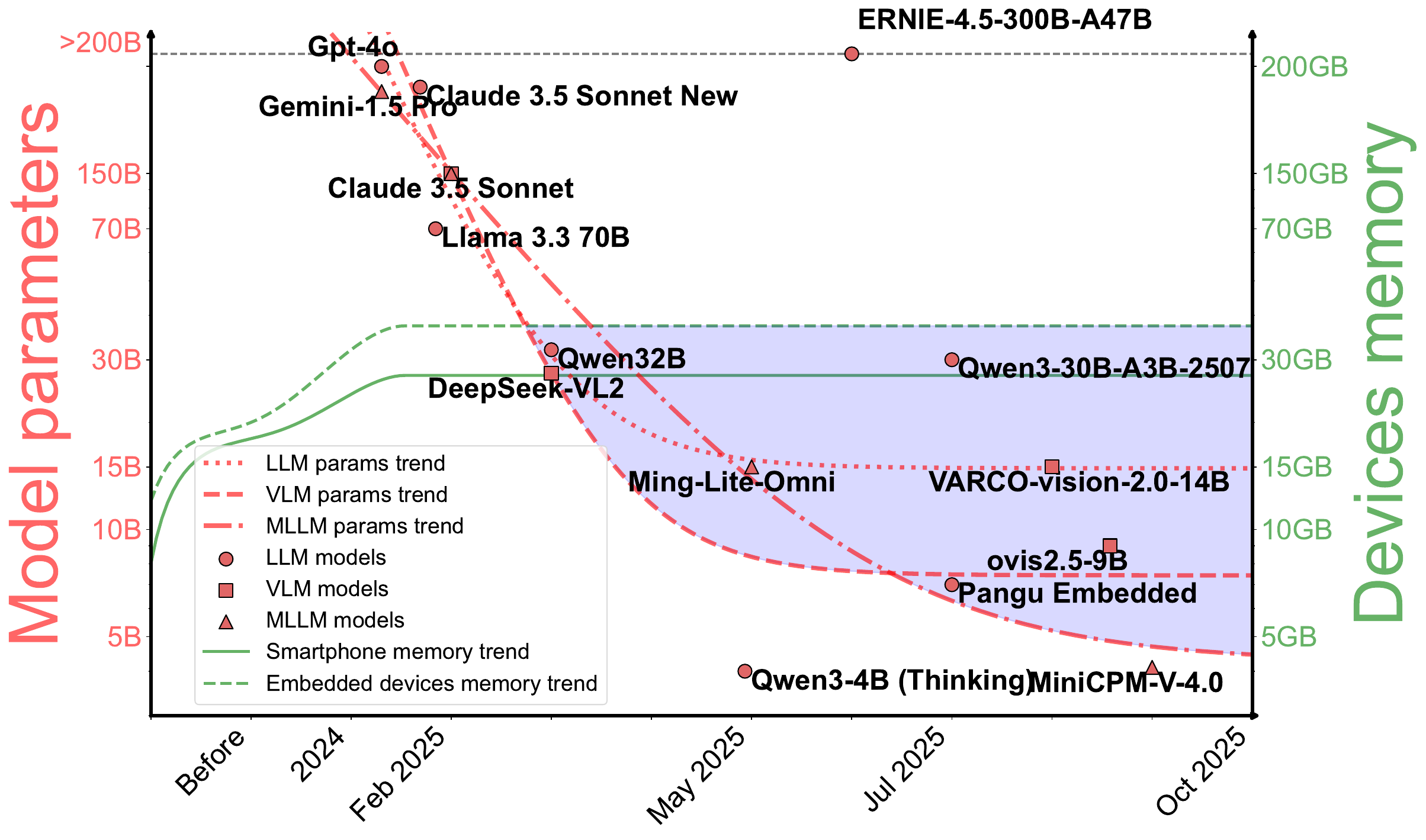}
    \caption{Development of FMs (8-bit) and embedded hardware.}
    \label{fig:model_performance_trends}
    \vspace{-6mm}
\end{figure}

In parallel, as shown in \figref{fig:model_performance_trends}, the evolution of FMs exhibits a trend reminiscent of Moore’s Law.
The parameter scale required for large FM-level performance is rapidly shrinking (red lines), while the computational capacity of mobile and embedded devices continues to increase (green lines).
The convergence of these trends indicates that deployable FMs on mobile and embedded platforms are becoming feasible, unlocking new opportunities.
At the same time, advances in the mobile/edge ecosystem, including commercial SoCs (\eg GPUs, DSPs, NPUs)~\cite{dhilleswararao2022efficient}, edge intelligence frameworks~\cite{mach2017mobile}, lightweight deployment libraries~\cite{zhou2022edgeai}, and efficient communication protocols for multi-agent collaboration—are fostering scalable and cost-effective deployment of \textit{agentic AI systems}.

Despite these advances, current FMs remain far from practical deployment in real-world mobile scenarios.
On mobile/edge-assisted platforms, \textit{adaptivity} and \textit{resource efficiency} are not optional add-ons but fundamental bottlenecks.
It is necessary to:
\textit{i)} restructure FMs and reasoning chains on the fly to cope with dynamic sensor data and fluctuating hardware resources;
\textit{ii)} sustain low-latency or real-time responses within stringent resource budgets; and
\textit{iii)} enable self-evolving cognitive abilities to handle non-stationary or even unseen data and tasks.

Existing surveys have addressed related aspects but remain limited.
Most works~\cite{2024mengweisurvey, zhou2024survey} focus on \textit{static} FM optimization techniques (\eg distillation, quantization, pruning, fine-tuning), overlooking the need for \textit{agentic FMs} that dynamically reconfigure structures, reasoning chains, and routing paths in response to runtime resource and environmental conditions.
Although some studies explore dynamic or self-adaptive FM algorithms~\cite{2022switchtransformers,du2022glammoe}, they often rely on manually designed heuristics, leaving this direction in its infancy.
Meanwhile, surveys on AI agents~\cite{ye2025mobile} have spanned generative~\cite{harrer2024generative, yee2024agents}, logic-based~\cite{aladin2019logic, calegari2021logic}, multimodal~\cite{ye2025mobile, xie2024large}, and embodied agents~\cite{duan2022survey, wu2024embodied}, but rarely address the mobile-specific constraints of limited resources, responsiveness, and real-world adaptability.

This survey aims to fill these gaps by providing a systematic study of \textit{adaptive and resource-efficient agentic AI systems} for mobile and embedded platforms.
We map enabling techniques into a unified landscape of resource efficiency and adaptivity, offering both a comprehensive taxonomy and insights into emerging challenges as follows:

\noindent$\bullet$ \textbf{Elastic FM Inference} (\secref{sec:elastic_fm_inference_in_agentic_ai_systems}). On mobile and embedded devices, fluctuating resource availability requires FMs to adjust their structures, often causing weight–structure \textit{mismatches}.
While retraining can mitigate these issues, it is computationally prohibitive and unsuitable for real-time applications.
Also, re-compression introduces operator dependencies that complicate resource mapping, and hardware heterogeneity further hinders efficient allocation.
Prior advances, including \textit{dynamic prompts, selective reasoning, scalable depth/width, routing, and KV cache optimization}, provide initial pathways toward elastic and resource-efficient FM inference.

\noindent$\bullet$ \textbf{Test-time Adaptation of FM} (\secref{sec:adaptation}).  
Lightweight FMs often exhibit poor generalization, resulting in cognitive (weight) \textit{mismatches} during long-term use.
Although retraining can adapt models to data shifts and unseen tasks, resource constraints on mobile devices (\eg drones, robots) make it impractical.
Moreover, the heterogeneity of distributed agent platforms reduces adaptation efficiency.
We review techniques that enable online adaptation without full retraining, including \textit{test-time prompt learning, parameter-efficient fine-tuning (PEFT), memory augmentation, interactive learning, and system-level methods} (\eg scheduling, distributed updates), with extensions to multi-agent collaboration.

\noindent$\bullet$ \textbf{Dynamic Multi-modal FMs} (\secref{sec_MLLM}).
The integration of \textit{heterogeneous} and \textit{asynchronous} sensor streams (\eg vision, speech, LiDAR, RF) introduces both computational overhead and alignment challenges.
Waiting for slow modalities inflates latency, whereas discarding them reduces accuracy, creating an inherent accuracy–latency–bandwidth trade-off.
We summarize adaptive fusion strategies such as \textit{dynamic attention, routing, alignment, and token compression}, which improve efficiency and scalability under mobile/edge constraints.

\noindent$\bullet$ \textbf{Agentic AI Applications} (\secref{Agentic AI Applications}).
Real-world applications (\eg embodied agents, GUI assistants, generative agents, and personal assistive services) demand long-term operation, responsiveness, and interactive adaptability.
Static FM deployments struggle to satisfy these requirements due to rigid pipelines and limited resource awareness.
We demonstrate how \textit{elastic inference} and \textit{adaptive retraining} can integrate FM capabilities into context-sensitive, resource-aware applications, emphasizing the role of \textit{application-driven optimization} in bridging enabling techniques with practical deployment.

In summary, our contributions are threefold:
\begin{itemize}
    \item To the best of our knowledge, this is the first systematic survey of adaptive and resource-efficient agentic AI systems on mobile \& embedded devices, outlining pathways for re-mapping FM structures, cognition, and hardware resources within a hardware–software spectrum.
    \item We propose a novel taxonomy of enabling techniques, spanning elastic inference, test-time adaptation, and dynamic multimodal integration, clarifying trade-offs in accuracy, latency, communication, and energy efficiency.
    \item We identify open issues in adaptive and resource-efficient agentic AI systems and outline potential research directions to guide innovation in FM architecture, algorithm–system co-design, and mobile/edge deployment.
\end{itemize} 

The rest of this paper is organized as follows.
\secref{sec:fundamentals} reviews the fundamentals;
\secref{sec:elastic_fm_inference_in_agentic_ai_systems} and \secref{sec:adaptation} survey elastic inference and test-time adaptation;
\secref{sec_MLLM} discusses dynamic multimodal FMs;
\secref{Agentic AI Applications} highlights representative applications.
Open challenges and evaluation methodologies are presented in \secref{sec_open issue}, and conclusions are drawn in \secref{sec_conclusuion}.

\section{Fundamentals and Overview}
\label{sec:fundamentals}
This section provides an overview of adaptive and resource-efficient agentic AI systems on mobile and embedded devices, and clarifies their relation to existing concepts.
\begin{figure}[t]
    \centering
    \includegraphics[width=0.38\textwidth]{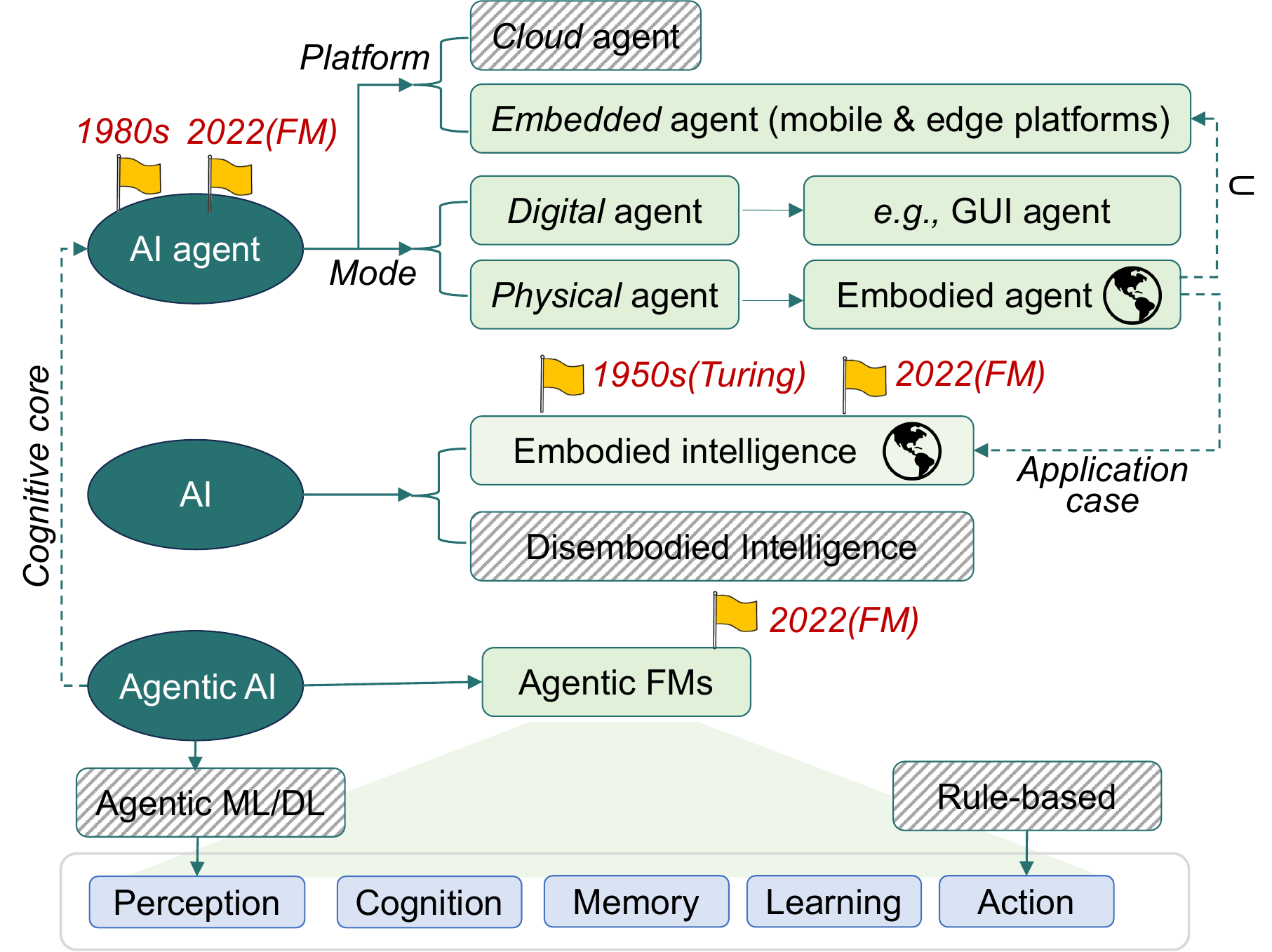}
    \caption{Related concepts.}
    \label{fig:related_concept}
    \vspace{-6mm}
\end{figure}

\subsection{Related Concepts}
Agentic AI systems build upon, and intersect with, several related notions (\figref{fig:related_concept}).

\textit{Mobile and embedded devices} are ubiquitous hardware platforms (\eg CPUs, GPUs, DSPs, NPUs, MCUs) that operate under strict compute, memory, and energy constraints (\eg smartphones, UAVs, in-vehicle units). 
They provide the execution platform.  

\textit{AI agents}, formalized in the 1990s~\cite{russell1995modern}, follow the \textit{sensing–decision–action loop} and may exist in \textit{digital} (\eg GUI assistants) or \textit{physical} (\ie embodied agents) forms, and can be deployed across \textit{cloud}, \textit{edge}, and \textit{mobile} environments.   
This survey focuses on \textit{edge and mobile settings}, where resource efficiency is a critical bottleneck.

\textit{Embedded AI agents} run on resource-constrained mobile/edge platforms, requiring \textit{resource-efficient} techniques for sustained operation.
\textit{Embodied AI agents} are equipped with sensors and actuators for direct physical environmental interaction (\eg drones, autonomous vehicles). 
Conceptually, embodied agents are a subset of embedded agents (\ie embodied $\subset$ embedded). 

\textit{Embodied intelligence}, envisioned by Turing in the 1950s~\cite{turing2007computing}, emphasizes cognition through physical perception and interaction, today instantiated by embodied agents integrated with FMs.  

\textit{Agentic AI} represents a paradigm shift since 2022 because FMs (\eg LLMs, VLMs, MLLMs) extend agents beyond \textit{perception} to \textit{high-level cognition}, including reasoning, task decomposition, and decision-making.
When augmented with planning, memory, tool use, and reflection, FMs function as the \textit{cognitive core of AI agents}, enabling autonomy and generalization.
An \textit{agentic AI system} typically comprises one or multiple FM-powered agents, coordinated through scheduling and communication, to interact with heterogeneous systems, adapt to dynamic environments, and collaborate with humans. This constitutes the scope of this survey. 

\subsection{Our Scope}
\begin{figure}[t]
    \centering
    \includegraphics[width=0.49\textwidth]{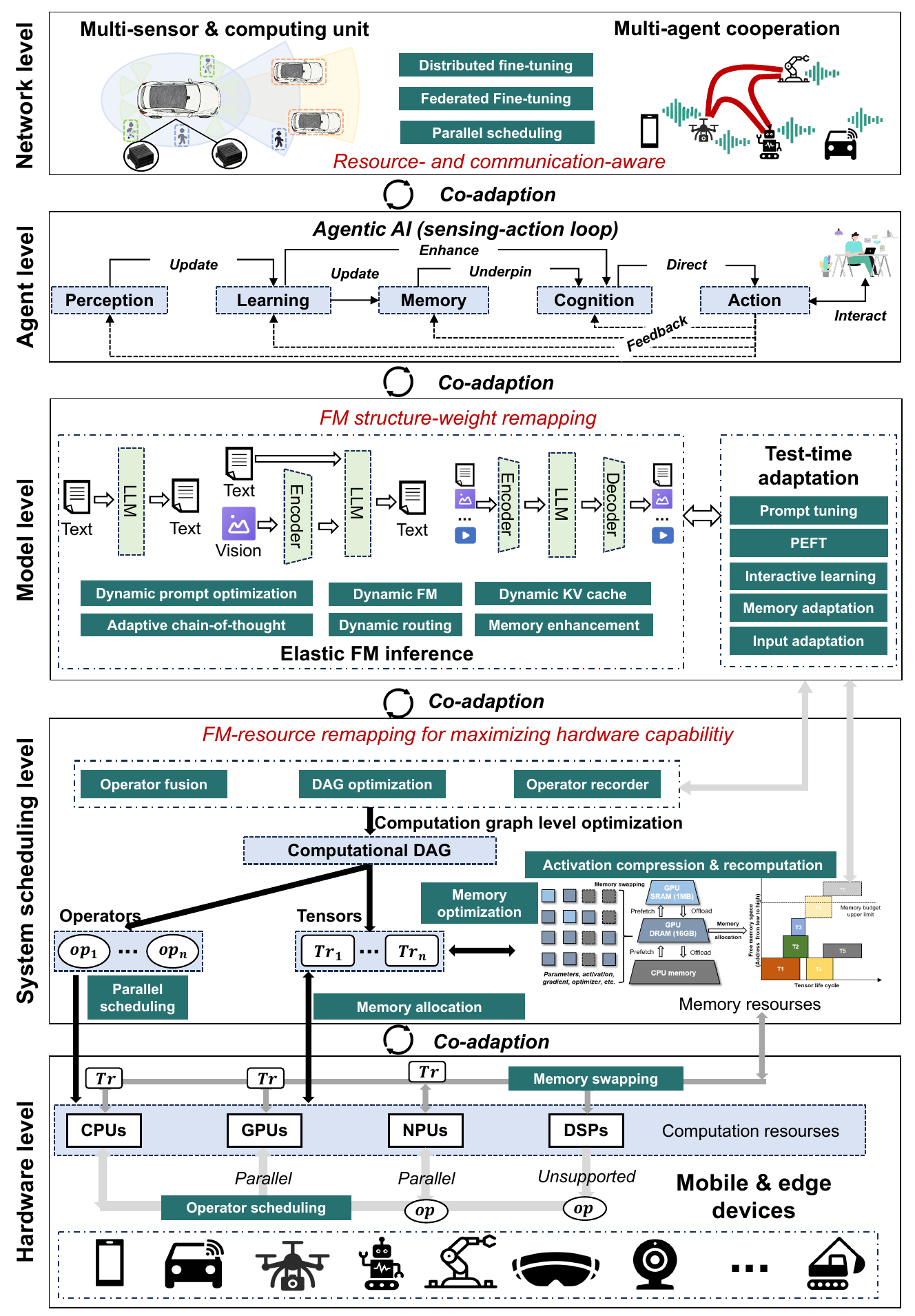}
    \caption{Cross-level landscape of agentic AI systems.}
    \label{fig:cross_level}
    \vspace{-6mm}
\end{figure}

This survey focuses on \textit{adaptive, resource-efficient agentic AI systems} for \textit{mobile and embedded platforms}, conceptualized across five interrelated levels (\figref{fig:cross_level}).
We emphasize mobile and embedded settings because the tension between large-scale FMs and constrained resources is the primary bottleneck, further compounded by rising privacy concerns that motivate on-device or edge execution rather than cloud reliance.

\begin{itemize}
\item \textit{Hardware level}: resource-constrained mobile/embedded devices or edge servers serve as the physical platform.
\item \textit{System scheduling level}: FMs are compiled into DAGs and dynamically mapped across heterogeneous compute/memory resources via \textit{frontend–backend co-compilation}, maximizing hardware utilization.
\item \textit{Model level}: FMs enhanced with planning, memory, and reasoning (\textit{agentic FMs}) act as the cognitive “brain” for perception and decision-making.
\item \textit{Agent level}: AI agents emerge from the integration of platforms, FMs, and sensing–action loops. On embedded devices, they form \textit{embedded agents}; equipped with physical sensors and actuators, they become \textit{embodied agents}.
\item \textit{Network level}: agents coordinate through scheduling and communication to form distributed \textit{agentic AI systems}.
\end{itemize}
In this survey, we exclude resource-rich cloud deployments, as they face fewer efficiency constraints. 
Nevertheless, the techniques discussed here remain broadly relevant across the cloud–edge–mobile continuum.

\subsection{Why Adaptivity and Resource-Efficiency Matter}  
Unlike resource-rich clouds, these systems must remain responsive under real-world \textit{diversity} and \textit{dynamics}.   
Challenges arise from four-fold: 
\textit{i)} fluctuating hardware resources due to contention, throttling, and heterogeneity;  
\textit{ii)} unstable mobile networks with variable latency and bandwidth;  
\textit{iii)} dynamic inputs and tasks (\eg multimodal bursts, environmental shifts, heterogeneous user demands) requiring flexible FM structures and operator flows; and  
\textit{iv)} long-running, open-world operations where agents collaborate and process distributed multimodal streams.  

In such contexts, \textit{static} FM structure–weight–resource mappings degrade quickly. 
Full retraining is both costly and too slow, while structural adjustments introduce new operator dependencies and resource mismatches exacerbated by hardware heterogeneity.  
What is needed are \textit{elastic reconfiguration} mechanisms to realign FM structures with resources, and \textit{efficient adaptation} techniques to update parameters under nonstationary tasks and data.   

\subsection{Characteristics of FM-based Agentic AI Systems} 
By leveraging FM backbones, agentic AI systems inherit distinctive traits that enhance \textit{adaptivity} and \textit{resource efficiency}, surpassing traditional DL-based mobile systems~\cite{liu2023enabling, liu2024lightweight} that are constrained to narrow, pre-defined tasks.  
FMs, pre-trained on large-scale multimodal data and unified transformer architectures (\eg LLaMA-2~\cite{touvron2023llama}, ViT-G~\cite{2020vit}), exhibit emergent reasoning even at moderate scales~\cite{wei2022emergent, schaeffer2023emergent}.
These properties empower agents to generalize via prompting~\cite{brown2020language}, decompose complex goals~\cite{wei2022chain}, incorporate contextual memory~\cite{park2023generative}, and refine behavior through human or environmental feedback~\cite{udagawa2022human, 2024simuser}.  
Concretely, FM-powered agentic AI systems exhibit four key capabilities that support \textit{complex} tasks:  
\begin{itemize}
    \item \textit{Cross-task generalization}: Pre-trained FMs (\eg LLMs, VLMs, MLLMs) handle diverse modalities with minimal retraining cost, offering robust adaptability.  
    \item \textit{Goal decomposition}: High-level instructions (\eg \textit{Plan a trip}) are decomposed into sub-goals (\eg booking, transportation) via chain-of-thought reasoning.  
    \item \textit{Contextual memory}: Lightweight episodic memory structures~\cite{pritzel2017neural} capture and reuse preferences and past interactions, enabling personalization and continuity.  
    \item \textit{Feedback integration}: Reinforcement learning from human feedback (RLHF) combined with environmental signals (\eg battery, network state) refines behavior online, balancing accuracy, latency, and energy efficiency.  
\end{itemize}  
These underscore adaptivity and efficiency for complex tasks, but also pose challenges. FMs are resource-intensive, planning needs elastic inference, memory entails efficiency–accuracy trade-offs, and feedback requires online adaptation.

\subsection{Taxonomy of Enabling Techniques}
We summarize four categories of enabling techniques for adaptive and resource-efficient agentic AI systems:  
\begin{itemize}
\item \textit{Elastic FM Inference} (\secref{sec:elastic_fm_inference_in_agentic_ai_systems}).  
Elastic inference dynamically adapts FM \textit{architectures}, reasoning \textit{depth}, \textit{computation}, and \textit{communication} at runtime to meet accuracy–latency–energy goals under fluctuating device and network constraints.  
Unlike static pipelines, it treats execution as continuously tunable across memory, compute, and bandwidth.  
Techniques include \textit{dynamic prompts}~\cite{2023DYNAICL,2023recomp} to cut input redundancy, \textit{adaptive CoT}~\cite{2024agentinstruct,2023Skeleton-of-thought,2021Scratchpad} for selective reasoning, \textit{scalable architectures}~\cite{2022FLASH,frantar2023sparsegptmassivelanguagemodels}, \textit{dynamic routing}~\cite{2022switchtransformers,2024deepseekv3,2024consistentee}, and \textit{KV cache management}~\cite{2023h2o,2023scissorhands} for long-horizon memory.

\item \textit{Test-time Adaptation of FM} (\secref{sec:adaptation}).  
To remain robust under data \textit{shifts}, \textit{partial observability}, and \textit{long-horizon tasks}, agentic AI systems must adapt FMs' parameter weights \textit{online} without full retraining.  
This can be achieved via \textit{algorithmic strategies}—prompt learning~\cite{2020autoprompt,2024mobilegpt,2023promptfl}, PEFT~\cite{2024FedPEFT,2022lora}, memory augmentation~\cite{2023Gisting,promptagent,LCD_PCW}, interactive learning~\cite{2024simuser,2022InstructGPT,2024emma}, and \textit{system-level techniques} such as memory management~\cite{2025memo,2025silvestre,2024edge_LLM}, scheduling~\cite{2022flashattention,2023flashattention2,2024data_juicer}, and distributed adaptation~\cite{2025survey_federated,2023FedPepTAO}.  
Beyond single agents, collaborative test-time adaptation leverages data, pipeline, and heterogeneous processor parallelism in multi-agent systems.

\item \textit{Dynamic Multi-modal FMs} (\secref{Dynamic Multi-modal FMs}).  
Mobile and edge agents must integrate heterogeneous sensor streams (\eg vision, speech, LiDAR, RF), which increases computation, memory, and alignment overhead.  
Dynamic multi-modal FMs employ \textit{dynamic attention}~\cite{zhang2025a-vl,limminference}, \textit{routing}~\cite{shen2024mome,xin2025i2moe}, \textit{alignment}~\cite{2024xmodalmoe,han2024fusemoe}, and \textit{token compression}~\cite{zhangspargeattention} to improve efficiency, consistency, and scalability under tight constraints.

\item \textit{Agentic AI Applications} (\secref{Agentic AI Applications}).  
Applications such as embodied agents~\cite{driess2023palm,shen2023hugginggpt}, GUI agents~\cite{wang2025mp,chen2024gui}, generative agents~\cite{park2024generative,adornetto2025generative}, and personal assistive agents~\cite{2024pac,zhang2024privacyasst} demonstrate how elastic inference and adaptive retraining integrate FM capabilities into resource-aware, context-sensitive services, highlighting the importance of \textit{application-driven optimization}.
\end{itemize}

\begin{figure*}[t]
    \centering
    \includegraphics[width=0.9\textwidth]{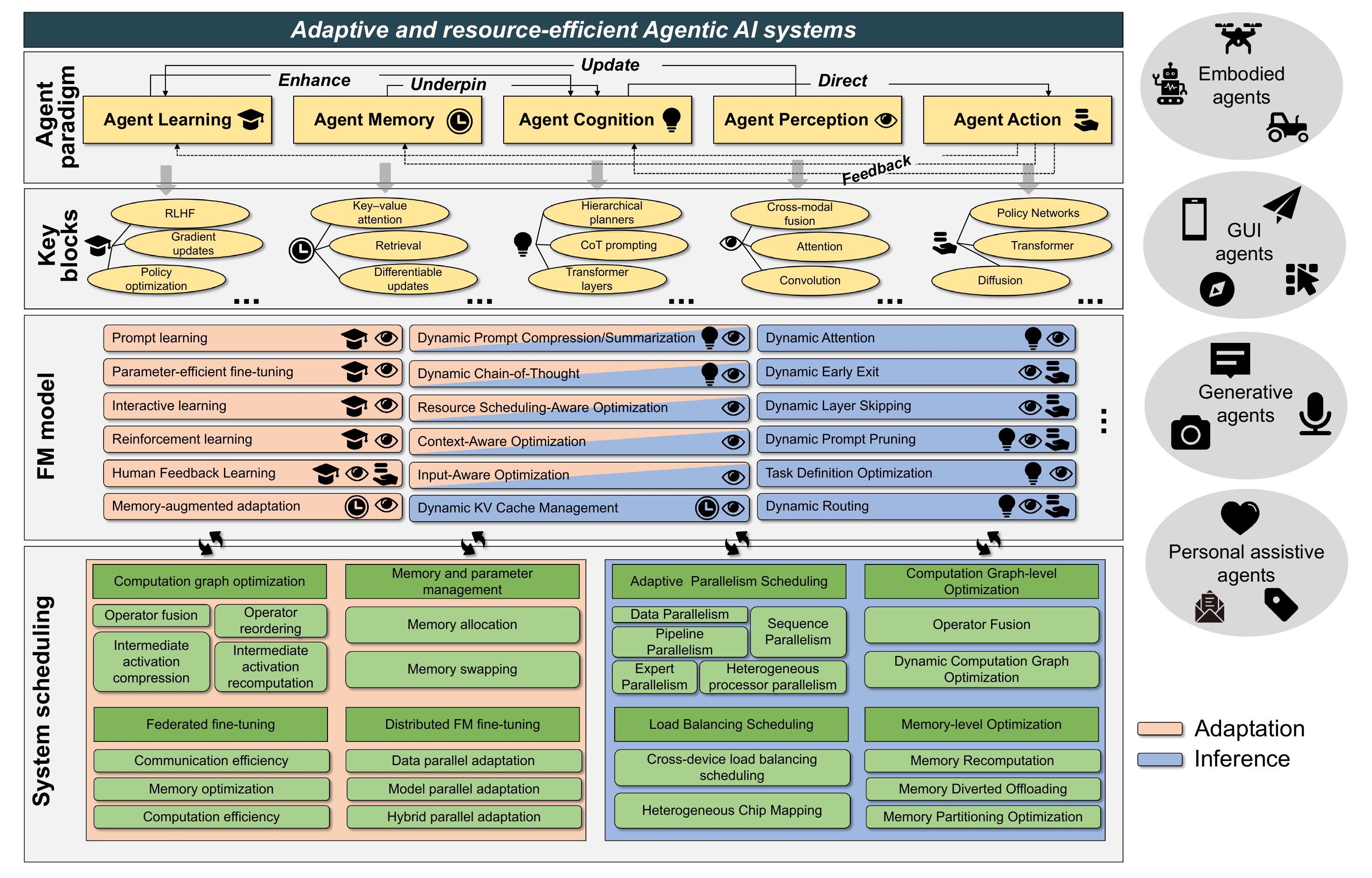}
    \caption{Dynamically adaptive and resource-efficient agentic AI system workflow.}
    \label{fig:agent_workflow}
    \vspace{-6mm}
\end{figure*}

\textbf{Workflow.}  
An FM-powered agentic AI system follows a closed-loop workflow of \textit{perception}, \textit{cognition}, \textit{memory}, \textit{learning}, and \textit{action} across one or multiple devices (\figref{fig:agent_workflow}).  
Each module is optimized as follows:  
\begin{itemize} 
\item \textit{Perception}: ingests multi-modal signals (\eg vision, speech, sensors, text) and encodes them into embeddings. Optimizations such as token quantization, prompt pruning, and lightweight feature extraction reduce latency and energy on constrained mobile/edge devices (see \secref{subsubsec:dynamic_prompt_pruning}, \secref{subsubsec:dynamic_pruning}, \secref{subsubsec:dynamic_quantization}). 

\item \textit{Cognition}: performs reasoning, goal decomposition, and decision planning via transformer layers, CoT prompting, and hierarchical planners. Dynamic routing, adaptive attention, and elastic CoT enable scalable inference under varying resource and performance demands (\secref{subsec:dynamic_routing}, \secref{subsubsec:dynamic_attention}, \secref{subsec:dynamic_chain_of_thought}). 

\item \textit{Memory}: maintains episodic and semantic context through key–value attention, retrieval, and differentiable updates. 
Efficiency is improved via quantization, knowledge distillation, and KV cache management (\secref{subsubsec:dynamic_quantization}, \secref{subsubsec:dynamic_knowledge_distillation}, \secref{subsec:dynamic_kv_cache_management}, \secref{sec:memory_aug_adaptation}). 

\item \textit{Agent Learning}: adapts to non-stationary data/tasks through test-time adaptation, policy optimization, and RLHF. 
Interactive learning, prompt tuning, and PEFT also provide efficient adaptation under tight budgets (\secref{sec:prompt_learning}, \secref{sec:peft},\secref{sec:interactive_learning}). 

\item \textit{Action}: executes decisions and generates outputs via policy networks, decoders, and diffusion models. 
Early exiting and layer skipping enhance responsiveness on mobile/edge hardware (see \secref{Dynamic Routing}). 

\item \textit{System scheduling} orchestrates elastic inference and adaptation through operator fusion, runtime scheduling, heterogeneous resource allocation for inference (see \secref{subsubsec:adaptive_parallelism_scheduling}, \secref{subsubsec:computation_graph_level_optimization}, \secref{subsubsec:load_balancing_scheduling}) and memory scheduling, computation graph optimization for retraining (\secref{sec:memory_parameter_management}, \secref{sec:computation_graph_level}). 
\end{itemize}
Together, they push the trade-off boundary between accuracy, latency, memory, and energy, enabling practical deployment of agentic AI systems on mobiles, wearables, and edge. 

\subsection{Performance Metrics}
In both inference and retraining, agentic AI systems must balance user goals (\eg accuracy, latency, energy efficiency) with dynamic device constraints (\eg memory hierarchy, battery life).
We summarize the key metrics below.

\textit{Accuracy}.
Accuracy is fundamental for reliable task execution. 
Beyond standard measures (\eg classification accuracy, perplexity, BLEU/ROUGE, factuality~\cite{brown2020language}), task-level indicators such as planning correctness, termination error rate, and task success rate can evaluate the end-to-end reliability.

\textit{Latency}.
Low latency is essential for interactive responsiveness. 
Besides \textit{end-to-end latency}, fine-grained metrics include:
\textit{Time To First Token (TTFT)}~\cite{TTFT,2024etalon}, reflecting initial responsiveness;
\textit{Time Per Output Token (TPOT)}~\cite{2024etalon}, indicating sustained throughput; and
\textit{Time Between Tokens (TBT)}~\cite{2024etalon}, where long gaps reduce naturalness.

\textit{Memory footprint}.
Large FM parameters and activations make memory occupancy a primary bottleneck. Key metrics include
total memory budget (executability), SRAM utilization (reducing DRAM traffic), and cache hit rate (data reuse efficiency)~\cite{swapmoe,2022flashattention,2025cache_hit}.
Efficient memory access is also critical for both latency and energy~\cite{2024melting}.

\textit{Computational Cost}.
Measured by multiply–accumulate (MAC) operations or FLOPs, computational cost directly impacts latency and energy. Retraining is significantly more expensive than inference due to additional backward passes.

\textit{Energy Efficiency}.
Measured by multiply–accumulate (MAC) operations or FLOPs, computational cost directly impacts latency and energy. Retraining is significantly more expensive than inference due to additional backward passes.

\begin{figure}[t]
    \centering
    \includegraphics[width=0.35\textwidth]{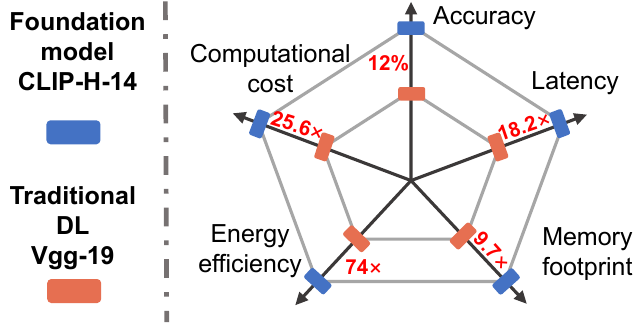}
    \caption{Performance comparison of FMs and traditional DL.}
    \vspace{-6mm}
    \label{fig:Performance_Metrics}
\end{figure}

As shown in \figref{fig:Performance_Metrics}, VGG-19~\cite{simonyan2014very} trained on ImageNet-1K achieves 72.37\% Top-1 accuracy, while CLIP (ViT-H-14-378)~\cite{fang2023data} attains 84.37\% zero-shot accuracy without supervised training on ImageNet. 
This accuracy advantage of foundation models is offset by substantial overheads: CLIP requires 18.2$\times$ FLOPs (GMacs), 9.7$\times$ latency, 74$\times$ peak memory, and 25.6$\times$ energy compared to VGG-19, intensifying the deployment challenges on constrained platforms.
The system must therefore perceive both environmental inputs and resource availability, and leverage timely online performance prediction and validation to balance accuracy, latency, memory, and energy cost for practical deployment.

\section{Elastic FM Inference in Agentic AI Systems}
\label{sec:elastic_fm_inference_in_agentic_ai_systems}

\begin{figure}[t]
    \centering
    \includegraphics[width=0.5\textwidth]{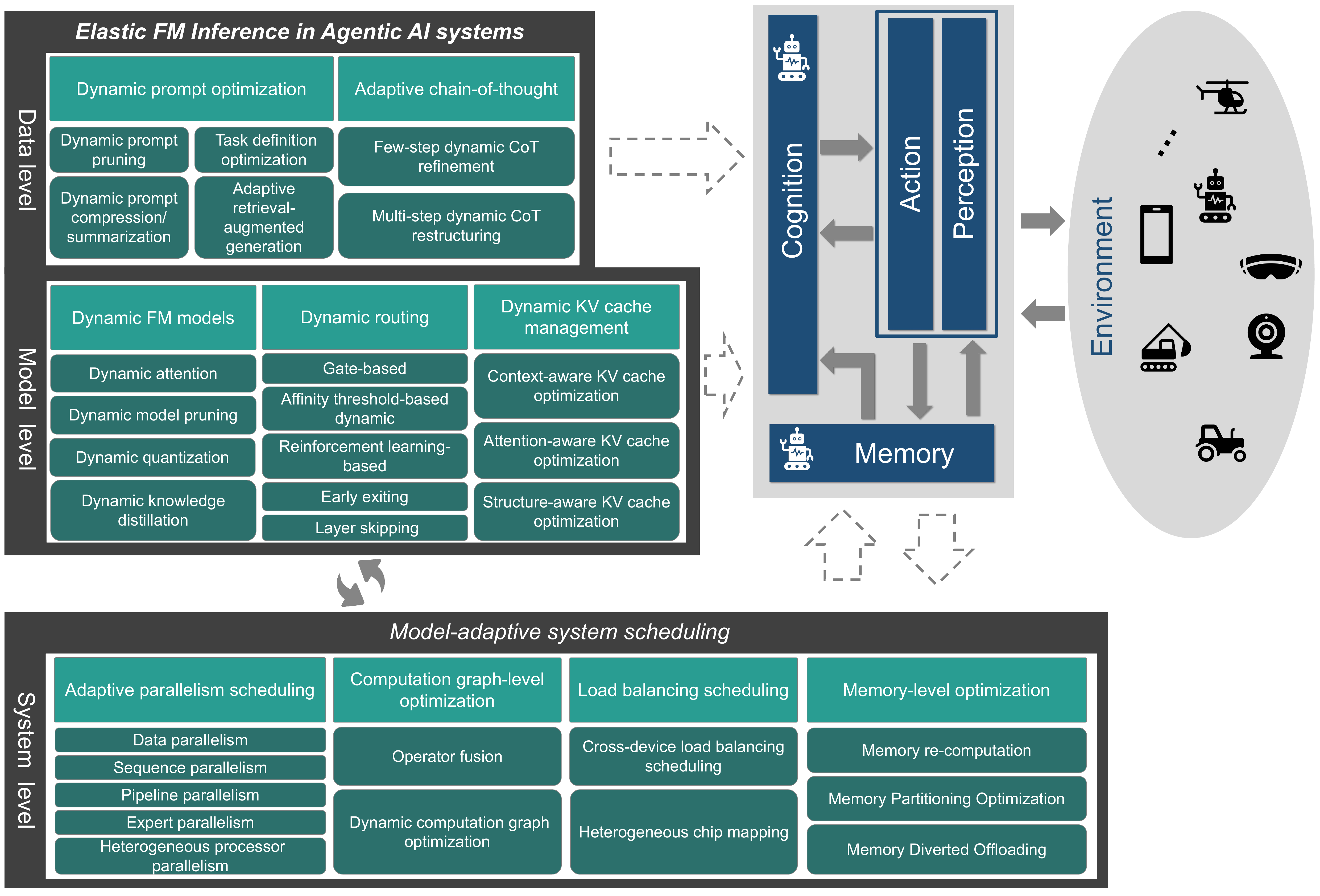}
    \caption{Overview of elastic FM inference techniques.} 
    \vspace{-6mm}
\label{fig:inference_agent}
\end{figure}

It is critical for agentic AI systems to meet diverse user demands (\eg accuracy, latency, energy efficiency) and adapt to dynamic device/network constraints (\eg memory, battery, bandwidth). 
Unlike static pipelines, elastic FM inference derives a new paradigm where \textit{FM structure}, \textit{reasoning depth}, \textit{computation cost}, and \textit{resource allocation} become \textit{runtime-adaptive} and \textit{tunable}, enabling sustainable deployment on mobile, wearable, and distributed agent platforms.  
As shown in~\figref{fig:inference_agent}, we summarize recent advances to highlight multiple aspects of elasticity: 
\textit{i) dynamic prompt optimization} reduces input complexity while preserving accuracy; \textit{ii) adaptive chain-of-thought} expands reasoning selectively; 
\textit{iii) dynamic FM models} allow scalable depth/width; 
\textit{iv) dynamic routing} enables path selection at token or layer level; 
and \textit{v) dynamic KV cache management} controls memory for long-horizon interactions. 
They define a novel landscape that pushes beyond static inference, reshaping FMs into adaptive and resource-efficient backbones for agentic AI systems.

\subsection{Dynamic Prompt Optimization}\label{subsec:dynamic_prompt_optimization}

\begin{table*}[t]
\centering
\caption{Summary of dynamic prompt optimization techniques for efficient inference in agentic AI systems.}
\vspace{-2mm}
\tiny
\label{tab:dynamic_prompt_optimization}
\renewcommand{\arraystretch}{1.1}
\setlength{\tabcolsep}{6pt}
\resizebox{\textwidth}{!}{%
\begin{tabular}{|c|c|c|c|c|}
\hline
\multicolumn{2}{|c|}{\textbf{Categories}} & 
\multicolumn{1}{c|}{\textbf{Technique highlight}} & 
\multicolumn{1}{c|}{\textbf{Year}} & 
\multicolumn{1}{c|}{\textbf{Ref}} \\
\hline

\multirow{13}{*}[\dimexpr-0ex\relax]{\centering\textbf{\begin{tabular}{c}Dynamic prompt \\ optimization \\~(\S\ref{subsec:dynamic_prompt_optimization})\end{tabular}}} 
& \multirow{3}{*}[\dimexpr0ex\relax]{\centering\textbf{\begin{tabular}{c}Dynamic prompt \\ compression~(\S\ref{subsubsec:dynamic_prompt_compression_summarization})\end{tabular}}} 
& \begin{tabular}[c]{@{}c@{}}Compresses documents into concise textual summaries.\end{tabular} & 2024 & \cite{2023recomp} \\
\cline{3-5}
& & \begin{tabular}[c]{@{}c@{}}Compress text into a concise summary vector.\end{tabular} & 2023 & \cite{2023AutoCompressors} \\
\cline{3-5}
& & \begin{tabular}[c]{@{}c@{}}Compresses long prompts into a small set of “gist” tokens via modified attention masks.\end{tabular} & 2023 & \cite{2023Gisting} \\
\cline{2-5}

& \multirow{3}{*}[\dimexpr0ex\relax]{\centering\textbf{\begin{tabular}{c}Dynamic prompt \\ pruning ~(\S\ref{subsubsec:dynamic_prompt_pruning})\end{tabular}}} 
& \begin{tabular}[c]{@{}c@{}}Train a meta-controller to predict the number of context examples required.\end{tabular} & 2023 & \cite{2023DYNAICL} \\
\cline{3-5}
& & \begin{tabular}[c]{@{}c@{}}Uses reinforcement learning to directly edit discrete prompts for efficient compression .\end{tabular} & 2024 & \cite{2024PCRL} \\
\cline{3-5}
& & \begin{tabular}[c]{@{}c@{}}Converts GUI elements into a lightweight HTML tag system with functional attributes.\end{tabular} & 2024 & \cite{2024autodroid} \\
\cline{2-5}

& \multirow{3}{*}[\dimexpr0ex\relax]{\centering\textbf{\begin{tabular}{c}Adaptive retrieval- \\augmented generation \\~(\S\ref{subsubsec:adaptive_retrieval_augmented_generation})\end{tabular}}} 
& \begin{tabular}[c]{@{}c@{}}Combine parametric memory and non-parametric memory .\end{tabular} & 2020 & \cite{2020rag} \\
\cline{3-5}
& & \begin{tabular}[c]{@{}c@{}}Introduces a self-reflection mechanism with dynamic reflection tokens for real-time accuracy.\end{tabular} & 2024 & \cite{2023self-rag} \\
\cline{3-5}
& & \begin{tabular}[c]{@{}c@{}}Employs a prospective prediction mechanism to activate knowledge retrieval based on upcoming text analysis.\end{tabular} & 2023 & \cite{jiang2023flare} \\
\cline{3-5}
& & \begin{tabular}[c]{@{}c@{}}Use the prediction results of LLM to supervise the retrieval module.\end{tabular} & 2023 & \cite{2023replug} \\
\cline{2-5}

& \multirow{2}{*}[\dimexpr0ex\relax]{\centering\textbf{\begin{tabular}{c}Task definition \\ optimization ~(\S\ref{subsubsec:task_definition_optimization})\end{tabular}}} 
& \begin{tabular}[c]{@{}c@{}}Leverages language models’ self-reflection by converting feedback into verbal guidance.\end{tabular} & 2023 & \cite{shinn2023reflexion} \\
\cline{3-5}
& & \begin{tabular}[c]{@{}c@{}}Synergizes reasoning and acting by interleaving thought and action generation.\end{tabular} & 2023 & \cite{yao2023react} \\
\hline

\end{tabular}%
}
\vspace{-2mm}
\end{table*}

Due to the fundamental differences in input processing paradigms between FMs and traditional DL models, \textit{dynamic prompt optimization} for FMs has positioned as an emerging research focus.
Prompting is the primary way to guide responses effectively, but it consumes precious space in the input context window, and repeatedly encoding the same prompt leads to inefficiencies in computation~\cite{2023Gisting}.
Thus adaptive FM prompt optimization represents a novel paradigm addressing three core mobile challenges, \ie real-time application demands, environmental variability, and flexible user interactions.
This approach enhances Agentic AI's expressive power, enabling real-time adjustment of guidance strategies.
We systematically categorize dynamic prompting strategies for Agentic AI systems into four, \ie \textit{dynamic prompt compression}\cite{ 2023recomp, 2023AutoCompressors, 2023Gisting}, \textit{prompt prunning}\cite{2023DYNAICL, 2024autodroid, 2024PCRL}, \textit{adaptive Retrieval-Augmented Generation (RAG)}\cite{2023self-rag, 2020rag, jiang2023flare, 2023replug}, and \textit{task definition Optimization}\cite{2023Gisting, shinn2023reflexion, yao2023react}(in \tabref{tab:dynamic_prompt_optimization}).

\begin{figure}[t]
    \centering
    \begin{minipage}{0.6\linewidth}
        \centering
        \subfloat[Dynamic prompt compression.]{
            \includegraphics[height=0.6\textwidth]{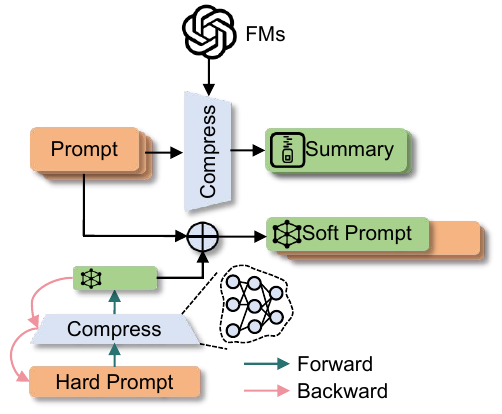}
            \label{fig:Dynamic prompt compression/summarization}
        }
    \end{minipage}
    \begin{minipage}{0.49\linewidth} 
        \centering
        \subfloat[Dynamic prompt pruning.]{
            \includegraphics[height=0.19\textwidth]{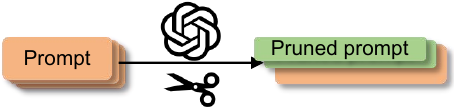}
            \label{fig:Dynamic prompt pruning}
        }
    \end{minipage}
    \begin{minipage}{0.49\linewidth} 
        \centering
        \subfloat[Adaptive RAG.]{
            \includegraphics[height=0.52\textwidth]{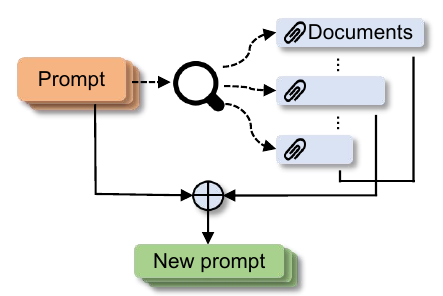}
            \label{fig:Adaptive RAG}
        }
    \end{minipage}
    \caption{Illustration of dynamic prompt optimization.}
    \label{fig:Dynamic Prompt}
    \vspace{-6mm}
\end{figure}

\subsubsection{Dynamic Prompt Compression/Summarization}
\label{subsubsec:dynamic_prompt_compression_summarization}
Dynamic prompt compression condenses long prompts into compact summaries that retain semantic relevance, serving as learnable soft prompts or direct FM inputs~\cite{2023AutoCompressors,2023recomp,2023Gisting}. 
Approaches include \textit{summarization accumulation}, \textit{segmentation}, \textit{task-adaptive extractive/abstractive compressors}, and \textit{attention-mask modification} (\figref{fig:Dynamic prompt compression/summarization}). AutoCompressors~\cite{2023AutoCompressors} compresses text into summary vectors used as soft prompts, trained via unsupervised objectives and optimized with summarization accumulation~\cite{bulatov2022RecurrentMemoryTransformer} and random segmentation. RECOMP~\cite{2023recomp} learns task-aware extractive and abstractive compressors for retrieval-augmented LMs, balancing compression rates with computation through selective document enhancement. Gisting~\cite{2023Gisting} achieves up to 26$\times$ compression by altering attention masks during fine-tuning, condensing prompts into gist tokens without retraining.

\subsubsection{Dynamic Prompt Pruning}\label{subsubsec:dynamic_prompt_pruning}
Dynamic prompt pruning~\cite{2023DYNAICL,2024autodroid,2024PCRL} selectively removes unimportant prompt content to reduce complexity, complementing compression and summarization methods (\figref{fig:Dynamic prompt pruning}). Strategies include meta-controller–based context allocation~\cite{2023DYNAICL}, token-level pruning~\cite{2024PCRL}, and UI element pruning~\cite{2024autodroid}. DYNAICL~\cite{2023DYNAICL} trains a FLAN-T5 meta-controller to predict context size per input, balancing accuracy with compute budget. PCRL~\cite{2024PCRL} applies reinforcement learning to edit prompts directly, pruning redundant tokens without model gradients or labeled data. AutoDroid~\cite{2024autodroid} prunes GUI/HTML elements by merging functionally equivalent nodes and discarding non-informative containers, lowering LLM processing overhead in mobile interaction tasks. Collectively, these methods adapt prompt length dynamically, improving efficiency while preserving task relevance.

\subsubsection{Adaptive Retrieval-Augmented Generation (RAG)}
\label{subsubsec:adaptive_retrieval_augmented_generation}
Retrieval-Augmented Generation (RAG) enhances FM accuracy and consistency by integrating external knowledge bases on demand~\cite{2020rag}, reducing long-context inputs while ensuring relevant retrieval at the right time (\figref{fig:Adaptive RAG}). Its advantages lie in \textit{indexing knowledge for efficient reasoning} and \textit{on-demand retrieval} for adaptive context construction.
Recent advances introduce \textit{adaptive} mechanisms to refine retrieval and generation. Self-RAG~\cite{2023self-rag,yu2023Chain-of-Note} adds \textit{self-reflection tokens} for real-time quality control. FLARE~\cite{jiang2023flare,gao2023HyDE} employs \textit{prospective prediction}, triggering retrieval based on upcoming semantic needs. REPLUG~\cite{2023replug,yu2023aar} applies a \textit{black-box strategy}, enhancing frozen FMs with tunable retrieval modules. AutoDroid~\cite{2024autodroid} integrates commonsense and app-specific knowledge via \textit{dynamic analysis}, enabling task automation across arbitrary Android apps.


\subsubsection{Task Definition Optimization}
\label{subsubsec:task_definition_optimization}
Task definition optimization seeks to refine prompts by \textit{compressing} or \textit{restructuring} them to improve efficiency without sacrificing effectiveness~\cite{2023Gisting,shinn2023reflexion,yao2023react}.  
In particular, Gisting~\cite{2023Gisting} modifies \textit{attention masks} during instruction fine-tuning to encode prompts into compact "Gist Tokens," which can be cached and reused then, saving computation and context window space.  
Reflexion~\cite{shinn2023reflexion} \textit{distills} failed trajectories into concise natural language summaries, integrating lessons learned into subsequent prompts via semantic compression.  
However, overly aggressive compression may induce hallucinations or loss of task control. 
To address this, ReAct~\cite{yao2023react} interleaves \textit{think–act–observe} trajectories in prompts, guiding explicit reasoning and interaction steps. 
This synergy between reasoning and acting enables dynamic planning, fact acquisition, and strategy adjustment, reducing factual hallucinations in traditional chain-of-thought prompting.  

\textbf{\textit{Discussion}}. 
Actuallu, dynamic prompting in agentic AI follows two paradigms (\figref{fig:Dynamic Prompt}), \ie \textit{rule-based} and \textit{data-driven}.
The \textit{rule-based} paradigm uses deterministic templates and adaptive triggers, ensuring verifiable behavior and ms-level latency for time-sensitive tasks, but struggles with unseen or dynamic scenarios. In contrast, the \textit{data-driven} paradigm leverages trainable models and retrieval augmentation for adaptive prompt optimization in knowledge-rich settings, offering flexibility at the cost of higher complexity and weaker determinism. 
These complementary strengths provide a framework for aligning prompting strategies with diverse agentic AI application requirements.

\begin{figure}[t]
    \centering
    \includegraphics[width=0.3\textwidth]{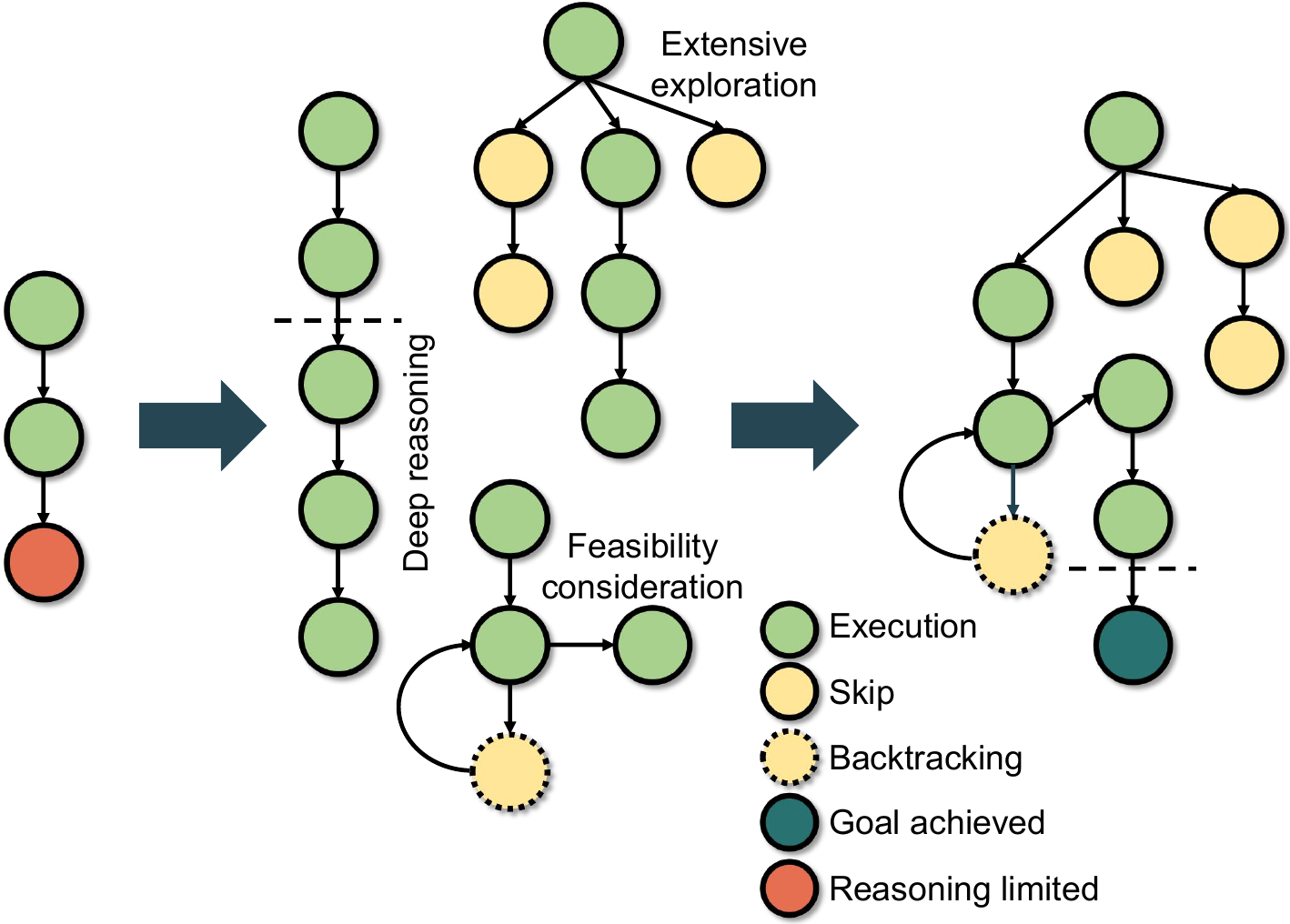}
    \caption{Compared to short CoT with few-step reasoning, long CoT features three characteristics: 1.deep reasoning, 2.extensive exploration, 3. feasibility consideration.}
    \label{fig:long_CoT}
    \vspace{-4mm}
\end{figure}

\subsection{Adaptive Chain-of-Thought}
\label{subsec:dynamic_chain_of_thought}
Chain of Thought (CoT) enables FMs to \textit{reason incrementally}, decomposing complex questions into stepwise rationales rather than producing direct answers. 
Dynamic CoT extends this capability with \textit{adaptive control}, allowing models to adjust reasoning depth and structure in response to task complexity, resource budgets, and stability demands. As summarized in \tabref{tab:adaptive_cot}, approaches fall into two categories: \textit{few-step adaptation}, which dynamically selects shorter or longer reasoning chains at runtime, and \textit{multi-step restructuring}, which revises or branches reasoning trajectories during execution for more reliable outcomes.

\begin{table*}[t]
\centering
\caption{Summary of adaptive Chain-of-Thought (CoT) techniques for efficient inference in agentic AI systems.}
\vspace{-2mm}
\tiny
\label{tab:adaptive_cot}
\renewcommand{\arraystretch}{1.1}
\setlength{\tabcolsep}{6pt}
\resizebox{\textwidth}{!}{%
\begin{tabular}{|c|c|c|c|c|}
\hline
\multicolumn{2}{|c|}{\textbf{Categories}} & 
\multicolumn{1}{c|}{\textbf{Technique highlight}} & 
\multicolumn{1}{c|}{\textbf{Year}} & 
\multicolumn{1}{c|}{\textbf{Ref}} \\
\hline

\multirow{14}{*}[\dimexpr 0ex\relax]{\centering\textbf{\begin{tabular}{c}Adaptive \\ chain-of-thought~(\S\ref{subsec:dynamic_chain_of_thought})\end{tabular}}} 
& \multirow{6}{*}[\dimexpr 0ex\relax]{\centering\textbf{\begin{tabular}{c}Few-step dynamic CoT \\ refinement ~(\S\ref{subsubsec:few_cot})\end{tabular}}} 
& \begin{tabular}[c]{@{}c@{}}Construct an autonomous agent to guide LLM reasoning.\end{tabular} & 2024 & \cite{2024agentinstruct} \\
\cline{3-5}
& & \begin{tabular}[c]{@{}c@{}}Obtain or construct corresponding demonstrations.\end{tabular} & 2024 & \cite{2023gem-cot} \\
\cline{3-5}
& & \begin{tabular}[c]{@{}c@{}}First output the answer skeleton and then generate each key point.\end{tabular} & 2024 & \cite{2023Skeleton-of-thought} \\
\cline{3-5}
& & \begin{tabular}[c]{@{}c@{}}Dynamically adjust the number of samplings for each question.\end{tabular} & 2023 & \cite{2023Adaptive-Consistency} \\
\cline{3-5}
& & \begin{tabular}[c]{@{}c@{}}Write the intermediate steps into the scratchpad.\end{tabular} & 2021 & \cite{2021Scratchpad} \\
\cline{3-5}
& & \begin{tabular}[c]{@{}c@{}}Think and record thoughts during the process of reading the context.\end{tabular} & 2023 & \cite{2023Self-Notes} \\
\cline{2-5}

& \multirow{5}{*}[\dimexpr 0ex\relax]{\centering\textbf{\begin{tabular}{c}Multi-step dynamic CoT \\ restructuring~(\S\ref{subsubsec:muti_cot})\end{tabular}}} 
& \begin{tabular}[c]{@{}c@{}}Synergizes large language models with program interpreters .\end{tabular} & 2023 & \cite{gao2023palprogramaidedlanguagemodels} \\
\cline{3-5}
& & \begin{tabular}[c]{@{}c@{}}Explore and evaluate multiple reasoning paths.\end{tabular} & 2023 & \cite{yao2023tree} \\
\cline{3-5}
& & \begin{tabular}[c]{@{}c@{}}Represents “thoughts” as a directed graph with dependencies and feedback.\end{tabular} & 2024 & \cite{Besta_2024} \\
\cline{3-5}
& & \begin{tabular}[c]{@{}c@{}}Generates and utilizes self-feedback without additional training.\end{tabular} & 2023 & \cite{madaan2023self} \\
\cline{3-5}
& & \begin{tabular}[c]{@{}c@{}}Combines reinforcement learning-driven rule rewards and tree search for deep logical exploration.\end{tabular} & 2025 & \cite{2025deepseek-r1} \\
\hline

\end{tabular}%
}
\vspace{-4mm}
\end{table*}

\subsubsection{Few-Step Dynamic CoT Refinement}
\label{subsubsec:few_cot}
Few-Step Dynamic CoT refers to a \textit{shallow, linear} reasoning process composed of a limited number of sequential nodes. 
Each step unidirectionally leads to the next without repetition or backtracking, making it suitable for simple, well-defined problems that require high speed and low resource consumption.  
It provides fast, resource-efficient reasoning, and its refinements focus on improving adaptability, efficiency, and structural robustness for simple or time-critical agentic AI tasks.
Specifically, recent studies explore three refinement directions, \ie \textit{task-driven adaptation}~\cite{2024agentinstruct,2023gem-cot,le2019automatic}, \textit{prompt-based efficiency}~\cite{2023Skeleton-of-thought,2023Adaptive-Consistency}, and \textit{structural optimization}~\cite{2021Scratchpad,2023Self-Notes}.

\textbf{\textit{a. Task-driven adaptation.}}  
Several methods enhance LLMs’ reasoning across diverse tasks.  
AgentInstruct~\cite{2024agentinstruct} guides zero-shot reasoning via agent-generated task-specific instructions.
GeM-CoT~\cite{2023gem-cot} selects demonstrations by question type for stronger generalization.
Automatic CoT~\cite{le2019automatic} retrieves similar examples or falls back to zero-shot reasoning to update its demo pool.
COSP samples multiple reasoning paths and re-prompts with the best candidate.
Reprompting iteratively learns effective prompts via Gibbs sampling.  

\begin{figure}[t]
    \centering
    \includegraphics[width=0.48\textwidth]{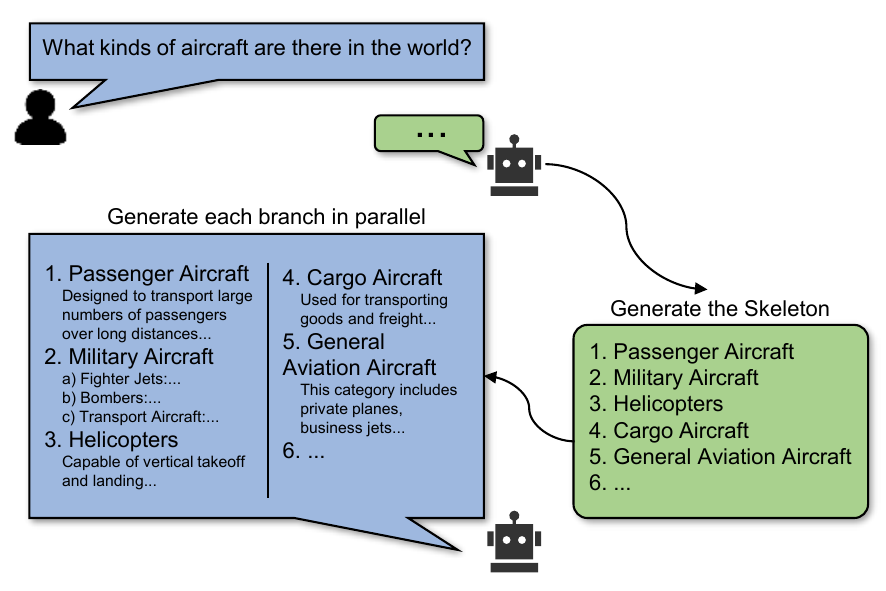}
    \caption{SoT generates an answer framework and then generates finer branches in parallel to achieve acceleration.}
    \label{fig:SoT}
    \vspace{-5mm}
\end{figure}

\textbf{\textit{b. Prompt-based efficiency.}}  
Skeleton-of-Thought~\cite{2023Skeleton-of-thought} decomposes answers into frameworks first, enabling parallel generation and reducing latency. \figref{fig:SoT} illustrates the working principle of SoT.
Adaptive-Consistency~\cite{2023Adaptive-Consistency} improves self-consistency sampling by dynamically adjusting sample size with lightweight stopping criteria.  

\textbf{\textit{c. Structural optimization.}}  
Scratchpad~\cite{2021Scratchpad} introduces an intermediate buffer for storing reasoning steps, allowing adaptive allocation of computation.  
Self-Notes~\cite{2023Self-Notes} extends this idea by letting LLMs generate notes in real time during reading and answering, enhancing memory capacity and multi-step reasoning ability.  

\subsubsection{Multi-Step Dynamic CoT Restructuring}
\label{subsubsec:muti_cot}
Unlike the fixed or shallow reasoning of short CoT, multi-step CoT enhances complex task solving by introducing \textit{dynamic adaptability} through three mechanisms, \ie \textit{depth expansion}~\cite{chowdhery2022palm, liu2023mathematical}, \textit{breadth exploration}~\cite{yao2023tree, Besta_2024}, and \textit{self-refinement}~\cite{shinn2023reflexion, madaan2023self}. \figref{fig:long_CoT} illustrates these typical characteristics of multi-step CoT.
This paradigm is particularly effective in \textit{mathematical}, \textit{programming}, and \textit{cross-domain} reasoning where long, adaptive chains are required. 

\textbf{\textit{a. Depth expansion. }} 
Multi-step CoT can extend reasoning depth beyond the node limits of short chains, developing \textit{incremental hierarchical logic} only when problem complexity demands it.  
For example, Program-Aided Language Models (PAL)~\cite{gao2023palprogramaidedlanguagemodels} dynamically invoke program execution to extend reasoning, and MathPrompter~\cite{liu2023mathematical} adaptively expands reasoning depth for difficult mathematical problems.  

\textbf{\textit{b. Breadth exploration.}}  
Moving from linear to tree-structured reasoning ($n_i \rightarrow n_{i+j}$), multi-step CoT can perform \textit{dynamic branching} to generate and evaluate parallel reasoning paths, pruning or prioritizing them based on resource budgets.
Examples include Tree-of-Thoughts (ToT)~\cite{yao2023tree} and Graph-of-Thoughts~\cite{Besta_2024}, which adaptively expand or collapse search paths to balance coverage and efficiency.  

\textit{\textbf{c. Self-refinement.}  }
Feedback loops enable dynamic verification and correction of intermediate steps, reallocating computation when inconsistencies arise. 
Approaches such as Reflexion~\cite{shinn2023reflexion} adapt reasoning by leveraging past errors, while Self-Refine~\cite{madaan2023self} iteratively critiques and edits outputs until convergence, embodying dynamic correction.

These mechanisms can also be combined adaptively.  
For example, DeepSeek-R1~\cite{2025deepseek-r1} integrates reinforcement learning with multi-stage reward signals (\textit{depth expansion}), tree search-based exploration (\textit{breadth exploration}), and dynamic feedback with self-correction (\textit{self-refinement}), achieving deeper reasoning hierarchies and higher-precision outputs in challenging mathematical and programming tasks.

\subsection{Dynamic FM Model}
\label{subsec:dynamic_fms_model}
Dynamically scaling FM models is a basic solution to resource limitations in embedded and embodied agents. 
Agent platforms always face strict power budgets and latency-sensitive demands, making structural optimization of FMs essential.
We summarize main directions of dynamic model design:  
(i) \textit{dynamic attention}~\cite{2022FLASH,2024sparsekattention,2021RoutingTransformer,2025NSA} allocates elastic attention computation for multi-granularity feature perception;  
(ii) \textit{dynamic pruning}~\cite{frantar2023sparsegptmassivelanguagemodels} removes redundant structures at runtime to cut FLOPs;  
(iii) \textit{dynamic quantization}~\cite{lin2024awqactivationawareweightquantization,wei2025t-mac,2024edge_LLM} adapts precision levels to balance efficiency and accuracy; and  
(iv) \textit{dynamic knowledge distillation}~\cite{sanh2020distilbertdistilledversionbert,sun2019patientknowledgedistillationbert} transfers knowledge into compact models for efficient inference(in \tabref{tab:dynamic_fm_model}).
Beyond these structural approaches, mechanisms include (v) \textit{dynamic routing} for adaptive path selection and resource scheduling, and (vi) \textit{dynamic KV cache management} for compressing long-sequence caches, jointly improving efficiency under tight resource budgets.

\begin{table*}[t]
\centering
\caption{Summary of dynamic FM model structures for elastic inference in agentic AI systems.}
\vspace{-2mm}
\tiny
\label{tab:dynamic_fm_model}
\renewcommand{\arraystretch}{1.1}
\setlength{\tabcolsep}{6pt}
\resizebox{\textwidth}{!}{%
\begin{tabular}{|c|c|c|c|c|}
\hline
\multicolumn{2}{|c|}{\textbf{Categories}} & 
\multicolumn{1}{c|}{\textbf{Technique highlight}} & 
\multicolumn{1}{c|}{\textbf{Year}} & 
\multicolumn{1}{c|}{\textbf{Ref}} \\
\hline
\multirow{18}{*}[-1ex]{\centering\textbf{\begin{tabular}{c}Dynamic FM model \\~(\S\ref{subsec:dynamic_fms_model})\end{tabular}}} 
& \multirow{8}{*}[-1ex]{\centering\textbf{\begin{tabular}{c}Dynamic \\attention~(\S\ref{subsubsec:dynamic_attention})\end{tabular}}} 
& \begin{tabular}[c]{@{}c@{}}By unifying gating mechanisms through the Gated Attention Unit(GAU).\end{tabular} & 2022 & \cite{2022FLASH} \\
\cline{3-5}
& & \begin{tabular}[c]{@{}c@{}}Integrate a scoring network and a top-k mask to select KV pairs for each query.\end{tabular} & 2024 & \cite{2024sparsekattention} \\
\cline{3-5}
& & \begin{tabular}[c]{@{}c@{}}By controlling sparsity through a threshold.\end{tabular} & 2024 & \cite{2024squeezedattention} \\
\cline{3-5}
& & \begin{tabular}[c]{@{}c@{}}Content-based sparse attention, incorporate a sparse routing module.\end{tabular} & 2021 & \cite{2021RoutingTransformer} \\
\cline{3-5}
& & \begin{tabular}[c]{@{}c@{}}Achieve sparsity through a fixed window combined with attention aggregation.\end{tabular} & 2025 & \cite{2025NSA} \\
\cline{3-5}
& & \begin{tabular}[c]{@{}c@{}}Decouple storage demands across attention heads based on their operational roles.\end{tabular} & 2024 & \cite{2024duoattention} \\
\cline{3-5}
& & \begin{tabular}[c]{@{}c@{}}Make coarse-grained and fine-grained attention divisions for tokens.\end{tabular} & 2023 & \cite{2023biformer} \\
\cline{3-5}
& & \begin{tabular}[c]{@{}c@{}}Use the Gaussian probability density function to compute attention weights.\end{tabular} & 2024 & \cite{2024gaam} \\
\cline{2-5}

& \multirow{3}{*}[0ex]{\centering\textbf{\begin{tabular}{c}Dynamic model \\pruning~(\S\ref{subsubsec:dynamic_pruning})\end{tabular}}} 
& \begin{tabular}[c]{@{}c@{}}Enables one-shot pruning of large GPT models to $50\%$ sparsity.\end{tabular} & 2023 & \cite{frantar2023sparsegptmassivelanguagemodels} \\
\cline{3-5}
& & \begin{tabular}[c]{@{}c@{}}Employs a lightweight prediction module to dynamically prune redundant tokens.\end{tabular} & 2021 & \cite{rao2021dynamicvitefficientvisiontransformers} \\
\cline{3-5}
& & \begin{tabular}[c]{@{}c@{}}Leverages the early-bird lottery ticket hypothesis to identify sub-networks.\end{tabular} & 2021 & \cite{chen2021earlybertefficientberttraining} \\
\cline{2-5}

& \multirow{3}{*}[-0ex]{\centering\textbf{\begin{tabular}{c}Dynamic \\quantization~(\S\ref{subsubsec:dynamic_quantization})\end{tabular}}} 
& \begin{tabular}[c]{@{}c@{}}Protects salient weights based on activation distribution.\end{tabular} & 2024 & \cite{lin2024awqactivationawareweightquantization} \\
\cline{3-5}
& & \begin{tabular}[c]{@{}c@{}}Data-free quantization-aware training supports 8-bit weight and activation quantization.\end{tabular} & 2023 & \cite{liu2023llmqatdatafreequantizationaware} \\
\cline{3-5}
& & \begin{tabular}[c]{@{}c@{}}Uses tensor decomposition and mixed precision for low-bit inference on CPU/NPU.\end{tabular} & 2025 & \cite{wei2025t-mac} \\
\cline{2-5}

& \multirow{3}{*}[-0ex]{\centering\textbf{\begin{tabular}{c}Dynamic knowledge \\distillation~(\S\ref{subsubsec:dynamic_knowledge_distillation})\end{tabular}}} 
& \begin{tabular}[c]{@{}c@{}}Two-stage Transformer distillation framework compresses BERT.\end{tabular} & 2020 & \cite{jiao2020tinybertdistillingbertnatural} \\
\cline{3-5}
& & \begin{tabular}[c]{@{}c@{}}Pre-trains a smaller general-purpose model.\end{tabular} & 2020 & \cite{sanh2020distilbertdistilledversionbert} \\
\cline{3-5}
& & \begin{tabular}[c]{@{}c@{}}Replaces KL divergence objective for efficient knowledge distillation.\end{tabular} & 2024 & \cite{gu2024minillmknowledgedistillationlarge} \\
\hline

\end{tabular}%
}
\vspace{-4mm}
\end{table*}

\subsubsection{Dynamic Attention}
\label{subsubsec:dynamic_attention}
Multi-head attention (MHA) is a core module in both \textit{decoder-only} and \textit{encoder–decoder} FMs, capturing complex sequence dependencies. 
Dynamic MHA has been enhanced through \textit{sparse attention}~\cite{2022flashattention,2024sparsekattention,2024squeezedattention} and \textit{hierarchical attention}~\cite{2024duoattention,2023biformer,2024gaam}, which reduce redundant operations and improve adaptability. 
\textit{Sparse attention} limits the number of active connections in the attention matrix, while \textit{hierarchical attention} prioritizes critical layers or regions, jointly improving memory and computational efficiency.

\begin{figure}[t]
    \centering
\includegraphics[width=0.32\textwidth]{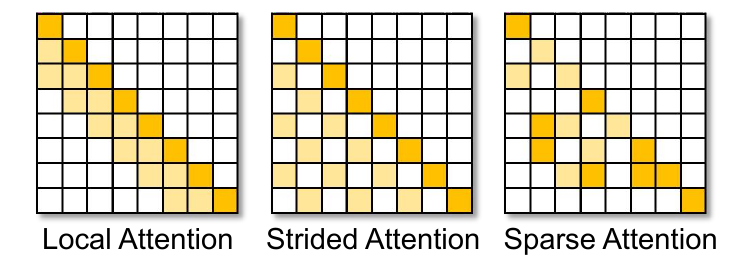}
    \caption{Sparse attention.}
    \label{fig:sparse_attention}
    \vspace{-4mm}
\end{figure}

\textbf{\textit{a. Sparse attention.}}
Sparse attention \textit{selectively activates} critical units, reducing complexity while preserving representational power (\figref{fig:sparse_attention}).
Key strategies include differentiable routing (\eg gated networks) for \textit{dynamic unit selection} and \textit{input-adaptive allocation}.
FLASH~\cite{2022FLASH} integrates gating with simplified attention via the Gated Attention Unit (GAU), replacing costly multi-head softmax with a lightweight single-head design.
SPARSEK Attention~\cite{2024sparsekattention} combines a scoring network with differentiable top-k masking, selecting a fixed number of KV pairs per query to achieve linear time and constant memory.
Squeezed Attention~\cite{2024squeezedattention} applies query-aware dynamic KV selection with Softmax-based sparsity control, outperforming fixed pruning (\eg SnapKV) by better balancing accuracy and efficiency.

\textit{\textbf{b. Focused dynamic attention}.}
It is another common approach that partitions sequences into \textit{overlapping} or \textit{non-overlapping windows} with \textit{predefined} or \textit{dynamic} sizes.
Each query attends to local neighbors and a few global nodes, while cross-window correlations are built via relative positional encoding and window shifting.
Routing Transformer~\cite{2021RoutingTransformer} combines content-based clustering with local and temporal sparse attention, improving flexibility and efficiency.
NSA~\cite{2025NSA} employs fixed windows with attention aggregation, using coarse-grained tokens for global context, fine-grained tokens for details, and sliding windows for local dependencies.
These designs preserve global context while overcoming the limitations of static sparse patterns in long-sequence modeling.

\begin{figure}[t]
    \centering
    \includegraphics[width=0.4\textwidth]{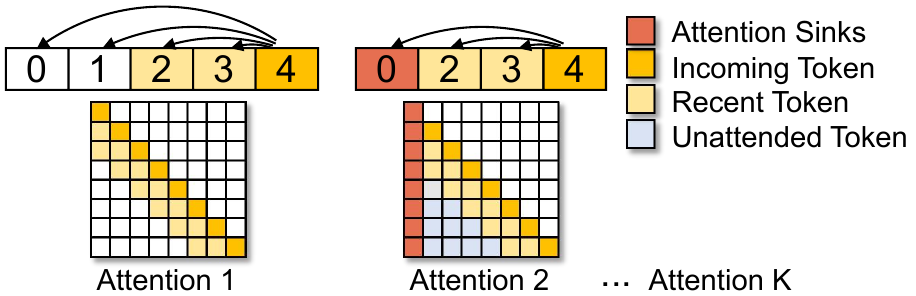}
    \caption{Hierarchical attention.}
    \label{fig:hierarchical_attention}
    \vspace{-4mm}
\end{figure}

\textit{\textbf{c. Hierarchical/Progressive attention}.}
Hierarchical attention organizes computation into \textit{layered structures} to capture multi-scale features. 
Unlike sparse attention, it stresses inter-layer collaboration, lower layers extract fine-grained details, while higher layers encode coarse-grained semantics. 
Two main strategies are used, \ie \textit{decoupled processing} and \textit{progressive granularity}.
\textit{First}, \textit{decoupled attention} separates functions or context spans for efficiency. 
DuoAttention~\cite{2024duoattention} assigns retrieval heads to long-range KV caches and streaming heads to recent tokens, showing that only a few retrieval heads suffice for long contexts, enabling pruning and selective memory usage.
\textit{Second}, \textit{progressive attention} builds multi-granularity layers. 
Early stages apply fine-grained attention (\eg token/pixel), while later stages use coarse-grained forms (\eg sentence/region). 
BiFormer~\cite{2023biformer} partitions tokens by granularity for efficient routing, while GAAM~\cite{2024gaam} employs Gaussian-based heads that dynamically adjust focus via learned means and variances, providing distribution-aware expressiveness.

\subsubsection{Dynamic Model Pruning}
\label{subsubsec:dynamic_pruning}
Model pruning reduces resource cost and latency by removing redundant parameters or structures. The key idea is to adaptively eliminate weights or neurons with minimal contribution to specific inputs, tasks, or outputs.
It typically include two types:
\textit{Unstructured pruning} offers fine-grained control by removing individual weights, but the resulting sparse matrices often underutilize hardware~\cite{frantar2023sparsegptmassivelanguagemodels}.
\textit{Structured pruning}, in contrast, removes whole neurons~\cite{chen2021earlybertefficientberttraining}, attention heads~\cite{xia2024shearedllamaacceleratinglanguage}, or channels~\cite{rao2021dynamicvitefficientvisiontransformers,li2024snapkvllmknowslooking}, preserving dense formats that map more efficiently to hardware. 


\subsubsection{Dynamic Quantization}\label{subsubsec:dynamic_quantization}
Quantization reduces overhead by lowering parameter precision. Converting FP32 weights to INT8, for example, cuts memory by $4\times$ and accelerates inference with integer operations. Two main approaches exist: \textit{post-training quantization}~\cite{frantar2023gptqaccurateposttrainingquantization,lin2024awqactivationawareweightquantization} and \textit{quantization-aware training}~\cite{liu2023llmqatdatafreequantizationaware,dettmers2023qloraefficientfinetuningquantized}.
However, both often require dequantization during inference to handle mixed precision, adding latency and memory bandwidth overhead (\eg converting INT8 back to FP16 for matrix multiplication). To overcome this, T-MAC~\cite{wei2025t-mac} employs bitwise operation lookup tables for direct low-bit computation, eliminating dequantization. Edge-LLM~\cite{cai2024edge} further introduces adaptive quantization, feature caching, and value-density-based scheduling in a server–edge framework, improving utilization and inference speed.

\subsubsection{Dynamic Knowledge Distillation}
\label{subsubsec:dynamic_knowledge_distillation}
Dynamic knowledge distillation compresses FMs by transferring knowledge from heavy “teacher” models to lightweight “student” models. Key strategies include:
\textit{Adaptive temperature scaling}, which adjusts softmax temperatures to emphasize different levels of teacher knowledge~\cite{sanh2020distilbertdistilledversionbert,wang2020minilmdeepselfattentiondistillation};
\textit{Progressive transfer}, which gradually distills layer-wise knowledge to balance efficiency and accuracy~\cite{jiao2020tinybertdistillingbertnatural,sun2019patientknowledgedistillationbert}; and
\textit{Dynamic weighting}, which reweights knowledge components according to task context~\cite{chen2021earlybertefficientberttraining,gu2024minillmknowledgedistillationlarge}.
Representative methods such as TinyBERT~\cite{jiao2020tinybertdistillingbertnatural}, Patient KD~\cite{sun2019patientknowledgedistillationbert}, and MiniLM~\cite{wang2020minilmdeepselfattentiondistillation} demonstrate that distillation can significantly reduce model size and inference latency while retaining accuracy.
By preserving multi-level contextual features, dynamic KD provides a scalable and adaptive solution for efficient long-context FM deployment.

\subsection{Dynamic Routing}
\label{subsec:dynamic_routing}
Dynamic routing enables \textit{input-aware} allocation of computation at the \textit{topological level}, determining which modules, layers, or branches are executed. 
Unlike dynamic attention that adapts feature-level computation, dynamic routing learns \textit{conditional mappings} between inputs and network topology, allowing flexible path selection. 
This adaptivity reduces redundancy for simple inputs while preserving accuracy on complex ones, thus balancing speed, accuracy, and efficiency.  
We summarize five main paradigms, \ie \textit{gate-based}~\cite{2022switchtransformers,2024Expertpruningandskipping,2023lina}, 
\textit{affinity threshold-based}~\cite{2024deepseekv3,2024Expert-Token-Resonance-MoE,2024Exflow}, \textit{reinforcement learning-based}~\cite{zhou2022mixture,2024consistentee},  
\textit{early exiting}~\cite{2023FREE,schuster2022calm,chen2023ee-llm,del2023skipdecode,2024edge_LLM}, and  
\textit{layer skipping}~\cite{kim2024shortenedllama,zeng2023skiplayer,jiang2024dllms}.  

\begin{table*}[t]
\centering
\caption{Summary of dynamic routing techniques for MoE-based FM inference.}
\vspace{-2mm}
\tiny
\label{tab:dynamic_routing}
\renewcommand{\arraystretch}{1.1}
\setlength{\tabcolsep}{6pt}
\resizebox{\textwidth}{!}{%
\begin{tabular}{|c|c|c|c|c|}
\hline
\multicolumn{2}{|c|}{\textbf{Categories}} & 
\multicolumn{1}{c|}{\textbf{Technique highlight}} & 
\multicolumn{1}{c|}{\textbf{Year}} & 
\multicolumn{1}{c|}{\textbf{Ref}} \\
\hline

\multirow{21}{*}[0ex]{\centering\textbf{\begin{tabular}{c}Dynamic routing \\~(\S\ref{subsec:dynamic_routing})\end{tabular}}} 
& \multirow{5}{*}[0ex]{\centering\textbf{\begin{tabular}{c}Gate-based dynamic \\routing with MoE \\~(\S\ref{subsubsec:gate_based})\end{tabular}}} 
& \begin{tabular}[c]{@{}c@{}}Uses a load balancing loss to penalize imbalanced distributions.\end{tabular} & 2022 & \cite{2022switchtransformers} \\
\cline{3-5}
& & \begin{tabular}[c]{@{}c@{}}Adjusts the gating function to be used to activate the next MoE module.\end{tabular} & 2024 & \cite{2024Pre-gatedmoe} \\
\cline{3-5}
& & \begin{tabular}[c]{@{}c@{}}Only keeps key experts during runtime and dynamically maintain the swapping in and out of experts in memory.\end{tabular} & 2024 & \cite{swapmoe} \\
\cline{3-5}
& & \begin{tabular}[c]{@{}c@{}}Set a threshold as the basis for expert skipping and dynamically skip certain experts.\end{tabular} & 2024 & \cite{2024Expertpruningandskipping} \\
\cline{3-5}
& & \begin{tabular}[c]{@{}c@{}}Dynamically schedule expert resources, predict each token's distribution in the next MoE layer by its expert path.\end{tabular} & 2023 & \cite{2023lina} \\
\cline{2-5}

& \multirow{4}{*}[0ex]{\centering\textbf{\begin{tabular}{c}Affinity threshold-based \\dynamic routing with MoE \\~(\S\ref{subsubsec:affinity_threshold_based})\end{tabular}}} 
& \begin{tabular}[c]{@{}c@{}}Introduces a learnable bias term for each expert, which is superimposed on the affinity scores.\end{tabular} & 2025 & \cite{2024deepseekv3} \\
\cline{3-5}
& & \begin{tabular}[c]{@{}c@{}}Pipeline optimization and hierarchical loading strategies are achieved through affinity-awareness.\end{tabular} & 2024 & \cite{2024aptmoe} \\
\cline{3-5}
& & \begin{tabular}[c]{@{}c@{}}A bidirectional selection routing framework based on expert-token resonance achieves efficient routing.\end{tabular} & 2024 & \cite{2024Expert-Token-Resonance-MoE} \\
\cline{3-5}
& & \begin{tabular}[c]{@{}c@{}}Exploit inter-layer affinity in pre-trained MoEs to optimize placement and routing with one AlltoAll.\end{tabular} & 2024 & \cite{2024Exflow} \\
\cline{2-5}

& \textbf{\begin{tabular}{c}
RL-based dynamic routing \\ 
~(\S\ref{subsubsec:rl_based})
\end{tabular}} 
& \begin{tabular}[c]{@{}c@{}}Model early exit as a reinforcement learning problem, use a "memory layer" to measure instance difficulty.\end{tabular} 
& 2024 & \cite{2024consistentee} \\
\cline{2-5}

& \multirow{6}{*}[0ex]{\centering\textbf{\begin{tabular}{c}Early exiting \\~(\S\ref{subsubsec:early_existing})\end{tabular}}} 
& \begin{tabular}[c]{@{}c@{}}Analyze key features to dynamically determine when to stop inference.\end{tabular} & 2024 & \cite{2022adainfer} \\
\cline{3-5}
& & \begin{tabular}[c]{@{}c@{}}Shallow deep modules and synchronous parallel decoding are combined.\end{tabular} & 2023 & \cite{2023FREE} \\
\cline{3-5}
& & \begin{tabular}[c]{@{}c@{}}Uses entropy of internal representations to compute confidence for adaptive early exiting.\end{tabular} & 2022 & \cite{schuster2022calm} \\
\cline{3-5}
& & \begin{tabular}[c]{@{}c@{}}Scales early-exit LLMs to hundreds of billions of parameters with 3D parallelism.\end{tabular} & 2024 & \cite{chen2023ee-llm} \\
\cline{3-5}
& & \begin{tabular}[c]{@{}c@{}}Uses column-wise unified exits and monotonically decreasing exit layers.\end{tabular} & 2023 & \cite{del2023skipdecode} \\
\cline{3-5}
& & \begin{tabular}[c]{@{}c@{}}Uses adaptive layer tuning with early exits/voting for memory-efficient full-model updates in edge LLMs.\end{tabular} & 2024 & \cite{2024edge_LLM} \\
\cline{2-5}

& \multirow{4}{*}[0ex]{\centering\textbf{\begin{tabular}{c}Layer skipping \\~(\S\ref{subsubsec:layer_skipping})\end{tabular}}} 
& \begin{tabular}[c]{@{}c@{}}Real speedups in small batches with one-shot depth pruning + pretraining, outperforming width pruning.\end{tabular} & 2024 & \cite{kim2024shortenedllama} \\
\cline{3-5}
& & \begin{tabular}[c]{@{}c@{}}Merges later into earlier via parameter differencing, for training-free structure-pruning over 80\% performance.\end{tabular} & 2024 & \cite{yang2024laco} \\
\cline{3-5}
& & \begin{tabular}[c]{@{}c@{}}Learns token-wise skipping via binary router, for end-to-end training/inference, save FLOPs, boost few-shot performance.\end{tabular} & 2023 & \cite{zeng2023skiplayer} \\
\cline{3-5}
& & \begin{tabular}[c]{@{}c@{}}Dynamically skips non-critical Transformer layers while evicting KV-cache, yielding plug-and-play 50\% inference savings.\end{tabular} & 2024 & \cite{jiang2024dllms} \\
\hline

\end{tabular}%
}
\vspace{-5mm}
\end{table*}

\begin{figure}[t]
    \centering
    \includegraphics[width=0.35\textwidth]{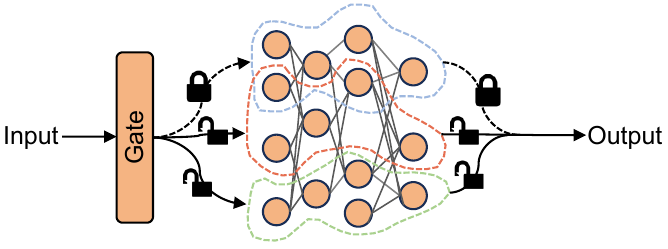}
    \caption{Illustration of gate-based dynamic routing mechnism.}
    \label{fig:gate_routing}
    \vspace{-4mm}
\end{figure}

\subsubsection{Gate-based Dynamic Routing with MoE}
\label{subsubsec:gate_based}
Gate-based dynamic routing enables \textit{input-aware subnetwork activation} via differentiable gating (\figref{fig:gate_routing}).
In the \textit{Mixture of Experts (MoE)} framework, a gating network assigns inputs across $N$ experts by producing a probability vector $\mathbf{g}(\mathbf{x}) \in \mathbb{R}^N$ (often with Gumbel-Softmax).
A Top-$K$ strategy (typically $K=1$ or $2$) activates only the selected experts, and the output is aggregated as
$\mathbf{y} = \sum_{i \in \text{TopK}(\mathbf{g}(\mathbf{x}))} g_i(\mathbf{x}) \cdot E_i(\mathbf{x})$,
where $E_i(\cdot)$ denotes expert $i$.
This paradigm is widely adopted in large-scale models, such as Switch Transformer~\cite{2022switchtransformers}, which replaces FFN blocks with MoE for efficiency.
However, MoE suffers from \textit{expert load imbalance}, where a few experts are overused while others remain idle, causing memory overflow, degraded throughput, and poor hardware utilization.
To address \textit{load imbalance} in MoE, three strategies are commonly used:
i) \textit{load-balancing loss}~\cite{2022switchtransformers}, which penalizes uneven token routing;
ii) \textit{expert capacity constraints}~\cite{2022switchtransformers}, which cap tokens per expert with controlled dropping; and
iii) \textit{gate refinement}~\cite{swapmoe,2024Expertpruningandskipping,2024Pre-gatedmoe,2023lina}, which improves routing quality.
Representative designs combine these ideas.
Switch Transformer~\cite{2022switchtransformers} uses balancing loss with capacity buffering, while Pre-gated MoE~\cite{2024Pre-gatedmoe} overlaps expert migration and execution to cut latency.
Beyond balancing, \textit{dynamic sparsity} methods skip low-contributing experts at runtime, as in SwapMoE~\cite{swapmoe} and expert pruning/skipping~\cite{2024Expertpruningandskipping}. Lina~\cite{2023lina} further schedules experts by popularity, enabling more adaptive resource allocation.

\begin{figure}[t]
    \centering
    \begin{minipage}{0.9\linewidth}
        \centering
        \subfloat[Dynamic routing with MoE.]{
    \includegraphics[height=0.22\textwidth]{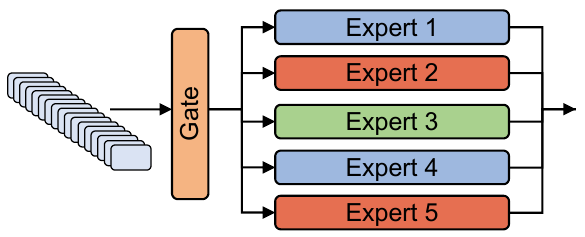}
            \label{fig:Affinity_routing_a}
        }
    \end{minipage}
    \begin{minipage}{0.9\linewidth}
        \centering
        \subfloat[Expert-level load balancing.]{
            \includegraphics[height=0.6\textwidth]{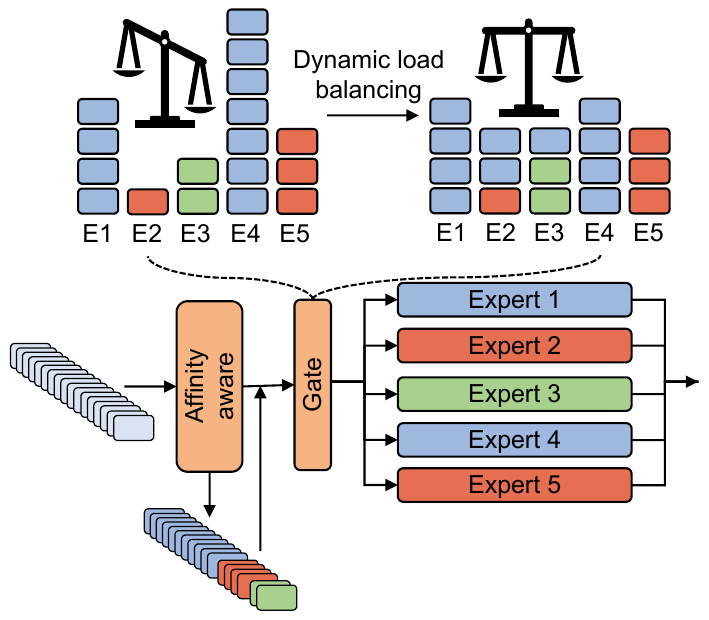}
            \label{fig:Affinity_routing_b}
        }
    \end{minipage}
    \begin{minipage}{0.9\linewidth} 
        \centering
        \subfloat[Device-level load balancing.]{
            \includegraphics[height=0.6\textwidth]{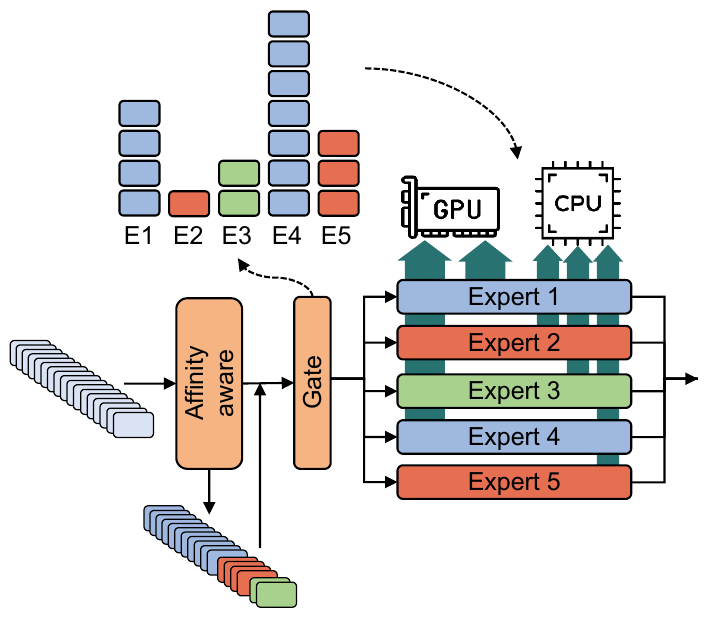}
            \label{fig:Affinity_routing_c}
        }
    \end{minipage}
    \caption{Illustraion of affinity threshold-based dynamic routing.}
    \label{fig:Affinity threshold-based}
    \vspace{-5mm}
\end{figure}

\subsubsection{Affinity Threshold-based Dynamic Routing with MoE}
\label{subsubsec:affinity_threshold_based}
Affinity-based dynamic routing addresses the limitations of gating-based MoE, where fixed gating functions fail to adapt to dynamic input distributions and handcrafted balancing losses conflict with primary objectives, leading to overload, underuse, or instability~\cite{2022switchtransformers}. Instead of static gating, it computes token–expert compatibility scores with learnable bias terms and real-time workload feedback~\cite{2024deepseekv3}, enabling adaptive, workload-aware routing without auxiliary losses (\figref{fig:Affinity threshold-based}). Representative methods include DeepSeek-V3~\cite{2024deepseekv3}, which employs sigmoid-based affinity scoring to improve utilization; Expert-Token Resonance MoE~\cite{2024Expert-Token-Resonance-MoE}, which diversifies specialization via cosine similarity and an orthogonal GrAP layer; APTMoE~\cite{2024aptmoe}, which reduces GPU memory and transfers by offloading sparse experts to CPUs; and Exflow~\cite{2024Exflow}, which co-locates stable cross-layer expert groups to cut communication latency. Collectively, these advances improve utilization, stability, and distributed efficiency for scalable agentic AI.

\begin{figure}[t]
    \centering
\includegraphics[width=0.4\textwidth]{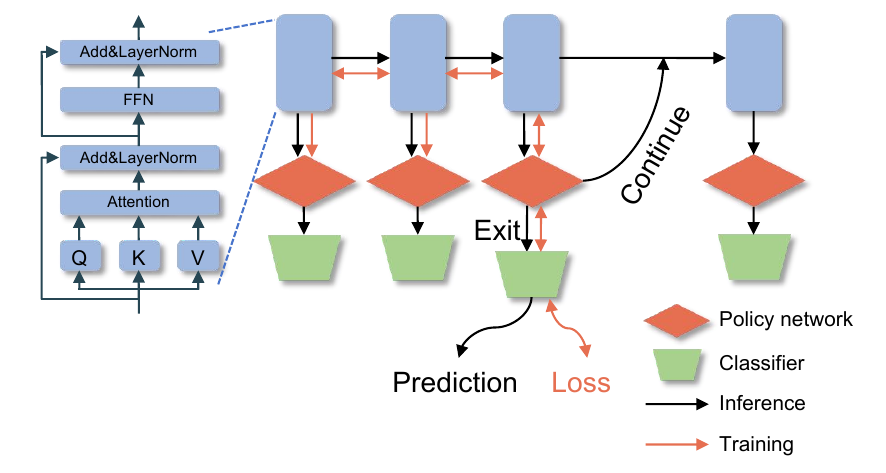}
    \vspace{-2mm}
    \caption{RL-based dynamic routing.}
    \label{fig:RL_routing}
    \vspace{-6mm}
\end{figure}

\subsubsection{Reinforcement Learning-based Dynamic Routing with MoE}
\label{subsubsec:rl_based}
Reinforcement learning (RL)-based dynamic routing formulates expert selection in MoE as a \textit{sequential decision-making task}. 
Unlike gating- or affinity-based approaches with fixed scoring, RL methods employ \textit{policy networks}~\cite{zhou2022mixture} and \textit{reward-driven training}~\cite{2024consistentee} to enable context-aware path selection and global multi-objective optimization. 
By adapting computation to input difficulty, they jointly optimize accuracy–latency trade-offs and mitigate local minima through policy exploration. ConsistentEE~\cite{2024consistentee} exemplifies this paradigm, using RL to optimize early-exit policies (\secref{subsubsec:early_existing}) via a memory layer for difficulty assessment and a difficulty-aware reward balancing prediction quality and latency. Policy gradients with multi-trajectory sampling further alleviate sparse rewards and align training with inference-time decisions (\figref{fig:RL_routing}). Despite challenges in reward design and training stability, RL-based routing shows strong potential for flexible, task-adaptive inference in resource-constrained agentic AI.

\subsubsection{Early Exiting}
\label{subsubsec:early_existing}
Early exiting accelerates inference by inserting adaptive decision points in shallow layers, allowing easy inputs to terminate once confidence is sufficient. This reduces latency and computation in FMs. 
Existing work mainly follows two directions, \ie \textit{confidence-based exit strategies}~\cite{2022adainfer,schuster2022calm,2023FREE} and \textit{system-level integration}~\cite{chen2023ee-llm,del2023skipdecode,2024edge_LLM}.
\textit{First, confidence-based} methods focus on \textit{when to exit}, balancing accuracy and efficiency through confidence measures. For example, AdaInfer~\cite{2022adainfer} uses token-level features (\eg gap, top prob), FREE~\cite{2023FREE} applies a BMM-based estimator with parallel decoding, and CALM~\cite{schuster2022calm} evaluates softmax response, hidden state saturation, and classifier scoring.
\textit{Second, system-level} integration improves compatibility with optimizations like KV cache reuse and batch decoding. EE-LLM~\cite{chen2023ee-llm} overlaps KV computation and token generation via pipelined scheduling, SkipDecode~\cite{del2023skipdecode} enforces monotonic exit depths to minimize recomputation, and Edge-LLM~\cite{2024edge_LLM} introduces confidence-based voting with sensitivity-aware compression and hardware-aligned scheduling.

\subsubsection{Layer skipping}
\label{subsubsec:layer_skipping}
Layer skipping accelerates inference by dynamically bypassing intermediate layers, unlike early exiting which terminates the entire sequence.
It adaptively adjusts computational depth at the token or layer level, offering fine-grained acceleration.
Research falls into three main directions.
\textit{i) Importance-based pruning}: Shortened LLaMA~\cite{kim2024shortenedllama} ranks block importance via Taylor+ and PPL metrics to prune uncritical layers, while LaCo~\cite{yang2024laco} merges adjacent layers to reduce depth without losing structure.
\textit{ii) Token-aware skipping}: SkipLayer~\cite{zeng2023skiplayer} uses binary routing to decide per-token layer execution, and D-LLMs~\cite{jiang2024dllms} combine decision modules with eviction policies to reduce compute and KV cache usage.
\textit{iii) Decoding-oriented skipping}: SkipDecode~\cite{del2023skipdecode} skips shallow layers and reuses deeper computations during generation, while Draft-Verify accelerates speculative decoding by selectively skipping intermediate layers.

\subsection{Dynamic KV Cache Management}\label{subsec:dynamic_kv_cache_management}
Dynamic Key–Value (KV) cache management underpins adaptive Transformer inference by reducing redundant computation in autoregressive decoding, lowering complexity from quadratic $O(n^2)$ to linear $O(n)$.
However, memory grows linearly with sequence length, causing overflow and bandwidth-induced latency, which are bottlenecks for mobile/edge devices and real-time applications.
To sustain responsiveness, recent work pursues adaptive cache management along three fronts.

\begin{table*}[t]
\centering
\caption{Summary of dynamic KV cache management techniques for elastic FM inference.}
\vspace{-2mm}
\tiny
\label{tab:dynamic_kv_cache_management}
\renewcommand{\arraystretch}{1.1}
\setlength{\tabcolsep}{6pt}
\resizebox{\textwidth}{!}{%
\begin{tabular}{|c|c|c|c|c|}
\hline
\multicolumn{2}{|c|}{\textbf{Categories}} & 
\multicolumn{1}{c|}{\textbf{Technique highlight}} & 
\multicolumn{1}{c|}{\textbf{Year}} & 
\multicolumn{1}{c|}{\textbf{Ref}} \\
\hline

\multirow{14}{*}[0ex]{\centering\textbf{\begin{tabular}{c} Dynamic KV cache \\management \\~(\S\ref{subsec:dynamic_kv_cache_management})\end{tabular}}} 

& \multirow{4}{*}[0ex]{\centering\textbf{\begin{tabular}{c} Context-aware KV \\cache optimization \\~(\S\ref{subsubsec:context_aware_kv_cache_optimization})\end{tabular}}} 
& \begin{tabular}[c]{@{}c@{}}Evaluates token importance via cumulative attention in sliding windows for critical tokens with long-term impact.\end{tabular} & 2023 & \cite{2023h2o} \\
\cline{3-5}
& & \begin{tabular}[c]{@{}c@{}}Focuses on sustained temporal influence of historical high-importance tokens for token relevance consistency.\end{tabular} & 2023 & \cite{2023scissorhands} \\
\cline{3-5}
& & \begin{tabular}[c]{@{}c@{}}Enhances cache efficiency via clustering-based token eviction, boosting accuracy and resource utilization.\end{tabular} & 2024 & \cite{2024clusterkv} \\
\cline{3-5}
& & \begin{tabular}[c]{@{}c@{}}Optimizes KV cache via CPU-GPU hierarchy, offloading less-accessed to CPU.\end{tabular} & 2024 & \cite{2024infllm} \\
\cline{2-5}

& \multirow{4}{*}[0ex]{\centering\textbf{\begin{tabular}{c} Attention-aware KV \\cache optimization \\~(\S\ref{subsubsec:attention_aware_kv_cache_optimization})\end{tabular}}} 
& \begin{tabular}[c]{@{}c@{}}Accelerates incremental Transformer decoding by sharing key-value heads with only minor quality loss.\end{tabular} & 2019 & \cite{shazeer2019mqa} \\
\cline{3-5}
& & \begin{tabular}[c]{@{}c@{}}Enables efficient uptraining from MHA checkpoints with $5\%$  
pre-training compute for MHA-quality at MQA-speed.\end{tabular} & 2023 & \cite{ainslie2023gqa} \\
\cline{3-5}
& & \begin{tabular}[c]{@{}c@{}}Projects queries, keys, and values into a low-dimensional latent space for attention computation.\end{tabular} & 2024 & \cite{2024deepseekv3} \\
\cline{3-5}
& & \begin{tabular}[c]{@{}c@{}}Compresses KV into a fixed associative memory and streams up to 1M tokens with local $+$ linear attention.\end{tabular} & 2025 & \cite{munkhdalai2024Infini-Attention} \\
\cline{2-5}

& \multirow{4}{*}[0ex]{\centering\textbf{\begin{tabular}{c} Model-adaptive KV \\cache optimization \\~(\S\ref{subsubsec:structure_aware_kv_cache_optimization})\end{tabular}}} 
& \begin{tabular}[c]{@{}c@{}}Dynamically assigns cache space via average attention scores to optimize layer-specific utilization and accuracy.\end{tabular} & 2025 & \cite{zhou2024dynamickv} \\
\cline{3-5}
& & \begin{tabular}[c]{@{}c@{}}Leverages attention heterogeneity to allocate cache pyramidally, prioritizing more resources for lower layers.\end{tabular} & 2025 & \cite{cai2024pyramidkv} \\
\cline{3-5}
& & \begin{tabular}[c]{@{}c@{}}Guides cache allocation by analyzing the theoretical upper bound of eviction loss for controllable resource management.\end{tabular} & 2025 & \cite{feng2024ada-kv} \\
\cline{3-5}
& & \begin{tabular}[c]{@{}c@{}}Implements attention-head-level cache allocation via dual capability evaluation and dynamic budget pooling.\end{tabular} & 2024 & \cite{fu2024headkv} \\
\hline
\end{tabular}%
}
\vspace{-4mm}
\end{table*}

\begin{figure*}[t]
    \centering
\includegraphics[width=0.7\textwidth]{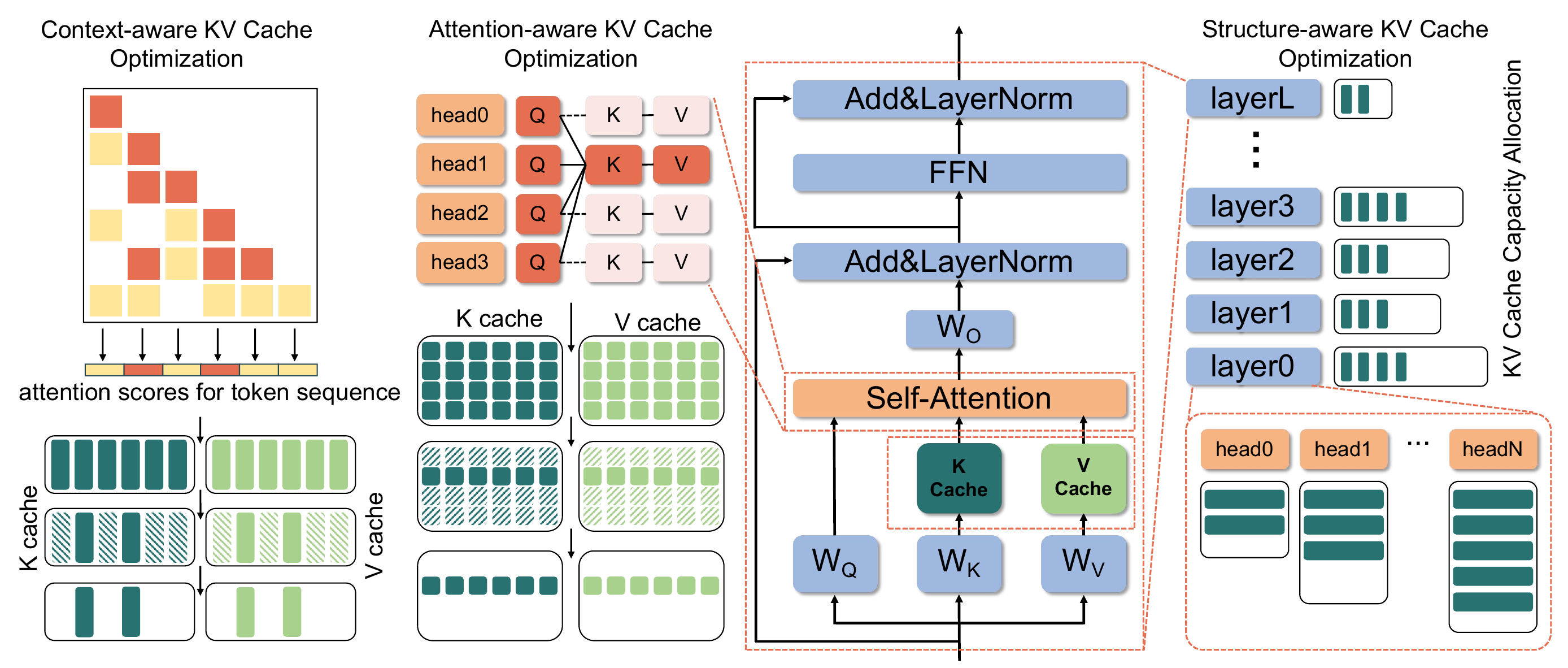}
    \caption{Dynamic KV cache management.}
    \label{fig:Dynamic KV Cache Management}
    \vspace{-4mm}
\end{figure*}

\subsubsection{Context-Aware KV Cache Optimization}
\label{subsubsec:context_aware_kv_cache_optimization} 
Context-aware KV cache optimization dynamically regulates cache usage by prioritizing high-importance tokens and offloading low-value ones. 
Importance scores, usually derived from attention weights, guide three main strategies: \textit{discarding} redundant tokens~\cite{2023h2o,2023scissorhands}, \textit{merging/clustering} similar ones~\cite{2024squeezedattention,2024clusterkv}, and \textit{quantization} for compact storage~\cite{liu2024kivi}. 
Representative designs include H2O~\cite{2023h2o} and Scissorhands~\cite{2023scissorhands}, which assess token relevance over time; PQ-Cache and RetrievalAttention~\cite{2024retrievalattention}, which employ similarity search (MIPS/ANNS) for cache selection; and SqueezedAttention~\cite{2024squeezedattention} and ClusterKV~\cite{2024clusterkv}, which cluster tokens to improve efficiency. 
At the storage level, InfLLM~\cite{2024infllm} hierarchically manages CPU–GPU memory, while Keyformer stabilizes eviction policies. Recent quantization methods (\eg KIVI, KVQuant, QServe, IntactKV) further compress key–value states to optimize memory and bandwidth.

\subsubsection{Attention-Aware KV Cache Optimization}
\label{subsubsec:attention_aware_kv_cache_optimization}
Core strategies include \textit{KV sharing} (MQA~\cite{shazeer2019mqa}, GQA~\cite{ainslie2023gqa}), which compress cache by sharing KV pairs across heads or groups; \textit{latent compression} (MLA~\cite{2024deepseekv3}), which replaces full KV with low-rank latent vectors; and \textit{approximate or hybrid attention} (FLASH~\cite{2022FLASH}, Infini-Attention~\cite{munkhdalai2024Infini-Attention}), which combine gating, linear approximations, or local–global hybrids to balance efficiency and long-range modeling. 
For instance, MQA reduces cache size to $1/n_{\text{head}}$ but risks accuracy loss, while GQA offers a tunable trade-off via group size. MLA achieves compression without major degradation, and FLASH/Infini-Attention improve scalability by approximating or hybridizing attention computation.

\subsubsection{Model-adaptive KV Cache Optimization}
\label{subsubsec:structure_aware_kv_cache_optimization}
Model-adaptive KV cache optimization improves efficiency by tunning cache budgets to \textit{layer}, \textit{head}, and \textit{model-specific} traits such as token/attention distributions, eviction-loss bounds, and retrieval ability. PrefixKV and DynamicKV~\cite{zhou2024dynamickv} adjust allocation via prefix patterns and attention scores, while PyramidKV~\cite{cai2024pyramidkv}, PyramidInfer, and MEDA~\cite{wan2025meda} exploit cross-layer heterogeneity with pyramid- or entropy-based quotas. AdaKV~\cite{feng2024ada-kv} bounds eviction loss for dynamic head budgets, and HeadKV~\cite{fu2024headkv} further prioritizes heads by retrieval and inference value. 
Together, they support elastic cache–compute trade-offs aligned with model structure.


\subsection{Model-Adaptive System Scheduling}
\label{subsec:model_adaptative_system_scheduling}
Model-Adaptive System Scheduling integrates algorithmic design with system-level scheduling to overcome FM inference bottlenecks by maximizing hardware utilization under dynamic workloads while preserving accuracy. 
It operates at two levels, the \textit{front-end}, which optimizes abstract computation graphs through redundancy elimination, intermediate simplification, and parallelism strategies such as data, sequence, pipeline, and expert parallelism~\cite{huang2025hd,yan2025moepicaccelerating}; and the \textit{back-end}, which tunes execution to device capabilities via graph-level (\eg operator fusion, graph rewriting~\cite{jia2019taso}), memory-level (\eg dynamic allocation, swapping), and instruction-level optimizations (\eg loop unrolling, register tiling).

\subsubsection{Adaptive Parallelism Scheduling}
\label{subsubsec:adaptive_parallelism_scheduling}
It dynamically adjusts execution strategies to system states (\eg memory, compute, bandwidth) for efficient FM inference in resource-constrained agentic environments. 
Key paradigms include \textit{data parallelism}, \textit{sequence parallelism}, \textit{pipeline parallelism}, \textit{expert parallelism}, and \textit{heterogeneous processor parallelism}. 
The main challenges are memory efficiency, real-time scheduling, and communication-aware partitioning, with adaptive strategies seeking to optimize throughput and latency while preserving accuracy for mobile/edge deployment (\tabref{tab:adaptive_parallelism}).

\begin{table*}[t]
\centering
\caption{Summary of adaptive parallelism scheduling techniques for efficient FM inference.}
\vspace{-2mm}
\tiny
\label{tab:adaptive_parallelism}
\renewcommand{\arraystretch}{1.1}
\setlength{\tabcolsep}{6pt}
\resizebox{\textwidth}{!}{%
\begin{tabular}{|c|c|c|c|c|}
\hline
\multicolumn{2}{|c|}{\textbf{Categories}} &
\multicolumn{1}{c|}{\textbf{Technique highlight}} &
\multicolumn{1}{c|}{\textbf{Year}} &
\multicolumn{1}{c|}{\textbf{Ref}} \\
\hline

\multirow{24}{*}[0ex]{\centering\textbf{\begin{tabular}{c} Adaptive \\parallelism \\scheduling \\~(\S\ref{subsubsec:adaptive_parallelism_scheduling}) \end{tabular}}}

& \multirow{4}{*}[0ex]{\centering\textbf{\begin{tabular}{c} Data \\parallelism \\~(\S\ref{subsubsec:data_parallelism}) \end{tabular}}}
& \begin{tabular}[c]{@{}c@{}}Boosts LLM inference throughput via OS-style paging (PagedAttention) for on-demand, shareable KV cache blocks.\end{tabular} & 2023 & \cite{kwon2023efficient} \\
\cline{3-5}
& & \begin{tabular}[c]{@{}c@{}}Partitions parameters and KV caches across nodes to reduce memory footprint using full sharding.\end{tabular} & 2025 & \cite{su2025seesaw} \\
\cline{3-5}
& & Employs iteration-level scheduling to enable continuous batching. & 2022 & \cite{yu2022orca} \\
\cline{3-5}
& & Dynamically adjusts data processing granularity across heterogeneous nodes. & 2025 & \cite{huang2025hd} \\
\cline{2-5}

& \multirow{3}{*}[0ex]{\centering\textbf{\begin{tabular}{c} Sequence \\parallelism \\~(\S\ref{subsubsec:sequence_parallelism}) \end{tabular}}}
& \begin{tabular}[c]{@{}c@{}}Speeds up exact attention via tiling/recomputation, avoiding attention matrix storage and cutting GPU accesses.\end{tabular} & 2024 & \cite{2022flashattention} \\
\cline{3-5}
& & Partitions sequences of up to one million tokens to achieve significant speedups. & 2022 & \cite{aminabadi2022deepspeedinference} \\
\cline{3-5}
& & \begin{tabular}[c]{@{}c@{}}Sequence partitioning, local/global attention optimization reduce communication/memory overhead in long sequences.\end{tabular} & 2024 & \cite{sun2024burstattention} \\
\cline{2-5}

& \multirow{4}{*}[0ex]{\centering\textbf{\begin{tabular}{c} Pipeline \\parallelism \\~(\S\ref{subsubsec:pipeline_parallelism}) \end{tabular}}}
& \begin{tabular}[c]{@{}c@{}}Boosts LLM speed via async pipelined speculation/early cancellation for low-acceptance/low-bandwidth scenarios.\end{tabular} & 2024 & \cite{butler2024pipeinfer} \\
\cline{3-5}
& & Applies dynamic micro-batching for efficient multi-task inference. & 2024 & \cite{TTFT} \\
\cline{3-5}
& & Refines execution granularity through task scheduling to minimize pipeline stalls. & 2023 & \cite{miao2023specinfer} \\
\cline{3-5}
& & \begin{tabular}[c]{@{}c@{}}Refines execution granularity and manages memory via swapping to minimize stalls.\end{tabular} & 2025 & \cite{du2025flexinfer} \\
\cline{2-5}

& \multirow{3}{*}[0ex]{\centering\textbf{\begin{tabular}{c} Expert \\parallelism \\~(\S\ref{subsubsec:expert_parallelism}) \end{tabular}}}
 & Improves cache hit rates through adaptive partitioning and VRAM budgeting. & 2025 & \cite{yan2025moepicaccelerating} \\
\cline{3-5}
& & Reformulates experts into block-sparse General Matrix Multiplications (GEMMs). & 2025 & \cite{cao2025moe-lightning} \\
\cline{3-5}
& & Implements dynamic load balancing using sparse activation of experts. & 2022 & \cite{du2022glammoe} \\
\cline{2-5}

& \multirow{4}{*}[0ex]{\centering\textbf{\begin{tabular}{c} Heterogeneous \\processor \\parallelism \\~(\S\ref{subsubsec:heterogeneous_processor_parallelism}) \end{tabular}}}
& \begin{tabular}[c]{@{}c@{}}Leverages LLM power-law activation, GPU-CPU hybrid and sparse operators for fast, accurate consumer GPU inference.\end{tabular} & 2024 & \cite{song2024powerinfer} \\
\cline{3-5}
& & \begin{tabular}[c]{@{}c@{}}DistServe disaggregates prefill/decoding, cuts interference, optimizes resources to boost GPU goodput.\end{tabular} & 2024 & \cite{TTFT} \\
\cline{3-5}
& & \begin{tabular}[c]{@{}c@{}}HD-MoE optimizes MoE LLMs on 3D NMP via offline hybrid parallel mapping and online dynamic scheduling\end{tabular} & 2025 & \cite{huang2025hd} \\
\cline{3-5}
& & \begin{tabular}[c]{@{}c@{}}Addresses LLM inference I/O bottlenecks on resource-constrained devices via CPU-GPU hetero-parallelism and async overlap\end{tabular} & 2024 & \cite{zhao2024hetegen} \\
\hline

\end{tabular}%
}
\vspace{-2mm}
\end{table*}

\begin{figure}[t]
    \centering
    \includegraphics[width=0.35\textwidth]{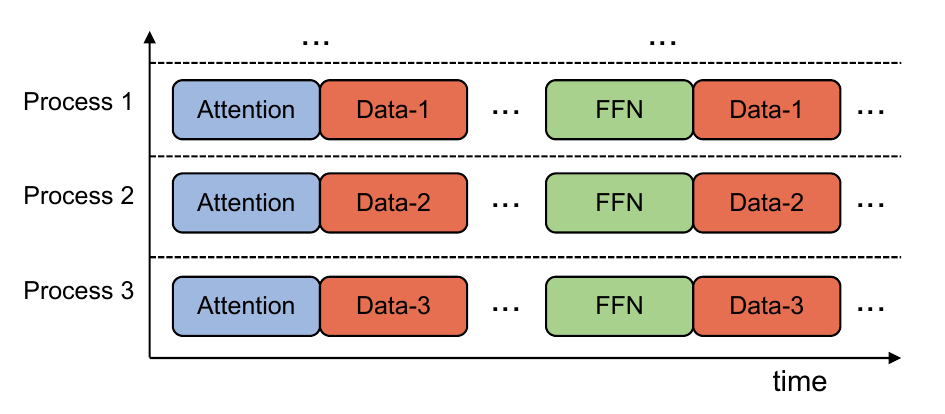}
    \caption{Illustration of data parallelism. Each process performs the same operations on different data subsets simultaneously.}
    \vspace{-5mm}
    \label{fig:data parallelism_v1}
\end{figure}

\textbf{\textit{a. Data parallelism.}}
\label{subsubsec:data_parallelism}
It distributes data batches across devices to overcome memory limits and communication overhead (\figref{fig:data parallelism_v1}). 
Approaches include \textit{full replication} (\eg vLLM~\cite{kwon2023efficient}, TensorRT-LLM), which is fast but memory-hungry; \textit{full sharding} (\eg Seesaw~\cite{su2025seesaw}), which reduces footprint but increases communication; and \textit{hybrid strategies} that balance both. Systems like Orca~\cite{yu2022orca} apply iteration-level scheduling, while HD-MoE~\cite{huang2025hd} adapts granularity across heterogeneous nodes. Recent advances explore \textit{dynamic re-sharding} to cut overhead and sustain efficiency under non-uniform bandwidth.

\begin{figure}[t]
    \centering
    \includegraphics[width=0.49\textwidth]{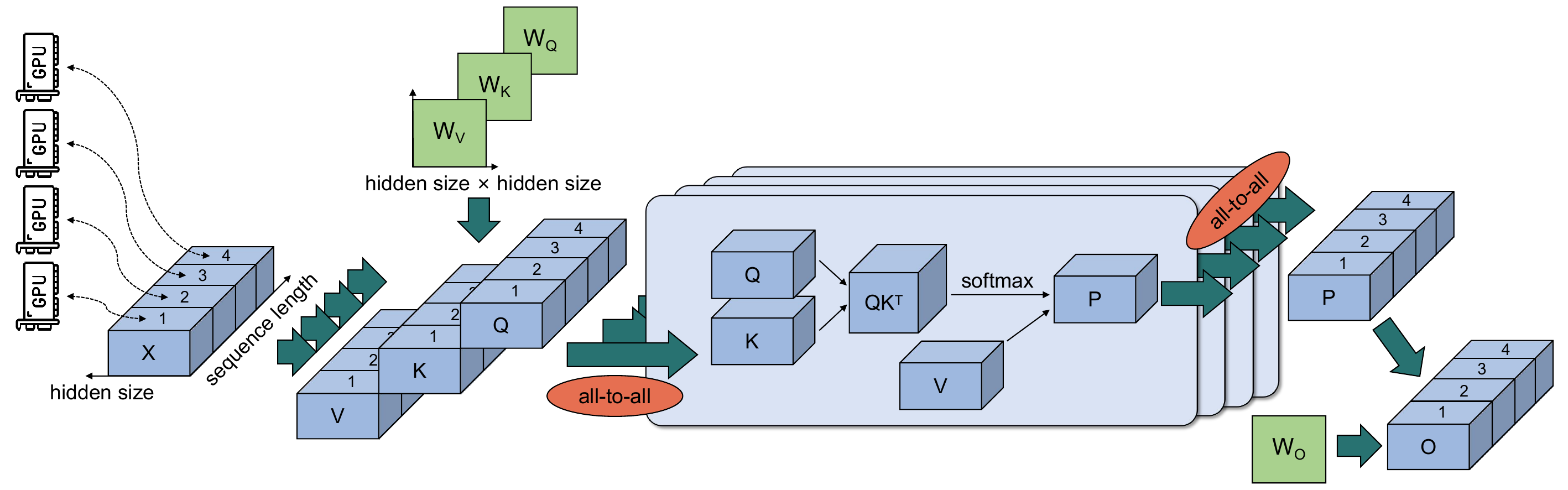}
    \caption{Illustration of sequence parallelism.}
    \label{fig:sequence parallelism}
    \vspace{-4mm}
\end{figure}

\textbf{\textit{b. Sequence parallelism.}}
\label{subsubsec:sequence_parallelism}
It accelerates long-sequence inference by splitting inputs across devices, reducing compute and memory pressure (\figref{fig:sequence parallelism}). 
Techniques span \textit{distributed attention}~\cite{2022flashattention,2023flashattention2,2024flashattention3}, \textit{sequence partitioning}~\cite{aminabadi2022deepspeedinference,su2025seesaw}, \textit{token-level balancing}~\cite{huang2025hd}, and \textit{dynamic reshaping}~\cite{sun2024burstattention}. 
Representative systems include DeepSpeed-Inference (2.5$\times$ speedup on million-token inputs), Sarathi (10$\times$ throughput via reshaping), StreamLLM (latency reduction via streaming), and RingAttention (near-unlimited context without memory overhead).

\textbf{\textit{c. Pipeline parallelism.}}
\label{subsubsec:pipeline_parallelism}
Pipeline parallelism accelerates FM inference by partitioning models into sequential stages across devices, improving memory utilization and efficiency for large models and long sequences. In dynamic or heterogeneous settings, adaptive scheduling addresses pipeline bubbles and latency bottlenecks. 
Strategies include \textit{adaptive partitioning} for balanced workload and memory~\cite{butler2024pipeinfer,TTFT}, \textit{task scheduling} to reduce stalls~\cite{miao2023specinfer}, and \textit{memory management} via swapping~\cite{song2024powerinfer,du2025flexinfer}. 
Representative systems such as PipeInfer~\cite{butler2024pipeinfer}, DistServe~\cite{TTFT}, SpecInfer~\cite{miao2023specinfer}, PowerInfer~\cite{song2024powerinfer}, and FlexInfer~\cite{du2025flexinfer}, demonstrate these optimizations through dynamic recomputation, micro-batching, and fine-grained scheduling.


\begin{figure}[t]
    \centering
    \includegraphics[width=0.49\textwidth]{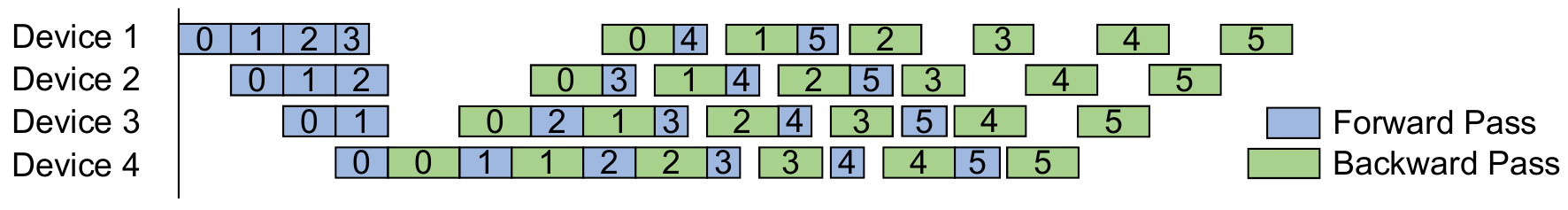}
    \caption{Illustration of the compute efficiency challenges introduced by pipeline bubbles under 1F1B.}
    \label{fig:pipe parallelism}
    \vspace{-6mm}
\end{figure}

\textbf{\textit{d. Expert parallelism.}}
\label{subsubsec:expert_parallelism}
It scales Mixture-of-Experts (MoE) models by activating only a subset of experts per input, reducing computation but introducing challenges of all-to-all communication, load imbalance, and dynamic scheduling. 
Experts are distributed across devices, requiring efficient cross-node coordination. Recent work advances along four fronts: \textit{communication and scheduling} (Occult, SpeculativeMoE, HD-MoE~\cite{huang2025hd}), \textit{flexible execution} with adaptive partitioning and hardware-agnostic communication (MoEpic~\cite{yan2025moepicaccelerating}), \textit{block-sparse and pipelined execution} to overlap compute and communication (MoE-Lightning~\cite{cao2025moe-lightning}, MoE-Lens), and \textit{dynamic load balancing} via sparse activation and cross-device routing (GLaM~\cite{du2022glammoe}, Uni-MoE~\cite{li2025uni-moe}).


\textbf{\textit{e. Heterogeneous processor parallelism}.}
\label{subsubsec:heterogeneous_processor_parallelism}
Embedded devices typically integrate diverse processors (\eg CPUs, GPUs, NPUs), creating opportunities to accelerate FM inference but also challenges in balancing loads and minimizing communication overhead. Recent systems (PowerInfer~\cite{song2024powerinfer}, Splitwise~\cite{patel2024splitwise}, DistServe~\cite{TTFT}, HD-MoE~\cite{huang2025hd}, FlexGen~\cite{sheng2023flexgen}, HeteGen~\cite{zhao2024hetegen}) dynamically map tasks to the most suitable processor. 
For example, PowerInfer exploits GPUs for large-scale sparse activations while offloading sequential I/O-heavy tasks to CPUs, boosting throughput and efficiency. Adaptive schedulers refine this mapping in real time based on processor capabilities and workload conditions, ensuring sustained performance and energy efficiency.

\subsubsection{Computation Graph-level Optimization}
\label{subsubsec:computation_graph_level_optimization}
Computation graph-level optimization restructures the overall FM graph—beyond operator-level tuning—to reduce redundant computation, memory access, and scheduling overhead. Two main approaches exist: \textit{operator fusion}, which merges adjacent ops into composite operators to cut runtime and improve locality (\eg tensor fusion~\cite{2022flashattention,2023flashattention2,2024flashattention3}, kernel fusion~\cite{zhou2024survey}; FlashAttention-2~\cite{2023flashattention2}, vLLM~\cite{kwon2023efficient}); and \textit{dynamic graph optimization}, which adapts graphs at runtime via on-demand construction, lazy evaluation, or rewriting (\eg DyNet~\cite{neubig2017dynet}, TensorFlow Fold~\cite{looks2017deeplearningdynamiccomputation}, Dali~\cite{raiman2018dali}). 
These methods jointly reduce compute/memory costs while preserving flexibility for deployment in constrained agentic systems.

\begin{table*}[t]
\centering
\caption{Summary of computation graph-level optimization and load balancing scheduling for elastic FM inference.}
\vspace{-2mm}
\tiny
\label{tab:graph_load_balance}
\renewcommand{\arraystretch}{1.1}
\setlength{\tabcolsep}{6pt}
\resizebox{\textwidth}{!}{%
\begin{tabular}{|c|c|c|c|c|}
\hline
\multicolumn{2}{|c|}{\textbf{Categories}} & 
\multicolumn{1}{c|}{\textbf{Technique highlight}} & 
\multicolumn{1}{c|}{\textbf{Year}} & 
\multicolumn{1}{c|}{\textbf{Ref}} \\
\hline

\multirow{7}{*}[0ex]{\centering\textbf{\begin{tabular}{c} Computation graph-level \\optimization \\~(\S\ref{subsubsec:computation_graph_level_optimization}) \end{tabular}}} 

& \multirow{3}{*}[0ex]{\centering\textbf{\begin{tabular}{c} Operator fusion \\~(\S\ref{subsubsec:computation_graph_level_optimization}) \end{tabular}}} 
& \begin{tabular}[c]{@{}c@{}}Boosts attention efficiency via optimizing non-matmul FLOPs, sequence-parallelism, and intra-block warp partitioning for GPUs.\end{tabular} & 2023 & \cite{2023flashattention2} \\
\cline{3-5}
& & \begin{tabular}[c]{@{}c@{}}Surveys LLM efficient inference via data/model/system-level optimizations, analyzes bottlenecks, with experiments.\end{tabular} & 2023 & \cite{kwon2023efficient} \\
\cline{3-5}
& & \begin{tabular}[c]{@{}c@{}}Enables efficient LLM serving via KV cache paging (non-contiguous), cuts fragmentation, boosts throughput 2--4$\times$.\end{tabular} & 2023 & \cite{kwon2023efficient} \\
\cline{2-5}

& \multirow{3}{*}[0ex]{\centering\textbf{\begin{tabular}{c} Dynamic computation graph optimization \\~(\S\ref{subsubsec:computation_graph_level_optimization}) \end{tabular}}} 
& \begin{tabular}[c]{@{}c@{}}Dynamic computation graph declaration, optimizes construction overhead, supports dynamic structures, faster than peers.\end{tabular} & 2017 & \cite{neubig2017dynet} \\
\cline{3-5}
& & \begin{tabular}[c]{@{}c@{}}Enables batched dynamic graph learning via dynamic batching, emulating dynamic graphs with static ones.\end{tabular} & 2017 & \cite{looks2017deeplearningdynamiccomputation} \\
\cline{3-5}
& & \begin{tabular}[c]{@{}c@{}}Applies lazy compilation to dynamic computation graphs to boost ML system efficiency.\end{tabular} & 2018 & \cite{raiman2018dali} \\
\hline

\multirow{5}{*}[0ex]{\centering\textbf{\begin{tabular}{c} Load balancing \\scheduling \\~(\S\ref{subsubsec:load_balancing_scheduling}) \end{tabular}}} 

& \multirow{2}{*}[0ex]{\centering\textbf{\begin{tabular}{c} Cross-device load balancing scheduling \\~(\S\ref{subsubsec:load_balancing_scheduling}) \end{tabular}}} 
& \begin{tabular}[c]{@{}c@{}}Extends MoE to NLG, uses PR-MoE/MoS to shrink model, optimizes inference for speed/cost.\end{tabular} & 2022 & \cite{rajbhandari2022deepspeedmoe} \\
\cline{3-5}
& & \begin{tabular}[c]{@{}c@{}}Accelerates distributed MoE training and inference via targeted optimizations for efficiency.\end{tabular} & 2023 & \cite{kwon2023efficient} \\
\cline{2-5}

& \multirow{3}{*}[0ex]{\centering\textbf{\begin{tabular}{c} Heterogeneous chip mapping \\~(\S\ref{subsubsec:load_balancing_scheduling}) \end{tabular}}} 
& \begin{tabular}[c]{@{}c@{}}Surveys CPU-GPU heterogeneous computing techniques across layers, covers systems/suites to boost performance/efficiency.\end{tabular} & 2017 & \cite{Heterogeneous10.1145/2788396} \\
\cline{3-5}
& & \begin{tabular}[c]{@{}c@{}}Enables scalable, memory-efficient DNNs via virtualization techniques for memory management.\end{tabular} & 2016 & \cite{rhu2016vdnn} \\
\cline{3-5}
& & \begin{tabular}[c]{@{}c@{}}Uses CGOPipe and HRM to achieve high-throughput MoE inference on memory-constrained GPUs, outperforming existing systems.\end{tabular} & 2025 & \cite{cao2025moe-lightning} \\
\hline

\end{tabular}%
}
\vspace{-3mm}
\end{table*}

\subsubsection{Load Balance Scheduling}
\label{subsubsec:load_balancing_scheduling}
Efficient computation-to-device mapping is vital for agentic systems, as it directly impacts bandwidth, compute efficiency, and communication overhead. Existing work falls into two lines: \textit{cross-device scheduling}, which mitigates skew and bottlenecks in distributed agent/edge environments~\cite{HammingMesh10.1145/3623490,davies2025efficientllminferencebandwidth,yun2025newllmbottlenecksystems} through methods such as tensor slicing with expert parallelism (DeepSpeed-MoE~\cite{rajbhandari2022deepspeedmoe}) or expert-popularity prediction for dynamic scheduling (Lina~\cite{2023lina}); and \textit{heterogeneous chip mapping}, which balances workloads across CPUs, GPUs, NPUs, and DSPs with divergent compute and memory capacities~\cite{Heterogeneous10.1145/2788396,rhu2016vdnn}, exemplified by MoE-LightNING~\cite{cao2025moe-lightning} via CGOPIPE overlapping CPU/GPU compute and I/O to boost utilization.

\subsubsection{Memory-level Optimization}
\label{subsubsec:memory_allocation_and_scheduling}
Memory management is critical for FM performance in resource-limited agentic environments. To sustain responsiveness and scalability, while supporting efficient retraining, key strategies include \textit{memory recomputation}~\cite{lee2024infinigen,jiang2024kvpr,xu2024pie,zhao2024llmpq_poster}, \textit{partitioning optimization}~\cite{su2025seesaw,stojkovic2025dynamollm,chen2024hardware,davies2025efficientllminferencebandwidth}, and \textit{diverted offloading}~\cite{jiang2024neo,luo2025headinfer,jang2025inf}, which collectively reduce memory usage, balance compute–memory trade-offs, and adapt to runtime constraints (\tabref{tab:memory_optimization}). 

\begin{table*}[t]
\centering
\caption{Summary of memory-level optimization techniques for elastic FM inference.}
\vspace{-2mm}
\tiny
\label{tab:memory_optimization}
\renewcommand{\arraystretch}{1.05}
\setlength{\tabcolsep}{6pt}
\resizebox{\textwidth}{!}{%
\begin{tabular}{|c|c|c|c|c|}
\hline
\multicolumn{2}{|c|}{\textbf{Categories}} &
\multicolumn{1}{c|}{\textbf{Technique highlight}} &
\multicolumn{1}{c|}{\textbf{Year}} &
\multicolumn{1}{c|}{\textbf{Ref}} \\
\hline

\multirow{15}{*}[6ex]{\centering\textbf{\begin{tabular}{c} Memory-level \\optimization \\~(\S\ref{subsubsec:memory_allocation_and_scheduling}) \end{tabular}}}

& \multirow{4}{*}[0ex]{\centering\textbf{\begin{tabular}{c} Memory \\re-computation \\~(\S\ref{subsubsec:memory_re_computation}) \end{tabular}}}
& \begin{tabular}[c]{@{}c@{}}Speculates attention patterns to prefetch critical KV cache, cutting CPU-GPU transfer overhead.\end{tabular} & 2024 & \cite{lee2024infinigen} \\
\cline{3-5}
& & \begin{tabular}[c]{@{}c@{}}I/O-aware partial KV cache recomputation enhances LLM inference efficiency.\end{tabular} & 2024 & \cite{jiang2024kvpr} \\
\cline{3-5}
& & \begin{tabular}[c]{@{}c@{}}Leverages CPU memory via transparent swapping and adaptive expansion for efficient LLM inference.\end{tabular} & 2023 & \cite{xu2024pie} \\
\cline{3-5}
& & \begin{tabular}[c]{@{}c@{}}Uses phase-aware partition and adaptive quantization for efficient LLM serving on heterogeneous clusters.\end{tabular} & 2024 & \cite{zhao2024llmpq_poster} \\
\cline{2-5}

& \multirow{4}{*}[0ex]{\centering\textbf{\begin{tabular}{c} Memory partitioning \\optimization \\~(\S\ref{subsubsec:memory_partitioning_optimization}) \end{tabular}}}
& \begin{tabular}[c]{@{}c@{}}Dynamically adjusts parallelism across prefill/decode, with KV buffering/scheduling cutting overhead.\end{tabular} & 2025 & \cite{su2025seesaw} \\
\cline{3-5}
& & \begin{tabular}[c]{@{}c@{}}Dynamically tunes instances, parallelism, frequency via hierarchical control for energy efficiency under SLOs.\end{tabular} & 2025 & \cite{stojkovic2025dynamollm} \\
\cline{3-5}
& & \begin{tabular}[c]{@{}c@{}}Uses trained prompt tokens for parallel prediction, with hardware-aware sparse tree for efficiency.\end{tabular} & 2024 & \cite{chen2024hardware} \\
\cline{3-5}
& & \begin{tabular}[c]{@{}c@{}}Conducts LLM inference limit study, focusing on bandwidth, sync, capacity via hardware-agnostic model.\end{tabular} & 2025 & \cite{davies2025efficientllminferencebandwidth} \\
\cline{2-5}

& \multirow{3}{*}[0ex]{\centering\textbf{\begin{tabular}{c} Memory diverted \\offloading \\~(\S\ref{subsubsec:memory_diverted_offloading}) \end{tabular}}}
& \begin{tabular}[c]{@{}c@{}}Offloads attention compute and KV cache to CPU via asymmetric pipelining, boosting throughput.\end{tabular} & 2024 & \cite{jiang2024neo} \\
\cline{3-5}
& & \begin{tabular}[c]{@{}c@{}}Offloads KV cache to CPU via head-wise strategy with optimizations, cutting GPU memory.\end{tabular} & 2025 & \cite{luo2025headinfer} \\
\cline{3-5}
& & \begin{tabular}[c]{@{}c@{}}Employs CSDs for near-storage attention computation, with delayed writeback/X-cache cutting I/O to boost LLM throughput.\end{tabular} & 2025 & \cite{jang2025inf} \\
\hline

\end{tabular}%
}
\vspace{-2mm}
\end{table*}


\textbf{\textit{a. Memory re-computation.}}
\label{subsubsec:memory_re_computation}
It reduces peak usage by discarding and regenerating intermediate tensors, trading compute for memory. 
Strategies include \textit{tensor rematerialization}~\cite{lee2024infinigen}, \textit{selective recomputation}~\cite{jiang2024kvpr}, and \textit{adaptive partitioning}~\cite{xu2024pie,zhao2024llmpq_poster}. InfiniGen~\cite{lee2024infinigen} rematerializes KV tensors with \textit{adaptive eviction}; KVPR~\cite{jiang2024kvpr} performs \textit{I/O-aware} partial regeneration; Pie~\cite{xu2024pie} mitigates CPU–GPU fragmentation via swapping; LLM-PQ~\cite{zhao2024llmpq_poster} integrates \textit{phase-aware} quantization with partitioned allocation; 
HybridCache combines checkpointing and hybrid caching for long-context case.

\textbf{\textit{b. Memory partitioning optimization.}}
\label{subsubsec:memory_partitioning_optimization}
This line distributes model states across hardware to reduce peak memory and communication bottlenecks. 
Techniques include \textit{dynamic sharding}~\cite{su2025seesaw,stojkovic2025dynamollm}, \textit{deduplication}~\cite{chen2024hardware}, and \textit{hierarchical synchronization}~\cite{davies2025efficientllminferencebandwidth}. Seesaw~\cite{su2025seesaw} and DynamoLLM~\cite{stojkovic2025dynamollm} enable re-sharding for linear scaling; KV compression and \textit{hardware-aware} decoding~\cite{chen2024hardware} cut redundancy; synchronization frameworks~\cite{davies2025efficientllminferencebandwidth} reduce latency and balance loads.

\textbf{\textit{c. Memory diverted offloading}}.
\label{subsubsec:memory_diverted_offloading}
This strategy shifts data or computation from GPUs to CPUs, NVMe, or near-storage accelerators to balance pressure, bandwidth, and latency. Approaches include \textit{dynamic partitioning}~\cite{jiang2024neo}, \textit{real-time reallocation}~\cite{luo2025headinfer}, and \textit{near-storage processing}~\cite{jang2025inf}. Neo~\cite{jiang2024neo} redistributes tensors across CPU–GPU to cut transfers; HeadInfer~\cite{luo2025headinfer} integrates swapping with scheduling; Aqua and INF$^{2}$~\cite{jang2025inf} exploit near-storage computing to minimize data movement.
\section{Test-time FM Adaptation in Agentic AI Systems}
\label{sec:adaptation}

\begin{figure}[t]
    \centering
    \includegraphics[width=0.48\textwidth]{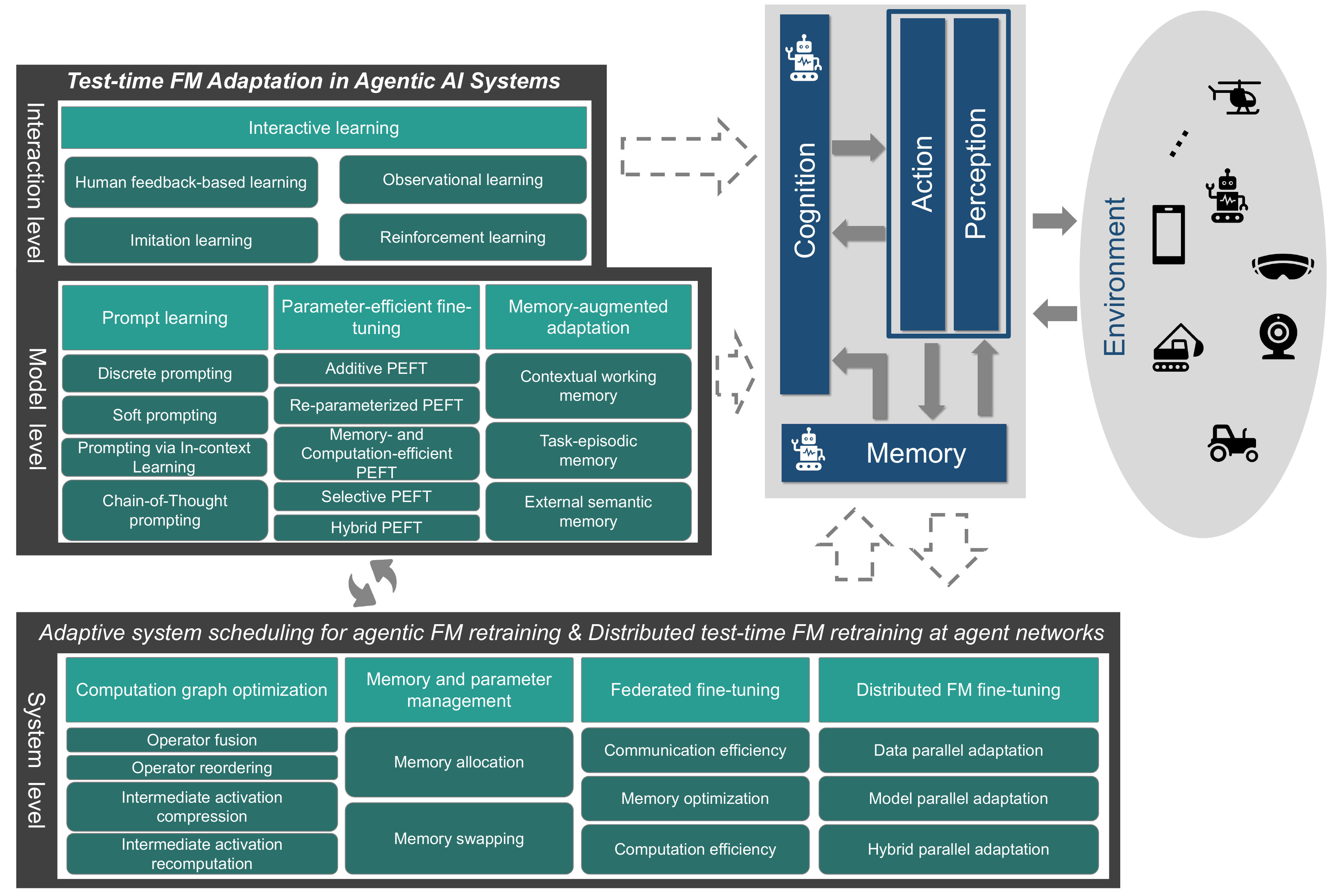}
    \caption{Overview of test-time FM adaptation techniques.}
    \vspace{-5mm}
    \label{fig:test_time_adaptation}
\end{figure}

In dynamic open-world environments, agentic AI systems on platforms such as autonomous vehicles, drones, and service robots must continually adapt their FMs to evolving conditions (\eg traffic, lighting, user intent, novel stimuli). This requires \textit{test-time adaptation}, \ie updating models during inference without full retraining and often without labeled data, to remain robust under distribution shifts, long-horizon tasks, and partial observability.
Two tracks enable adaptive and efficient FM adaptation, \ie \textit{algorithmic strategies} and \textit{system-level techniques}. 
Beyond conventional gradient-based updates, FMs can also refine knowledge through \textit{memory}, \textit{external integration}, and \textit{prompt-driven control}. 
Algorithmic methods include prompt learning, parameter-efficient fine-tuning (PEFT), memory-augmented adaptation, and interactive learning, all designed to minimize parameters, samples, and overhead for rapid adaptation. 
In parallel, \textit{system-level} techniques enhance adaptation under hardware limits via memory management, execution scheduling, and distributed adaptation.

\begin{table*}[t]
\centering
\caption{Summary of prompt tuning techniques for test-time FM adaptation in agentic AI systems.}
\vspace{-2mm}
\tiny
\label{tab:prompt_learning}
\renewcommand{\arraystretch}{1.05}
\setlength{\tabcolsep}{6pt}
\resizebox{\textwidth}{!}{%
\begin{tabular}{|c|c|c|c|c|} 
\hline
\multicolumn{2}{|c|}{\textbf{Categories}} & 
\multicolumn{1}{c|}{\textbf{Technique highlight for improving}} & 
\multicolumn{1}{c|}{\textbf{Year}} & 
\multicolumn{1}{c|}{\textbf{Ref}} \\
\hline

\multirow{12}{*}[\dimexpr-0ex\relax]{\centering\textbf{\begin{tabular}{c} Prompt tuning \\~(\S\ref{sec:prompt_learning})\end{tabular}}} 
& \multirow{3}{*}[\dimexpr-0ex\relax]{\centering\textbf{\begin{tabular}{c} Discrete prompting\\~(\S\ref{sec:discrete_prompting})\end{tabular}}} 
& \begin{tabular}[c]{@{}c@{}}Shared trigger tokens, top-k candidates, optimized prompt construction.\end{tabular}  & 2020 & \cite{2020autoprompt} \\
\cline{3-5}
& & \begin{tabular}[c]{@{}c@{}}Maintain continuous embeddings, project to nearest vocabulary tokens.
\end{tabular} & 2023 & \cite{2023PEZ} \\
\cline{3-5}
& & \begin{tabular}[c]{@{}c@{}}Reinforcement learning for prompt optimization, policy network, z-score rewards.\end{tabular} & 2022 & \cite{2022rlprompt} \\
\cline{2-5}

& \multirow{2}{*}[\dimexpr-0ex\relax]{\centering\textbf{\begin{tabular}{c}Soft prompting\\~(\S\ref{sec:soft_prompting})\end{tabular}}} 
& \begin{tabular}[c]{@{}c@{}}Collaborative soft prompt training, learnable embeddings, aggregated updates. \end{tabular}& 2023 & \cite{2023promptfl} \\
\cline{3-5}
& & \begin{tabular}[c]{@{}c@{}}Global/domain prompts, optimization, momentum aggregation, prompt similarity.\end{tabular} & 2024 & \cite{2024diprompt} \\
\cline{2-5}

& \multirow{3}{*}[\dimexpr-0ex\relax]{\centering\textbf{\begin{tabular}{c}Prompting via \\ in-context learning\\~(\S\ref{sec:in_context_learning})\end{tabular}}} 
& \begin{tabular}[c]{@{}c@{}}Task decomposition, sub-task examples, self-correction, human feedback.\end{tabular} & 2024 & \cite{2024mobilegpt} \\
\cline{3-5}
& & \begin{tabular}[c]{@{}c@{}}Example selection with conditional DPP, capture relevance and diversity.\end{tabular} & 2023 & \cite{2023ceil} \\
\cline{3-5}
& & \begin{tabular}[c]{@{}c@{}}View hierarchies to structured text, chain-of-thought prompting.\end{tabular} & 2023 & \cite{2023in_context_wang} \\
\cline{2-5}

& \multirow{4}{*}[\dimexpr-0ex\relax]{\centering\textbf{\begin{tabular}{c}Chain-of-Thought (CoT) \\
prompting\\~(\S\ref{sec:Chain_of_Thought})\end{tabular}}} 
& \begin{tabular}[c]{@{}c@{}}Five-stage CoT, expand concepts, continuation and revision modules, lightweight adapters.\end{tabular} & 2024 & \cite{2024promptcot} \\
\cline{3-5}
& & \begin{tabular}[c]{@{}c@{}}Attention saliency for adaptive CoT prompt selection, zero-shot reasoning.\end{tabular} & 2024 & \cite{2024instance_cot} \\
\cline{3-5}
& & \begin{tabular}[c]{@{}c@{}}Integrate language, perception, control in multimodal Transformer, four-stage reasoning.\end{tabular} & 2024 & \cite{2024robotic_cot} \\
\cline{3-5}
& & \begin{tabular}[c]{@{}c@{}}Multi-hop rationales, commonsense relations, dual-stage alignment.\end{tabular} & 2023 & \cite{2023DOCTOR} \\
\hline
\end{tabular}%
}
\vspace{-5mm}
\end{table*}
\subsection{Prompt Tuning}
\begin{figure}[t]
    \centering
  \subfloat[Discrete prompting.]{
    \includegraphics[height=0.17\textwidth]{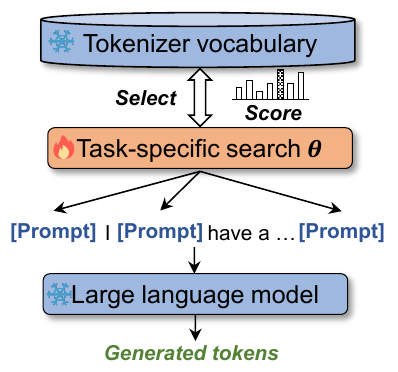}
    \label{fig:discrete_prompting}
    }
  \hspace{0.01\textwidth}
  \subfloat[Soft prompting.]{
    \includegraphics[height=0.17\textwidth]{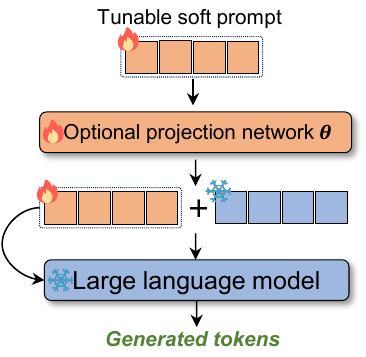}
    \label{fig:soft_prompting}
    }
  \\
  \subfloat[In-context learning.]{
    \includegraphics[height=0.17\textwidth]{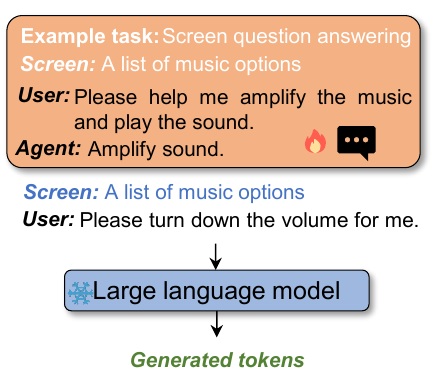}
    \label{fig:in_context_learning}
    }
  \hspace{0.01\textwidth}
  \subfloat[Chain-of-thought.]{
    \includegraphics[height=0.17\textwidth]{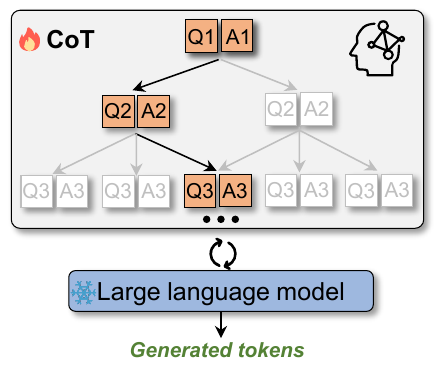}
    \label{fig:cot}
    }
    \caption{Illustration of prompt tuning.}
    \vspace{-6mm}
    \label{fig:prompt_learning}
\end{figure}
\label{sec:prompt_learning}

Prompt learning enables efficient test-time adaptation of FMs by steering behavior through \textit{input sequence modification} \textit{without parameter updates}, making it well-suited for resource-constrained AI agents. As illustrated in \figref{fig:prompt_learning} and \tabref{tab:prompt_learning}, it spans four strategies—\textit{discrete prompting}, \textit{soft prompting}, \textit{in-context learning}, and \textit{chain-of-thought prompting}, each balancing control granularity, model accessibility, and compute cost.

\subsubsection{Discrete Prompting}
\label{sec:discrete_prompting}
Discrete prompting inserts optimized natural-language tokens into inputs to steer frozen FMs without parameter updates (\figref{fig:discrete_prompting}), offering black-box compatibility and negligible overhead. Methods include \textit{gradient-based} approaches, such as AutoPrompt~\cite{2020autoprompt} (gradient-guided token search) and PEZ~\cite{2023PEZ} (embedding-space optimization projected to tokens), as well as \textit{RL-based} methods like RLPrompt~\cite{2022rlprompt}, which frames prompting as reinforcement learning with reward shaping and input-conditioned generation for stronger few-shot generalization.

\subsubsection{Soft Prompting}
\label{sec:soft_prompting}
Soft prompting prepends \textit{trainable continuous embeddings} to inputs, offering finer control than discrete tokens while keeping FM weights frozen (\figref{fig:soft_prompting}). 
PROMPTFL~\cite{2023promptfl} applies this to federated CLIP via learnable embeddings prepended to text tokens. DiPrompT~\cite{2024diprompt} extends to federated domain generalization with disentangled prompts. 
A global prompt for invariant knowledge, local prompts optimized via prototypical guidance and momentum aggregation, and query prompts for label-free domain inference, ensembled at inference for cross-domain generalization.

\subsubsection{Prompting via In-context Learning (ICL)}
\label{sec:in_context_learning}
ICL adapts FMs \textit{parameter-free} by prepending task-specific exemplars, conditioning inference on recent context (\figref{fig:in_context_learning}). This is especially valuable for agents with limited resources and strict latency, enabling dynamic adjustment from signals such as user interaction or episodic memory. Applications include Wang \etal~\cite{2023in_context_wang}, which reformulates Android UI trees into HTML-style text for summarization and QA; MobileGPT~\cite{2024mobilegpt}, which enhances generalization via hierarchical memory, dynamic prompt reconstruction, and human-in-the-loop refinement; and CEIL~\cite{2023ceil}, which optimizes exemplar selection with conditional DPPs for relevance–diversity balance, yielding strong transferability.

\subsubsection{Chain-of-Thought (CoT) Prompting}
\label{sec:Chain_of_Thought}
Chain-of-thought (CoT) prompting augments inputs with natural language rationales, guiding models through intermediate reasoning steps before producing answers (\figref{fig:cot}). Although it increases inference length, CoT improves multi-step decision-making, semantic planning, and interpretability, making it valuable under mobile resource constraints. Recent work highlights \textit{dynamic adaptation}, where CoT adjusts reasoning strategies to context and distribution shifts~\cite{2024instance_cot,2023DOCTOR}. Instance-adaptive Zero-shot CoT~\cite{2024instance_cot} selects prompts via attention-based saliency with substitution or voting, enabling task-agnostic adaptation. DOCTOR~\cite{2023DOCTOR} builds multi-hop rationales through iterative QA with commonsense guidance and dual-stage alignment for coherence. ECoT~\cite{2024robotic_cot} integrates language, perception, and control in a multimodal Transformer, supporting test-time adaptation via meta-learning. PromptCoT~\cite{2024promptcot} extends CoT to diffusion models with staged reasoning and lightweight adapters for efficient multi-task adaptation.

\begin{table*}[t]
\centering
\caption{Summary of Parameter-efficient fine-tuning for test-time FM adaptation in agentic AI systems.}
\vspace{-2mm}
\tiny
\label{tab:peft}
\renewcommand{\arraystretch}{1.05}
\setlength{\tabcolsep}{6pt}
\resizebox{\textwidth}{!}{%
\begin{tabular}{|c|c|c|c|c|c|}
\hline
\multicolumn{3}{|c|}{\textbf{Categories}} & 
\multicolumn{1}{c|}{\textbf{Technique highlight for improving}} & 
\multicolumn{1}{c|}{\textbf{Year}} & 
\multicolumn{1}{c|}{\textbf{Ref}} \\
\hline
\multirow{26}{*}[0ex]{\textbf{\begin{tabular}{c} Parameter-efficient \\ fine-tuning  (PEFT)\\~(\S\ref{sec:peft}) \end{tabular}}}
& \multirow{10}{*}[0ex]{\textbf{\begin{tabular}{c} Additive PEFT \\~(\S\ref{sec:add_peft}) \end{tabular}}} 
& \multirow{5}{*}[0ex]{\textbf{\begin{tabular}{c} Adapter tuning \\~(\S\ref{sec:adapter_tuning})\end{tabular}}} 
& \begin{tabular}[c]{@{}c@{}}Prune adapter weights at initialization, introduce Large-Sparse configuration,boost capacity under parameter budget.\end{tabular} & 2022 & \cite{adapter_sparseadapter} \\
\cline{4-6}
& & & \begin{tabular}[c]{@{}c@{}}Introduce low-rank hypercomplex adapters, compute task-specific weights with Kronecker products, reduce parameter complexity.\end{tabular} & 2021 & \cite{adapter_compacter} \\
\cline{4-6}
& & & \begin{tabular}[c]{@{}c@{}}Extract lightweight proxy submodels, identify and merge important experts, enable efficient on-device adapter tuning.\end{tabular} & 2024 & \cite{adapter_litemoe} \\
\cline{4-6}
& & & \begin{tabular}[c]{@{}c@{}}Reformulate PEFT, pruning, and quantization as adapter-based transformations, enable consistent chaining of modules.\end{tabular} & 2024 & \cite{velingkerclam} \\
\cline{4-6}
& & & \begin{tabular}[c]{@{}c@{}}Use zeroth-order tensor-train adapters, apply parallel contraction and sublinear query scheduling, enable efficient fine-tuning.\end{tabular} & 2024 & \cite{yang2024adazeta} \\
\cline{3-6}

& & \multirow{3}{*}[0ex]{\textbf{\begin{tabular}{c}Prompt tuning\\~(\S\ref{sec:prompt-tuning})\end{tabular}}} 
& \begin{tabular}[c]{@{}c@{}}Prepend continuous prefix vectors, reparameterize via MLP, enable stable training with frozen backbone.\end{tabular} & 2021 & \cite{2021prefix_tuning} \\
\cline{4-6}
& & & \begin{tabular}[c]{@{}c@{}}Insert continuous prompts in all Transformer layers, use reparameterization encoders, improve parameter efficiency.\end{tabular} & 2021 & \cite{2021p_tuning_v2} \\
\cline{4-6}
& & & \begin{tabular}[c]{@{}c@{}}Combine sample selection and noise-aware training, use in-memory computing for scaled retrieval, enable edge tuning.\end{tabular} & 2024 & \cite{2024nvcim_pt} \\
\cline{3-6}

& & \multirow{2}{*}[0ex]{\textbf{\begin{tabular}{c}Other modules\\~(\S\ref{sec:othermodels})\end{tabular}}} 
& \begin{tabular}[c]{@{}c@{}}Introduce learnable scaling vectors, rescale attention and feedforward activations, support mixed-task fine-tuning.\end{tabular} & 2022 & \cite{2022_ia3} \\
\cline{4-6}
& & & \begin{tabular}[c]{@{}c@{}}Train lightweight policy adapters, shape output distributions toward user objectives, combine adapters with base model.\end{tabular} & 2023 & \cite{2023ipa} \\
\cline{2-6}

& \multirow{6}{*}[0ex]{\textbf{\begin{tabular}{c}Selective PEFT\\~(\S\ref{sec:selective_peft})\end{tabular}}} 
& \multirow{4}{*}[0ex]{\textbf{\begin{tabular}{c}Unstructured \\selection~(\S\ref{sec:selective_peft})\end{tabular}}} 
& \begin{tabular}[c]{@{}c@{}}Compute sensitivity scores for bias pruning, prune low-sensitivity biases and reinitialize important ones.\end{tabular} & 2023 & \cite{2023U_BitFit} \\
\cline{4-6}
& & & \begin{tabular}[c]{@{}c@{}}Update child network only, mask gradients of non-child parameters, preserve full model capacity.\end{tabular} & 2021 & \cite{2021CHILD_Tuning} \\
\cline{4-6}
& & & \begin{tabular}[c]{@{}c@{}}Select parameters with largest absolute differences, retrain with binary masks and L1 regularization.\end{tabular} & 2021 & \cite{2021LT_SFT} \\
\cline{4-6}
& & & \begin{tabular}[c]{@{}c@{}}Filter smallest-magnitude parameters with group-wise selection, enable efficient non-IID adaptation.\end{tabular} & 2023 & \cite{2023PaFi} \\
\cline{3-6}

& & \multirow{2}{*}[0ex]{\textbf{\begin{tabular}{c}Structured selection\\~(\S\ref{sec:selective_peft})\end{tabular}}} 
& \begin{tabular}[c]{@{}c@{}}Identify node-level importance with L1-norm changes, select top-r\% nodes for learning.\end{tabular} & 2022 & \cite{2022FAR} \\
\cline{4-6}
& & & \begin{tabular}[c]{@{}c@{}}Update selected rows and columns in weight matrices,  perform in-place fine-tuning.\end{tabular} & 2024 & \cite{2024rocoft} \\
\cline{2-6}

& \multirow{8}{*}[0ex]{\textbf{\begin{tabular}{c}Re-parameterized \\PEFT~(\S\ref{sec:re_param_peft})\end{tabular}}} 
& \multirow{5}{*}[0ex]{\textbf{\begin{tabular}{c}LoRA family\\~(\S\ref{sec:re_param_peft})\end{tabular}}} 
& \begin{tabular}[c]{@{}c@{}}Approximate weight updates with low-rank matrices, update only rank-constrained modules.\end{tabular} & 2022 & \cite{2022lora} \\
\cline{4-6}
& & & \begin{tabular}[c]{@{}c@{}}Sample dynamic target rank, truncate LoRA projection matrices, support flexible inference.\end{tabular} & 2022 & \cite{2022dylora} \\
\cline{4-6}
& & & \begin{tabular}[c]{@{}c@{}}Parameterize updates with SVD-like decomposition, prune singular values based on importance.\end{tabular} & 2023 & \cite{2023adalora} \\
\cline{4-6}
& & & \begin{tabular}[c]{@{}c@{}}Parallelize zeroth-order gradient estimation, update only one matrix in LoRA.\end{tabular} & 2024 & \cite{gao2024enabling} \\
\cline{4-6}
& & & \begin{tabular}[c]{@{}c@{}}Reuse LoRA weights for compressed models, optimize recovery modules for degraded weights.\end{tabular} & 2023 & \cite{2023ca_lora} \\
\cline{3-6}

& & \multirow{3}{*}[0ex]{\textbf{\begin{tabular}{c}LoRA variants\\~(\S\ref{sec:re_param_peft})\end{tabular}}} 
& \begin{tabular}[c]{@{}c@{}}Quantize weight deltas to 1-bit sign representations, use trainable scale factors.\end{tabular} & 2024 & \cite{2024bitdelta} \\
\cline{4-6}
& & & \begin{tabular}[c]{@{}c@{}}Represent updates in frequency domain with DFT, learn shared spectral coefficients.\end{tabular} & 2024 & \cite{2024FourierFT} \\
\cline{4-6}
& & & \begin{tabular}[c]{@{}c@{}}Perform SVD to extract principal component subspace, constrain updates within singular vectors.\end{tabular} & 2024 & \cite{2024pissa} \\
\cline{2-6}

& \multirow{2}{*}[0ex]{\textbf{\begin{tabular}{c}Hybrid PEFT\\~(\S\ref{sec:hybrid_peft})\end{tabular}}} 
& & \begin{tabular}[c]{@{}c@{}}Decompose PEFT design into layer grouping and strategy assignment, refine across backbones.\end{tabular} & 2023 & \cite{2023S4} \\
\cline{4-6}
& & & \begin{tabular}[c]{@{}c@{}}Search over insertion layers and module combinations, optimize parameter budgets, use Bayesian optimization.\end{tabular} & 2024 & \cite{2024autopeft} \\
\hline
\end{tabular}%
}
\vspace{-6mm}
\end{table*}

\begin{figure}[t]
    \centering
  \subfloat[Additive.]{
    \includegraphics[height=0.17\textwidth]{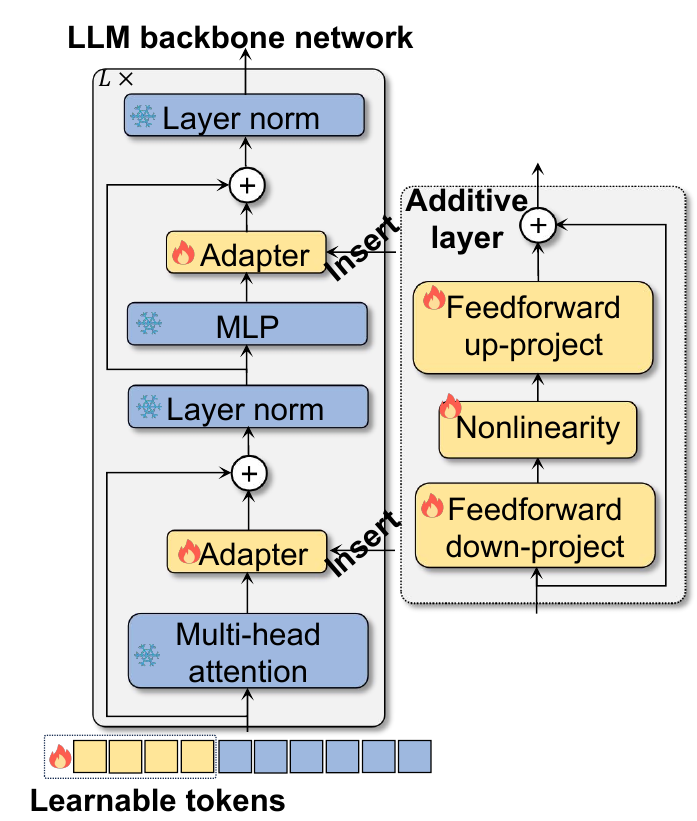}
    \label{fig:reparam}
    }
  \subfloat[Selective.]{
    \includegraphics[height=0.17\textwidth]{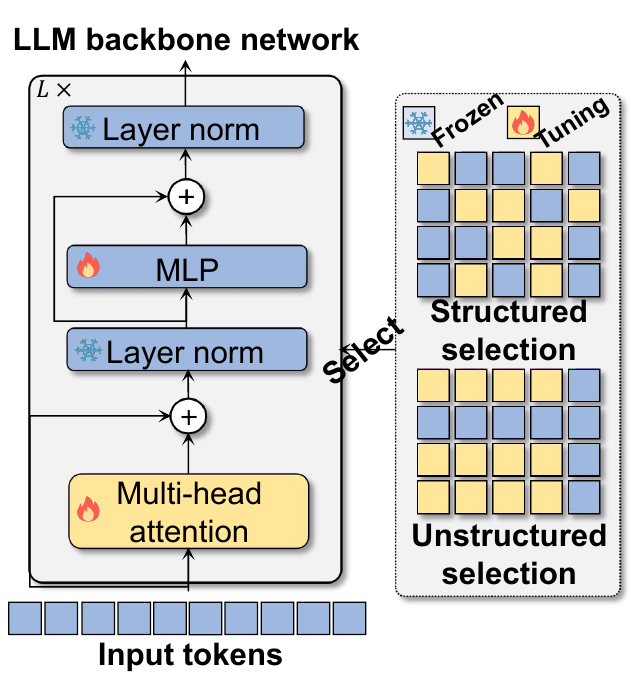}
    \label{fig:reparam}
    }
  \subfloat[Re-parameterized.]{
    \includegraphics[height=0.17\textwidth]{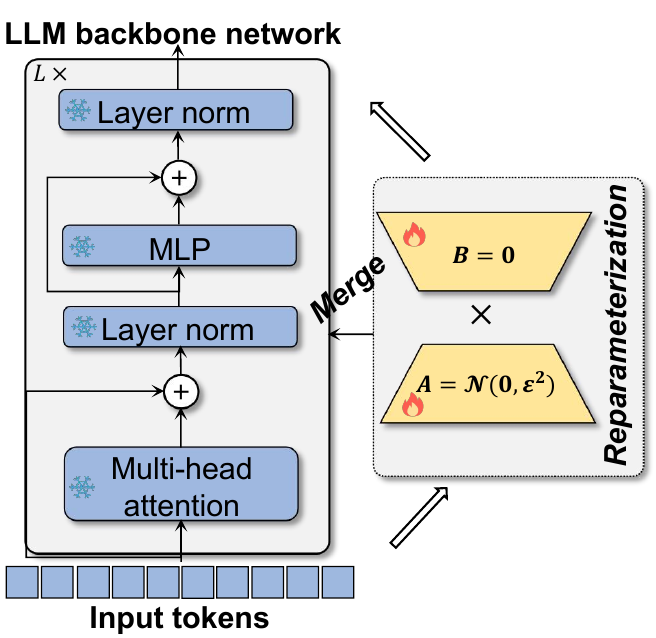}
    \label{fig:reparam}
    }
    \caption{Illustration of parameter-efficient fine-tuning.}
    \vspace{-6mm}
    \label{fig:peft_illustration}
\end{figure}

\subsection{Parameter-efficient Fine-Tuning (PEFT)}
\label{sec:peft}
On mobile and edge platforms, full-parameter fine-tuning is infeasible due to prohibitive memory, compute, and storage demands. Parameter-Efficient Fine-Tuning (PEFT) alleviates this by updating only \textit{small, strategically chosen parameter subsets} while keeping the backbone frozen (\figref{fig:peft_illustration}). This enables rapid, lightweight, and even on-device adaptation, supporting heterogeneous, personalized, and real-time applications.
PEFT methods can be grouped into following four categories.

\subsubsection{Additive PEFT}
\label{sec:add_peft}
Additive PEFT facilitates efficient adaptation of FMs under resource constraints by \textit{injecting} lightweight trainable modules into Transformer layers or input embeddings, while keeping the backbone parameters frozen.
It reduces training-time memory and computation overhead by updating only a small number of the inserted modules.
Also, it avoids gradient storage and optimizer tracking for the frozen backbone and limits activation retention to the trainable components, resulting in efficient fine-tuning under memory and energy constraints.
Depending on the insertion location and architectural design, additive PEFT methods are commonly categorized into adapter tuning~\cite{adapter_sparseadapter,adapter_compacter,adapter_litemoe}, prompt tuning~\cite{2021prefix_tuning,2021p_tuning_v2,2024nvcim_pt}, and other integration-based approaches~\cite{2022_ia3,2023ipa}.

\textit{\textbf{a. Adapter tuning.}}
\label{sec:adapter_tuning}
Adapter tuning adapts FMs by inserting lightweight modules into Transformer layers while freezing the backbone, enabling efficient and task-specific customization. Recent advances improve efficiency, modularity, and on-device suitability. SparseAdapter~\cite{adapter_sparseadapter} prunes adapter weights at initialization to scale capacity under fixed budgets. Compacter~\cite{adapter_compacter} applies low-rank hypercomplex decomposition, reducing complexity from $O(kd)$ to $O(k+d)$. LiteMoE~\cite{adapter_litemoe} extracts lightweight proxy submodels from MoE-based LLMs for multi-task personalization. AdaZeta~\cite{yang2024adazeta} introduces tensorized forward-only adapters with adaptive scheduling for memory-efficient tuning. CLAM~\cite{velingkerclam} integrates PEFT, pruning, and quantization into unified adapter-based transformations for consistent specialization and compression.

\begin{figure}[t]
    \centering
  \subfloat[P-tuning v1.]{
    \includegraphics[height=0.10\textwidth]{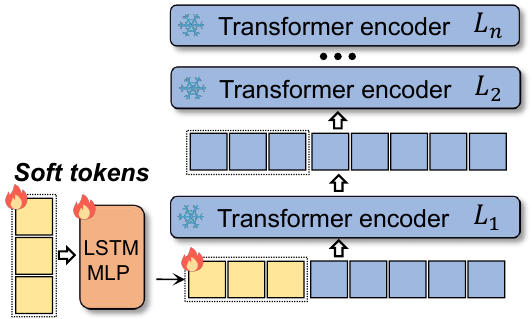}
    \label{fig:p_tuning_v1}
    }
  \hspace{-2mm}
  \subfloat[P-tuning v2.]{
    \includegraphics[height=0.10\textwidth]{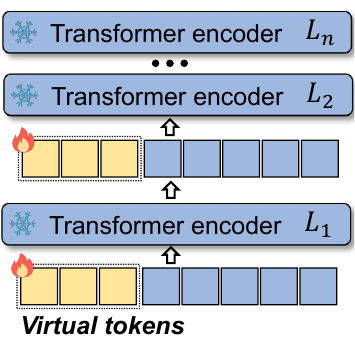}
    \label{fig:p_tuning_v2}
    }
    \hspace{-2mm}
    \subfloat[Prefix tuning.]{
    \includegraphics[height=0.10\textwidth]{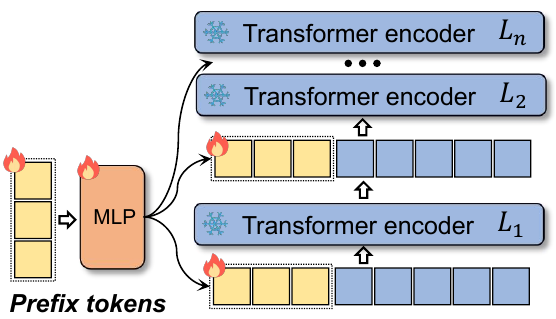}
    \label{fig:prefix_tuning}
    }
    \caption{Illustration of prompt tuning.}
    \label{fig:add_prompt_tuning}
    \vspace{-6mm}
\end{figure}

\textit{\textbf{b. Prompt tuning.}}
\label{sec:prompt-tuning}
Prompt tuning adapts FMs by optimizing a small set of task-specific embeddings while freezing the backbone (\figref{fig:add_prompt_tuning}), offering lightweight adaptation with minimal parameters and strong data efficiency. Structure-aware variants refine \textit{injection position and form} within Transformers. Prefix-Tuning~\cite{2021prefix_tuning} prepends trainable prefix embeddings across layers via MLP reparameterization; P-Tuning~\cite{2021p_tuning_v2} learns continuous embeddings with an LSTM encoder at the input, and P-Tuning v2 extends this to all layers for greater expressivity. NVCiM-PT~\cite{2024nvcim_pt} further tailors prompt tuning to edge devices through hardware-aware encoding, sample selection, and non-volatile memory optimization, enabling efficient and robust on-device adaptation.

\textit{\textbf{c. Other structurally integrated modules.}}
\label{sec:othermodels}
Beyond adapters and prompts, structurally integrated PEFT modules inject lightweight transformations into frozen backbones via \textit{scaling}, \textit{policy control}, or \textit{distribution reshaping}, enabling multi-task and inference-time adaptation with negligible cost. (IA)\textsuperscript{3}~\cite{2022_ia3} applies learnable task-specific scaling vectors to attention and feedforward activations, supporting mixed-task batches with minimal overhead, while IPA~\cite{2023ipa} leverages a lightweight policy adapter trained via reinforcement learning and integrated through a product-of-experts mechanism to steer distributions efficiently across diverse objectives.

\label{sec:selective_peft}
\begin{figure}[t]
    \centering
    \includegraphics[width=0.38\textwidth]{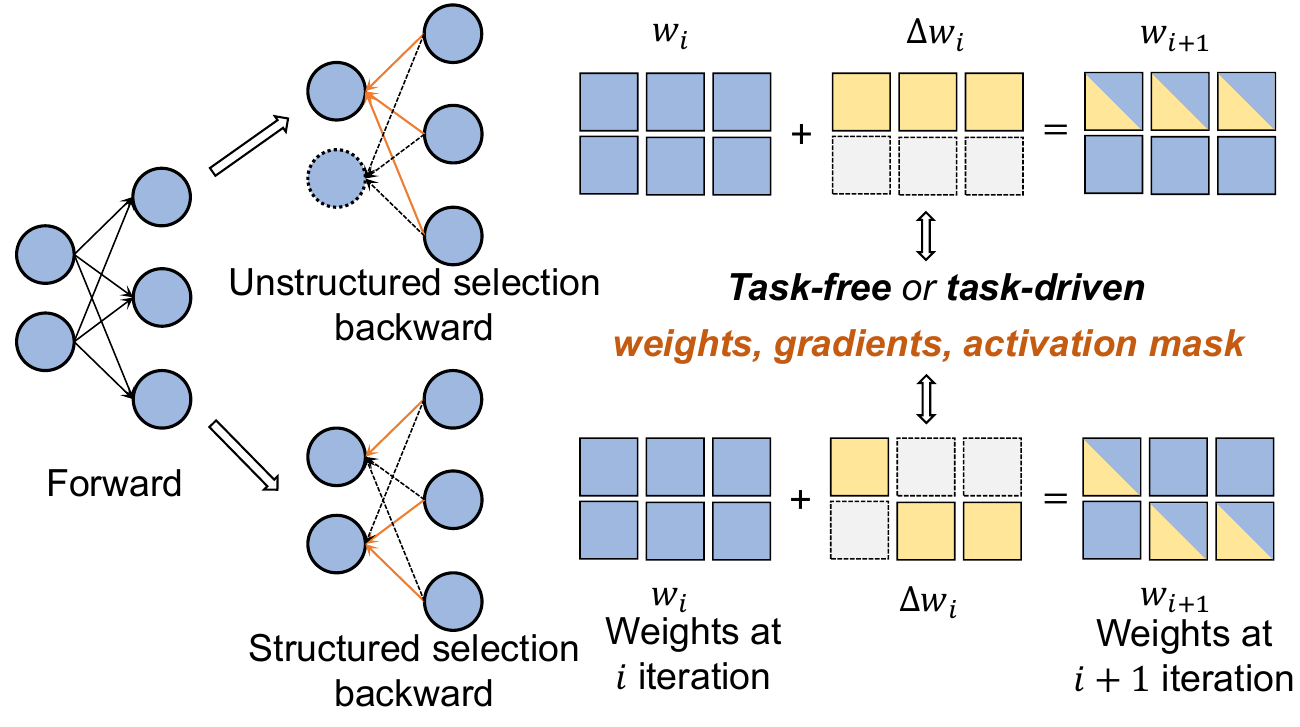}
    \caption{Illustration of selective parameter-efficient fine-tuning.}
    \vspace{-5mm}
    \label{fig:select_peft}
\end{figure}

\subsubsection{Selective PEFT}
Unlike additive PEFT, which inserts auxiliary modules, selective PEFT updates only a task-relevant subset of native parameters to reduce overhead while maintaining adaptability (\figref{fig:select_peft}). 
Approaches fall into two categories:
\textit{First, structured selection} updates larger architectural units (\eg layers, blocks, attention heads) for coarse-grained yet efficient tuning. FAR~\cite{2022FAR} ranks FFN nodes by $L1$-norm changes and fine-tunes the top-$r\%$, reconfiguring memory layout to reduce fragmentation, while RoCoFT~\cite{2024rocoft} restricts updates to selected rows/columns of weight matrices, achieving accuracy comparable to full fine-tuning.
\textit{Second, unstructured selection} targets individual parameters for maximal efficiency. U-BitFit~\cite{2023U_BitFit} prunes low-sensitivity biases via gradient signals; CHILD-Tuning~\cite{2021CHILD_Tuning} masks gradients to update only a task-aware “child network”; LT-SFT~\cite{2021LT_SFT} leverages lottery ticket sparsity to retrain selected weights; PaFi~\cite{2023PaFi} tunes merely 0.5\% of parameters chosen by group-wise magnitude, enabling efficient adaptation even in federated settings.

\begin{figure}[t]
    \centering
  \subfloat[LoRA.]{
    \includegraphics[height=0.105\textwidth]{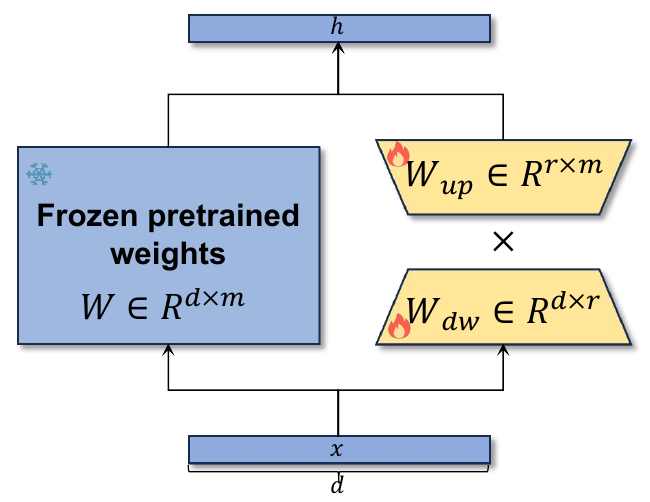}
    \label{fig:Lora}
    }
  \subfloat[DyLoRA.]{
    \includegraphics[height=0.105\textwidth]{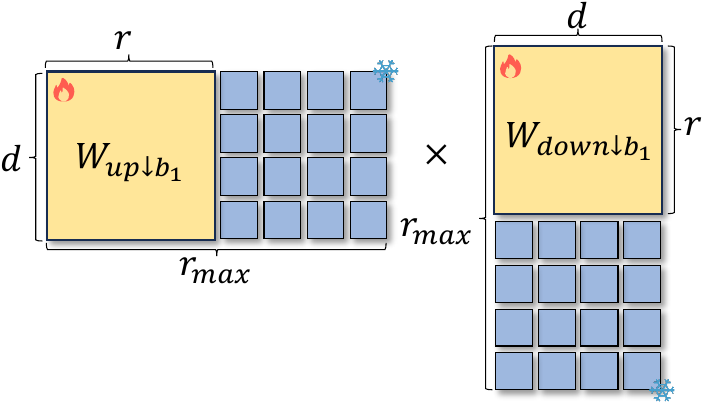}
    \label{fig:DyLoRA}
    }
  \subfloat[AdaLoRA.]{
    \includegraphics[height=0.105\textwidth]{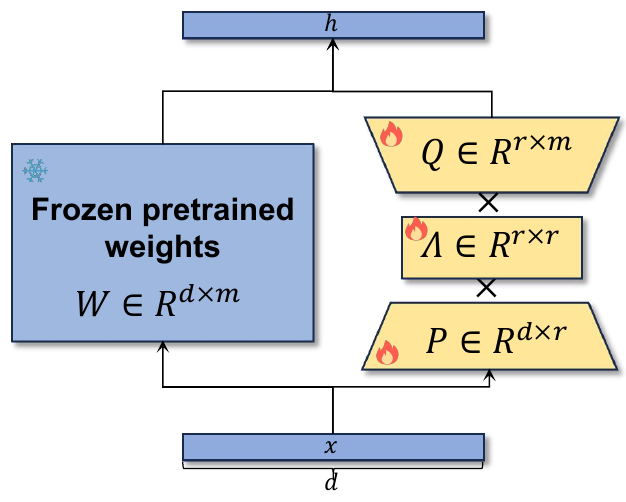}
    \label{fig:adaLoRA}
    }
    \caption{Illustration of LoRA Family.}
    \label{fig:peft_LoRA_Family}
    \vspace{-5mm}
\end{figure}

\subsubsection{Re-parameterized PEFT (LoRA Family)}
\label{sec:re_param_peft}
It reformulates parameter updates via low-rank decomposition, training only rank-constrained matrices while freezing the backbone. This achieves lightweight, memory-efficient adaptation without inference overhead.
The \textit{LoRA family} is the canonical design.
Given $\mathbf{W}_0 \in \mathbb{R}^{d \times m}$, updates are expressed as 
In Low-Rank Adaptation (LoRA)~\cite{2022lora}, the update to a weight matrix $\mathbf{W}_0 \in \mathbb{R}^{d \times m}$ is re-parameterized as
$$
\mathbf{W} = \mathbf{W}_0 + \Delta \mathbf{W} = \mathbf{W}_0 + \alpha \cdot \mathbf{A}\mathbf{B},
$$
where $\mathbf{A}\in \mathbb{R}^{d \times r}, \mathbf{B}\in \mathbb{R}^{r \times m}, r \ll \min(d,m)$. Only $\mathbf{A},\mathbf{B}$ are trained, cutting parameter and memory costs substantially~\cite{2022lora}.
Recent extensions enhance LoRA with \textit{dynamic rank adjustment}, \textit{budget allocation}, \textit{compression recovery}, and \textit{gradient-free tuning}. DyLoRA~\cite{2022dylora} samples target ranks to avoid exhaustive search; AdaLoRA~\cite{2023adalora} prunes singular values for adaptive rank budgeting; 
CA-LoRA restores compressed LLMs via knowledge inheritance; 
Delta-LoRA~\cite{zi2023delta} jointly updates pretrained weights and low-rank deltas; 
P-RGE~\cite{gao2024enabling} enables forward-pass–only tuning via zeroth-order estimation.
LoRA-inspired \textit{variants} introduce more \textit{flexible} low-rank structures. BitDelta~\cite{2024bitdelta} compresses weight differences into 1-bit signs with scaling factors for multi-tenant deployment. FourierFT~\cite{2024FourierFT} encodes updates as sparse Fourier signals, achieving LoRA-level accuracy with $\leq$0.1\% parameters. PiSSA~\cite{2024pissa} initializes updates with top singular vectors from SVD, improving convergence and quantization robustness.

\begin{figure}[t]
    \centering
  \subfloat[BitDelta.]{
    \includegraphics[height=0.12\textwidth]{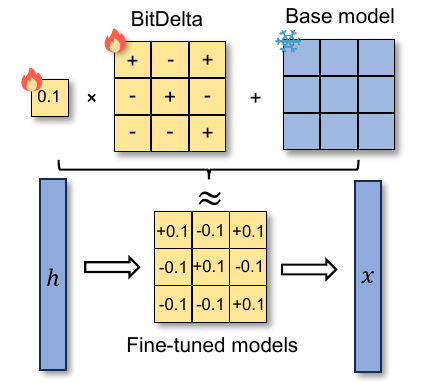}
    \label{fig:BitDelta}
    }
  \subfloat[PiSSA.]{
    \includegraphics[height=0.12\textwidth]{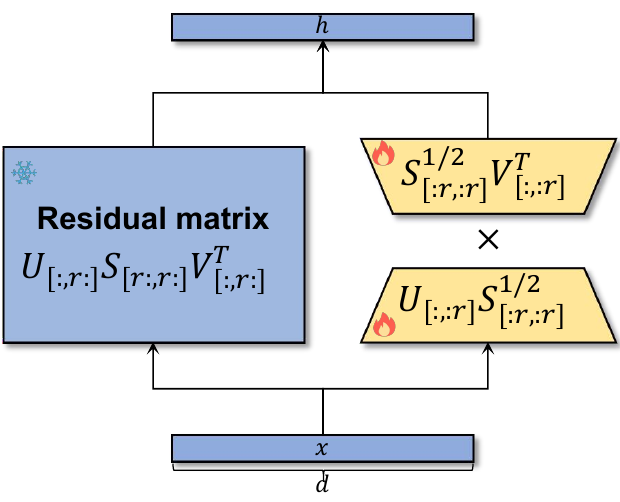}
    \label{fig:PiSSA}
    }
  \subfloat[FourierFT.]{
    \includegraphics[height=0.12\textwidth]{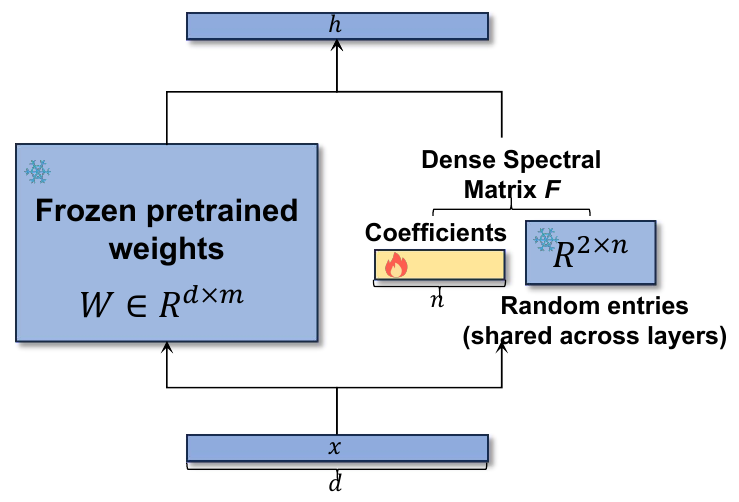}
    \label{fig:FourierFT}
    }
    \caption{Illustration of LoRA-inspired variants.}
    \label{fig:LoRA_inspired}
    \vspace{-5mm}
\end{figure}


\subsubsection{Hybrid PEFT}
\label{sec:hybrid_peft}
Hybrid PEFT combines multiple strategies (\eg LoRA, Adapters, Prompt Tuning) to leverage complementary strengths in efficiency, flexibility, and generalization, making it well-suited for resource-constrained agent scenarios~\cite{2024flexllm}.
S4~\cite{2023S4} formalizes hybrid design spaces along four axes—layer grouping, parameter allocation, group selection, and strategy assignment—identifying robust patterns such as spindle-shaped grouping and diverse strategy assignment across layers. AUTOPEFT~\cite{2024autopeft} automates hybrid design via hierarchical search and multi-objective Bayesian optimization, exploring insertion layers, parameter budgets, and module combinations (\eg serial/parallel adapters with prefix-tuning) to yield Pareto-optimal and transferable configurations.

\begin{table*}[t]
\centering
\caption{Summary of memory- and computation-efficient PEFT for FM adaptation in agentic AI systems.}
\vspace{-2mm}
\tiny
\label{tab:eff_peft}
\renewcommand{\arraystretch}{1.1}
\setlength{\tabcolsep}{6pt}
\resizebox{\textwidth}{!}{%
\begin{tabular}{|c|c|c|c|c|}
\hline
\multicolumn{2}{|c|}{\textbf{Categories}} & 
\multicolumn{1}{c|}{\textbf{Technique highlight for improving}} & 
\multicolumn{1}{c|}{\textbf{Year}} & 
\multicolumn{1}{c|}{\textbf{Ref}} \\
\hline
\multirow{13}{*}[\dimexpr 0ex\relax]{\centering\textbf{\begin{tabular}{c}Memory- and \\ computation-efficient \\ PEFT~(\S\ref{sec:efficient_peft})\end{tabular}}} 
& \multirow{4}{*}[\dimexpr-0ex\relax]{\centering\textbf{\begin{tabular}{c}Pruning-enhanced \\PEFT~(\S\ref{sec:pruning_peft})\end{tabular}}} 
& \begin{tabular}[c]{@{}c@{}}Structured pruning with LoRA gradients: iterative LoRA pruning/fine-tuning, hardware-friendly sparsity\end{tabular} & 2023 & \cite{2023loraprune} \\
\cline{3-5}
& & \begin{tabular}[c]{@{}c@{}}Prune redundant adapter params at init, scale capacity with bottleneck-sparsity balance, keep param efficiency\end{tabular} & 2022 & \cite{adapter_sparseadapter} \\
\cline{3-5}
& & \begin{tabular}[c]{@{}c@{}}Score adapter salience, adjust adapter ranks with layer importance, use self-distillation.\end{tabular} & 2024 & \cite{2024apt} \\
\cline{3-5}
& & \begin{tabular}[c]{@{}c@{}}Deploy client-specific LoRA modules, perform rank self-pruning, aggregate with sparsity weighting.\end{tabular} & 2024 & \cite{2024HETLORA} \\
\cline{2-5}

& \multirow{4}{*}[\dimexpr-0ex\relax]{\centering\textbf{\begin{tabular}{c}Quantization-aware \\PEFT~(\S\ref{sec:quant_peft})\end{tabular}}} 
& \begin{tabular}[c]{@{}c@{}}Quantize weights to 4-bit NF4, apply double quantization, offload  optimizer states with paged optimizer.\end{tabular} & 2023 & \cite{dettmers2023qloraefficientfinetuningquantized} \\
\cline{3-5}
& & \begin{tabular}[c]{@{}c@{}}Quantize pretrained weights, keep merged weights quantized for inference.\end{tabular} & 2024 & \cite{2023qa_lora} \\
\cline{3-5}
& & \begin{tabular}[c]{@{}c@{}}Quantize weights and LoRA adapters to FP8, use gradient scaling , fuse operators.\end{tabular} & 2024 & \cite{20248_bit} \\
\cline{3-5}
& & \begin{tabular}[c]{@{}c@{}}Preserve weak columns in FP16, group scaling, fine-tune sensitive columns.\end{tabular} & 2024 & \cite{2024qeft} \\
\cline{2-5}

& \multirow{5}{*}[\dimexpr-0ex\relax]{\centering\textbf{\begin{tabular}{c}Backpropagation-free \\PEFT~(\S\ref{sec:memory_peft})\end{tabular}}} 
& \begin{tabular}[c]{@{}c@{}}Generate task-specific parameters with hypernetworks, adapt without backpropagation.\end{tabular} & 2023 & \cite{2023hypertuning} \\
\cline{3-5}
& & \begin{tabular}[c]{@{}c@{}}Avoid backpropagation through backbone, leverage ladder side network with pruning and layer dropping.\end{tabular} & 2022 & \cite{2022lst} \\
\cline{3-5}
& & \begin{tabular}[c]{@{}c@{}}Zero-order gradient estimation with two forward passes, skip  backpropagation.\end{tabular} & 2023 & \cite{2023MeZO} \\
\cline{3-5}
& & \begin{tabular}[c]{@{}c@{}}Project gradients into low-rank subspace, leverage gradient matrix structure.\end{tabular} & 2024 & \cite{2024galore} \\
\cline{3-5}
& & \begin{tabular}[c]{@{}c@{}}Combine pruning and quantization for unified compression, tune layers  adaptively with hardware scheduling.\end{tabular} & 2024 & \cite{2024edge_LLM} \\
\hline
\end{tabular}%
}
\vspace{-6mm}
\end{table*}

\begin{figure}[t]
    \centering
  \subfloat[Normal PEFT.]{
    \includegraphics[height=0.185\textwidth]{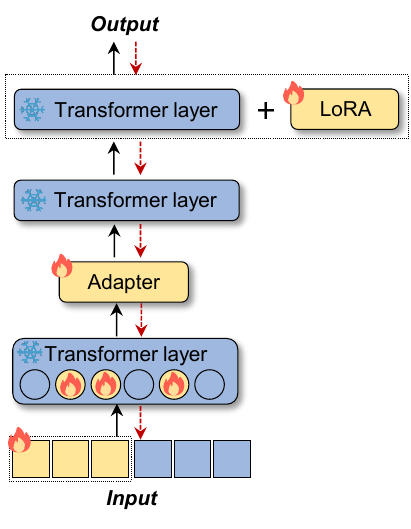}
    \label{fig:normal_peft}
    }
  \subfloat[Pruning PEFT.]{
    \includegraphics[height=0.185\textwidth]{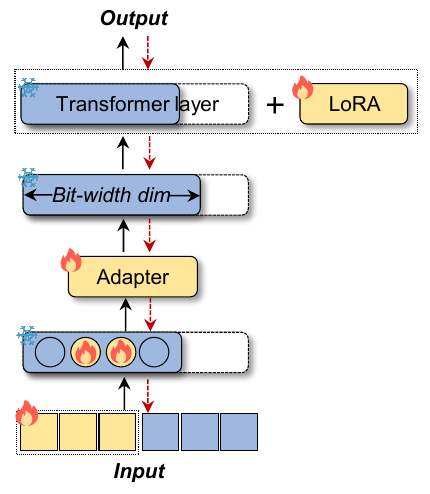}
    \label{fig:pruning_peft}
    }
  \subfloat[Quantization PEFT.]{
    \includegraphics[height=0.185\textwidth]{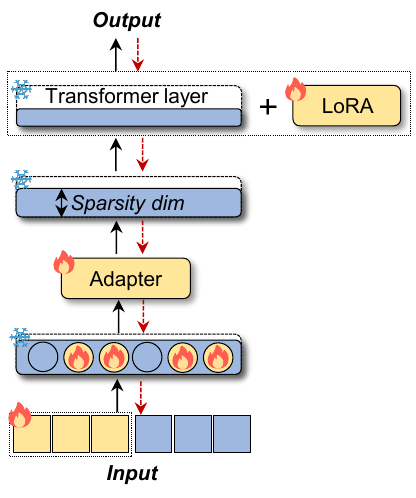}
    \label{fig:meZo}
    }
    \caption{Illustration of efficient PEFT design.}
    \vspace{-6mm}
    \label{fig:efficient_peft_design}
\end{figure}

\subsubsection{Memory- and Computation-efficient PEFT}
\label{sec:efficient_peft}
In mobile/edge agent scenarios, FMs must adapt under tight memory, computation, and latency budgets. Conventional PEFT can incur high delays and memory peaks, particularly during backpropagation when activations, gradients, and optimizer states coexist. To address this, memory- and computation-efficient PEFT strategies (\figref{fig:efficient_peft_design}, \tabref{tab:eff_peft}) have emerged:
\textit{pruning-enhanced PEFT}~\cite{2023loraprune,2024apt,adapter_sparseadapter,2024HETLORA} trims redundant parameters or adapter weights to cut training cost while retaining adaptation capacity;
\textit{quantization-aware PEFT}~\cite{dettmers2023qloraefficientfinetuningquantized,20248_bit,2024qeft} compresses trainable modules into low-bit formats, balancing accuracy with memory and bandwidth efficiency;
\textit{backpropagation-free PEFT}~\cite{2022lst,2023hypertuning,2023MeZO,2024galore} eliminates gradients via optimizer-free updates, forward-only optimization, or low-rank projection, lowering memory overhead and enabling scalable on-device tuning.

\textbf{\textit{a. Pruning-enhanced PEFT.}}
\label{sec:pruning_peft}
This approach combines parameter pruning with PEFT modules to boost sparsity and efficiency, supporting lightweight deployment and adaptive retraining.
LoRAPrune~\cite{2023loraprune} exploits LoRA’s low-rank structure for gradient-based structured pruning without full-model updates;
APT~\cite{2024apt} adaptively prunes adapters via outlier-aware salience scoring and dynamic rank adjustment;
SparseAdapter~\cite{adapter_sparseadapter} prunes at initialization to construct sparse yet expressive adapters;
HETLoRA~\cite{2024HETLORA} integrates rank self-pruning with sparsity-weighted aggregation to enable client-specific adaptation in federated settings.

\textbf{\textit{b. Quantization-aware PEFT.}}
\label{sec:quant_peft}
Quantization-aware PEFT reduces memory and compute by applying low-bit quantization to pretrained weights while fine-tuning selected parameters, enabling hardware-friendly, accuracy-preserving adaptation.
QLoRA~\cite{dettmers2023qloraefficientfinetuningquantized} introduces 4-bit NF4 quantization with double quantization and paged optimizers for long-sequence training;
QA-LoRA~\cite{2023qa_lora} adopts group-wise quantization to preserve adaptation freedom under GPU limits;
8-bit Transformer~\cite{20248_bit} quantizes both backbone and LoRA adapters to FP8/Posit8 with fused operators for stable 8-bit training;
QEFT~\cite{2024qeft} selectively retains weak FP16 columns while quantizing others, balancing efficiency with accuracy.

\begin{figure}[t]
    \centering
  \subfloat[LST.]{
    \includegraphics[height=0.22\textwidth]{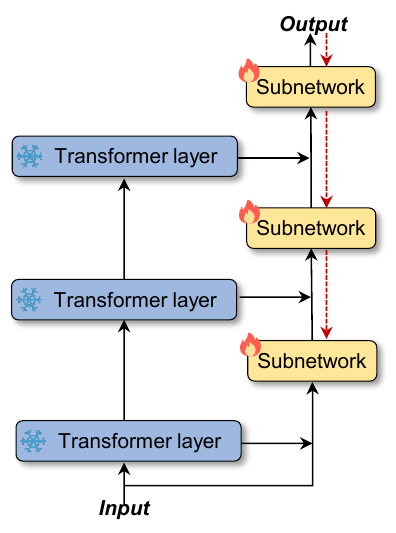}
    \label{fig:LST}
    }
  \subfloat[Layer selection.]{
    \includegraphics[height=0.22\textwidth]{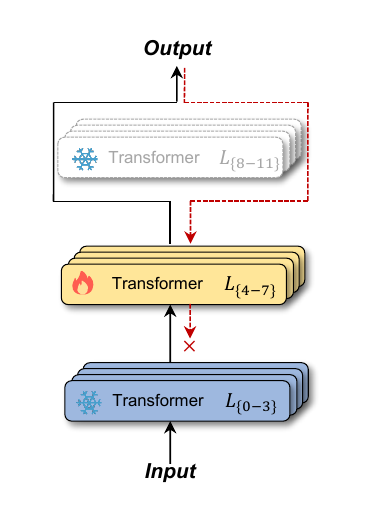}
    \label{fig:layer_select}
    }
  \subfloat[MeZo.]{
    \includegraphics[height=0.22\textwidth]{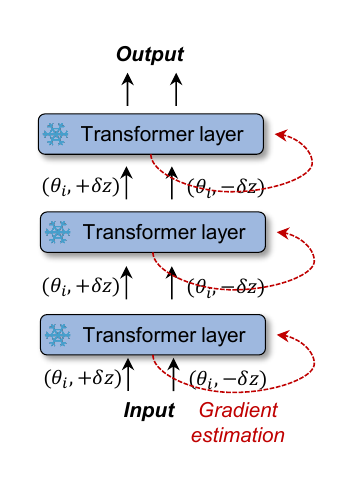}
    \label{fig:meZo}
    }
    \caption{Illustration of memory-efficient PEFT.}
    \vspace{-6mm}
    \label{fig:memory_efficient_peft}
\end{figure}

\textbf{\textit{c. Backpropagation-free PEFT.}}
\label{sec:memory_peft} 
This paradigm alleviates memory bottlenecks and optimizer overhead by removing or approximating gradient computation (\figref{fig:memory_efficient_peft}).
LST~\cite{2022lst} introduces a ladder side network to bypass backbone backpropagation, cutting activation storage while retaining adaptation.
HyperTuning~\cite{2023hypertuning} employs a hypernetwork to generate task-specific parameters (\eg prompts, LoRA) without gradients, lowering memory and compute costs.
MeZO~\cite{2023MeZO} applies zeroth-order optimization, estimating gradients via forward passes only, making adaptation memory usage close to inference.
GaLore~\cite{2024galore} projects gradients into low-rank subspaces, reducing optimizer state memory while preserving flexibility for both full and parameter-efficient tuning.

\begin{table*}[t]
\centering
\caption{Summary of memory-augmented adaptation for real-time FM adaptation on agentic AI systems.}
\vspace{-2mm}
\tiny
\label{tab:memory_augmented}
\renewcommand{\arraystretch}{1.1}
\setlength{\tabcolsep}{6pt}
\resizebox{\textwidth}{!}{%
\begin{tabular}{|c|c|c|c|c|c|}
\hline
\multicolumn{3}{|c|}{\textbf{Categories}} & 
\multicolumn{1}{c|}{\textbf{Technique highlight for improving}} & 
\multicolumn{1}{c|}{\textbf{Year}} & 
\multicolumn{1}{c|}{\textbf{Ref}} \\
\hline
\multirow{21}{*}[-0ex]{\textbf{\begin{tabular}{c} Memory-augmented \\ adaptation~(\S\ref{sec:quant_peft})\end{tabular}}} 
& \multirow{12}{*}[-0ex]{\textbf{\begin{tabular}{c} Contextual working \\memory \\ ~(\S\ref{sec:context_working_memory})\end{tabular}}} 
& \multirow{3}{*}[-0ex]{\textbf{\begin{tabular}{c} Prompt adaptation \\ ~(\S\ref{sec:prompt_adaptation})\end{tabular}}} 
& \begin{tabular}[c]{@{}c@{}}Modify Transformer attention masks, compress prompts into gist tokens,  enable zero-shot gist prefix prediction.\end{tabular} & 2023 & \cite{2023Gisting} \\
\cline{4-6}
& & & \begin{tabular}[c]{@{}c@{}}Apply dynamic prompt compression, use iterative token-level compression, align with instruction fine-tuning.\end{tabular} & 2023 & \cite{llmlingua} \\
\cline{4-6}
& & & \begin{tabular}[c]{@{}c@{}}Model prompt optimization as MDP, integrate error feedback via MCTS, refine prompts iteratively.\end{tabular} & 2023 & \cite{promptagent} \\
\cline{3-6}

& & \multirow{2}{*}[-0ex]{\textbf{\begin{tabular}{c} Long context \\ distillation ~(\S\ref{sec:long_context_distillation})\end{tabular}}} 
& \begin{tabular}[c]{@{}c@{}}Model example selection as MDP, use marginal utility rewards, improve generalization across models.\end{tabular} & 2022 & \cite{LCD_1} \\
\cline{4-6}
& & & \begin{tabular}[c]{@{}c@{}}Divide long contexts into parallel windows, reuse positional embeddings, restrict attention to within-window tokens.\end{tabular} & 2024 & \cite{LCD_PCW} \\
\cline{3-6}

& & \multirow{2}{*}[-0ex]{\textbf{\begin{tabular}{c} Role playing \\ ~(\S\ref{sec:role_palying})\end{tabular}}} 
& \begin{tabular}[c]{@{}c@{}}Adopt fact-grounded scene simulation, reconstruct experience pipeline, forget irrelevant knowledge for consistency.\end{tabular} & 2023 & \cite{Character_LLM} \\
\cline{4-6}
& & & \begin{tabular}[c]{@{}c@{}}Use orchestrator with task and progress ledgers, assign tasks and monitor outcomes, enable collaborative agent reasoning.\end{tabular} & 2024 & \cite{magentic_One} \\
\cline{3-6}

& & \multirow{3}{*}[-0ex]{\textbf{\begin{tabular}{c} Self correction \\ ~(\S\ref{sec:self_correction})\end{tabular}}} 
& \begin{tabular}[c]{@{}c@{}}Generate self-feedback for output refinement, act as generator, refiner, and feedback provider.\end{tabular} & 2023 & \cite{madaan2023self} \\
\cline{4-6}
& & & \begin{tabular}[c]{@{}c@{}}Assess confidence of model outputs, enable adaptive self-correction.\end{tabular} & 2024 & \cite{self_cor_confidence} \\
\cline{4-6}
& & & \begin{tabular}[c]{@{}c@{}}Interact with external tools for validation, generate actionable feedback with LLM.\end{tabular} & 2023 & \cite{self_cor_critic} \\
\cline{2-6}
& \multirow{6}{*}[-0ex]{\textbf{\begin{tabular}{c} Task-episodic \\memory \\ ~(\S\ref{sec:task_episodic_memory})\end{tabular}}} 
& \multirow{2}{*}[-0ex]{\textbf{\begin{tabular}{c} Data replay \\ ~(\S\ref{sec:task_episodic_memory})\end{tabular}}} 
& \begin{tabular}[c]{@{}c@{}}Combine embedding entropy and domain scores, build data buffer with diversity, replay dialogues for adaptation.\end{tabular} & 2024 & \cite{data_replay_1} \\
\cline{4-6}
& & & \begin{tabular}[c]{@{}c@{}}Cluster past data for replay, model replay as multi-armed bandit, select data to mitigate forgetting.\end{tabular} & 2024 & \cite{data_replay_2} \\
\cline{3-6}

& & \multirow{4}{*}[-0ex]{\textbf{\begin{tabular}{c} Self experiences\\ ~(\S\ref{sec:task_episodic_memory})\end{tabular}}} 
& \begin{tabular}[c]{@{}c@{}}Store self-experiences as triplet knowledge, automate retrieval with fuzzy matching.\end{tabular} & 2023 & \cite{self_exp_ret_llm} \\
\cline{4-6}
& & & \begin{tabular}[c]{@{}c@{}}Use SQL databases as symbolic memory, dynamically generate SQL commands.\end{tabular} & 2023 & \cite{self_exp_chatdb} \\
\cline{4-6}
& & & \begin{tabular}[c]{@{}c@{}}Encode corpus knowledge into model parameters, use pseudo query-document pairs.\end{tabular} & 2022 & \cite{self_exp_corpusbrain} \\
\cline{4-6}
& & & \begin{tabular}[c]{@{}c@{}}Build on-the-fly memos as task-episodic memory, maintain conversation consistency.\end{tabular} & 2023 & \cite{self_exp_memochat} \\
\cline{2-6}

& \multirow{5}{*}[-0ex]{\textbf{\begin{tabular}{c} External semantic\\ memory\\ ~(\S\ref{sec:external_semantic_memory})\end{tabular}}} 
& \multirow{3}{*}[-0ex]{\textbf{\begin{tabular}{c} Continual knowledge \\graph learning \\ ~(\S\ref{sec:external_semantic_memory})\end{tabular}}} 
& \begin{tabular}[c]{@{}c@{}}Construct domain knowledge graphs with LLMs, align with KG feedback for updates, address domain gaps.\end{tabular} & 2024 & \cite{con_know-graph_1} \\
\cline{4-6}
& & & \begin{tabular}[c]{@{}c@{}}Use evidence graph mining with LLMs, aggregate evidence graphs, enable graph-of-thoughts inference.\end{tabular} & 2023 & \cite{con_know-graph_2} \\
\cline{4-6}
& & & \begin{tabular}[c]{@{}c@{}}Extract triples for KG construction, explore nodes and relationships, enable multi-hop KGQA.\end{tabular} & 2024 & \cite{con_know-graph_3} \\
\cline{3-6}

& & \multirow{2}{*}[-0ex]{\textbf{\begin{tabular}{c} Continual document \\learning ~(\S\ref{sec:external_semantic_memory})\end{tabular}}} 
& \begin{tabular}[c]{@{}c@{}}Use indexing APIs for document-level updates, skip unchanged blocks, avoid redundant updates.\end{tabular} & 2024 & \cite{2024langchain} \\
\cline{4-6}
& & & \begin{tabular}[c]{@{}c@{}}Manage document storage with incremental updates, index and parse new data.\end{tabular} & 2023 & \cite{2023LlamaIndex} \\
\hline
\end{tabular}%
}
\vspace{-6mm}
\end{table*}

\begin{figure*}[t]
    \centering
    \subfloat[Contextual working memory.]{
        \includegraphics[height=0.12\textwidth]{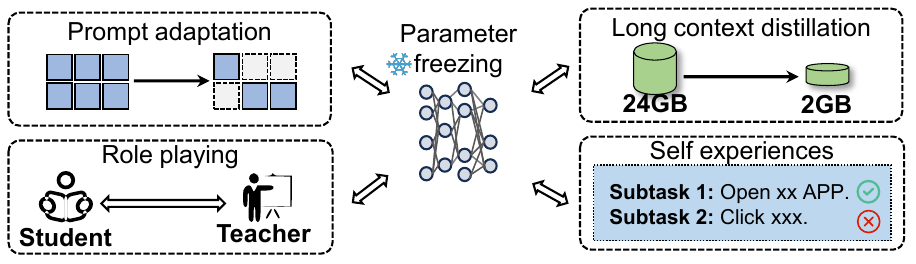}
        \label{fig:contextual_working_memory}
    }
    \subfloat[Task episodic memory.]{
        \includegraphics[height=0.12\textwidth]{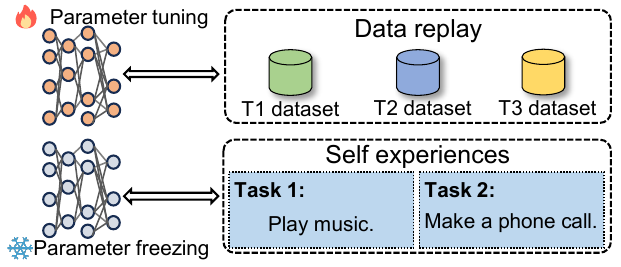}
        \label{fig:task_episodic_memory}
    }
    \subfloat[External semantic memory.]{
        \includegraphics[height=0.12\textwidth]{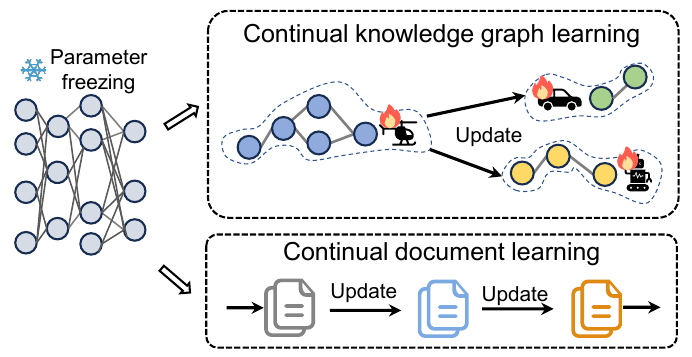}
        \label{fig:external_semantic_memory}
    }
    \vspace{-2mm}
    \caption{Illustration of memory-augmented adaptation.}
    \vspace{-4mm}
    \label{fig:memory_types}
\end{figure*}

\subsection{Memory-Augmented Adaptation}
\label{sec:memory_aug_adaptation} 
Unlike PEFT, which adapts parameters, memory-augmented adaptation equips FMs with external or auxiliary memory for recording, updating, and recalling task-relevant information during inference, enabling flexible test-time adaptation with minimal parameter updates (\tabref{tab:memory_augmented}).

\subsubsection{Contextual Working Memory}
\label{sec:context_working_memory}
Agentic FMs on embedded devices often forget instructions or misinterpret follow-ups due to limited context windows and lack of persistent state.
Contextual working memory provides short-term, task-specific storage of recent inputs, observations, and intermediate results (\figref{fig:contextual_working_memory}), active only within an ongoing interaction and cleared on context shift.
It supports coherent inference and adaptive responses via mechanisms such as prompt adaptation~\cite{2023Gisting,llmlingua,promptagent}, long-context distillation~\cite{LCD_PCW,LCD_1}, role playing~\cite{magentic_One,Character_LLM}, and self-correction~\cite{self_cor_critic,madaan2023self}.

\textbf{\textit{a. Prompt adaptation.}}
\label{sec:prompt_adaptation}
Prompt adaptation compresses and refines historical interactions and instructions within limited context windows, enhancing inference consistency without parameter updates.
Strategies include \textit{soft compression}~\cite{2023Gisting}, which distills prompts into gist tokens, \textit{hard compression}~\cite{llmlingua}, which prunes redundant tokens with dynamic ratio allocation, and \textit{optimization}~\cite{promptagent}, which refines prompt structures via self-reflective feedback.

\textbf{\textit{b. Long-context distillation.}}
\label{sec:long_context_distillation}
Long-context distillation extracts the most relevant spans from lengthy inputs to fit limited prompt windows.
Methods include \textit{context pruning}~\cite{LCD_1}, which models example selection as an RL-based MDP, and \textit{context fusion}~\cite{LCD_PCW}, which aggregates information via parallel context windows with restricted attention and shared task tokens.

\textbf{\textit{c. Role playing.}}
\label{sec:role_palying}
Role playing steers inference and interaction by assigning agents task-specific identities.
\textit{Single-agent role playing}~\cite{Character_LLM} simulates roles (\eg planner, explainer) to improve reasoning style and reduce hallucinations, while \textit{multi-agent role playing}~\cite{magentic_One} orchestrates collaboration with distinct identities and task ledgers to support coordinated reflection and plan revision.

\textbf{\textit{d. Self sorrection.}}
\label{sec:self_correction}
Self-correction refines outputs at inference via feedback or internal evaluation, improving reliability without parameter updates, crucial for dynamic agentic environments.
Approaches include \textit{feedback-based refinement}~\cite{madaan2023self}, where SELF-REFINE iteratively critiques and revises outputs within a single LLM; \textit{confidence-based adjustment}~\cite{self_cor_confidence}, where IoE triggers retries only when confidence is low to avoid over-correction; and \textit{tool-augmented correction}~\cite{self_cor_critic}, where CRITIC integrates calculators, search, or interpreters to drive “verify–correct” cycles.

\subsubsection{Task-Episodic Memory}
\label{sec:task_episodic_memory}
In long-horizon agentic tasks, recalling past actions, successes, and failures is essential to avoid redundancy and inefficiency. \textit{Episodic memory} extends beyond short-lived working memory by storing structured records of actions, observations, and outcomes (\figref{fig:task_episodic_memory}), enabling retrospective inference and experience reuse under sparse feedback.
Approaches fall into two categories: \textit{data replay}, which reuses logged interactions or embeddings to improve sample efficiency and mitigate forgetting, \eg diverse dialogue replay for personalization~\cite{data_replay_1} or bandit-based clustering for dynamic sampling~\cite{data_replay_2}; and \textit{self experiences}, where agents generate and store semantic triplets, logs, or documents for introspection and decision-making, as in RET-LLM~\cite{self_exp_ret_llm}, ChatDB~\cite{self_exp_chatdb}, CorpusBrain~\cite{self_exp_corpusbrain}, and MemoChat~\cite{self_exp_memochat}.

\subsubsection{External Semantic Memory}
\label{sec:external_semantic_memory}
In open-ended environments, pretrained FMs alone cannot ensure reliable inference; agents need access to factual knowledge, task schemas, and up-to-date information to avoid hallucination and improve generalization. \textit{Semantic memory} provides persistent external representations, \eg knowledge graphs, document stores, or vector databases, that complement model parameters and differ from episodic memory by encoding generalized facts and concepts (\figref{fig:external_semantic_memory}).
Two main approaches dominate. \textit{First, continual knowledge graph learning} incrementally expands structured knowledge with new entities and relations, supporting consistent reasoning in dynamic domains, as in domain-specific alignment~\cite{con_know-graph_1}, evidence-graph inference (MindMap)~\cite{con_know-graph_2}, and KG-RAG triple construction for fine-grained reasoning~\cite{con_know-graph_3}. \textit{Second, continual document learning} ingests, summarizes, and indexes new documents for retrieval-based inference, using strategies such as block-skipping (LangChain~\cite{2024langchain}) and incremental document-aware storage (LlamaIndex~\cite{2023LlamaIndex}).

\begin{table*}[t]
\centering
\caption{Summary of interactive learning for real-time FM adaptation on AI agents.} 
\vspace{-2mm}
\tiny
\label{tab:interactive_learning}
\renewcommand{\arraystretch}{1.05}
\setlength{\tabcolsep}{6pt}
\resizebox{\textwidth}{!}{%
\begin{tabular}{|c|c|c|c|c|}
\hline
\multicolumn{2}{|c|}{\textbf{Categories}} & 
\multicolumn{1}{c|}{\textbf{Technique highlight}} & 
\multicolumn{1}{c|}{\textbf{Year}} & 
\multicolumn{1}{c|}{\textbf{Ref}} \\
\hline

\multirow{18}{*}[-0ex]{\centering\textbf{\begin{tabular}{c} Interactive \\learning \\ ~(\S\ref{sec:interactive_learning})\end{tabular}}} 
& \multirow{6}{*}[-0ex]{\centering\textbf{\begin{tabular}{c} Human \\feedback-based\\ learning ~(\S\ref{sec:HFL})\end{tabular}}} 
& \begin{tabular}[c]{@{}c@{}} Simulate user interaction, heuristic feedback, CoT-based reasoning.\end{tabular} & 2024 & \cite{2024simuser} \\
\cline{3-5}
& & \begin{tabular}[c]{@{}c@{}}GPT-4 for fine-grained ratings, human preferences alignment.\end{tabular} & 2023 & \cite{2023ultrafeedback} \\
\cline{3-5}
& & \begin{tabular}[c]{@{}c@{}}LLM verifiers, corrective feedback, refine decision-making. \end{tabular} & 2025 & \cite{2025ExpeL} \\
\cline{3-5}
& & \begin{tabular}[c]{@{}c@{}}LLM inference for reward modeling, StableReinforce algorithm.\end{tabular} & 2025 & \cite{2025r1_reward} \\
\cline{3-5}
& & \begin{tabular}[c]{@{}c@{}}Bradley-Terry preference modeling, optimize policy via binary cross-entropy.\end{tabular} & 2023 & \cite{2023DPO} \\
\cline{3-5}
& & \begin{tabular}[c]{@{}c@{}}Reward model on human rankings, PPO for policy alignment.\end{tabular} & 2022 & \cite{2022InstructGPT} \\
\cline{2-5}

& \multirow{4}{*}[-0ex]{\centering\textbf{\begin{tabular}{c} Imitation \\learning \\ ~(\S\ref{sec:Imitation_Learning})\end{tabular}}} 
& \begin{tabular}[c]{@{}c@{}}Inverse soft Q-learning, occupancy and token likelihood,principled imitation.\end{tabular} & 2024 & \cite{2024Geist} \\
\cline{3-5}
& & \begin{tabular}[c]{@{}c@{}}Joint multimodal tokens, Swin Transformer, PPO for prompt tuning.\end{tabular} & 2024 & \cite{2024Zhang} \\
\cline{3-5}
& & \begin{tabular}[c]{@{}c@{}}Align VLM and LLM expert with DAgger-DPO, distill expert actions and feedback.\end{tabular} & 2024 & \cite{2024emma} \\
\cline{3-5}
& & \begin{tabular}[c]{@{}c@{}}Bootstrap LLM planner with demonstration, iterative self-training, positive feedback.\end{tabular} & 2024 & \cite{2024llm_personalize} \\
\cline{2-5}

& \multirow{3}{*}[-0ex]{\centering\textbf{\begin{tabular}{c} Observational \\learning \\ ~(\S\ref{sec:Observational_Learning})\end{tabular}}} 
& \begin{tabular}[c]{@{}c@{}}Visual to textual prompts, CoT-based reasoning, in-context learning.\end{tabular} & 2024 & \cite{2024velma} \\
\cline{3-5}
& & \begin{tabular}[c]{@{}c@{}}Cycle observation, action, and reflection, extract high-similarity subgraphs,  LLM and KG synergy.\end{tabular} & 2024 & \cite{2024oda} \\
\cline{3-5}
& & \begin{tabular}[c]{@{}c@{}}Pioneer-observer LLMs, alternating roles, shared rewards, policy co-adaptation.\end{tabular} & 2024 & \cite{2024copy} \\
\cline{2-5}

& \multirow{5}{*}[-0ex]{\centering\textbf{\begin{tabular}{c} Reinforcement \\learning \\ ~(\S\ref{sec:RL})\end{tabular}}} 
& \begin{tabular}[c]{@{}c@{}} Self-reflective feedback, layered memories, refine decision-making.\end{tabular} & 2023 & \cite{shinn2023reflexion} \\
\cline{3-5}
& & \begin{tabular}[c]{@{}c@{}}External experience memory with Q-learning,  refine long-term memory via RL.\end{tabular} & 2023 & \cite{2023REMEMBERER} \\
\cline{3-5}
& & \begin{tabular}[c]{@{}c@{}}Monte Carlo Tree Search, self-critique, off-policy preference optimization.\end{tabular} & 2024 & \cite{2024agent_Q} \\
\cline{3-5}
& & \begin{tabular}[c]{@{}c@{}}Manager-Analyst structure, CVRF, use episodic and working memories.\end{tabular} & 2023 & \cite{2024fincon} \\
\cline{3-5}
& & \begin{tabular}[c]{@{}c@{}}Ground LLMs as policies in text-based environments, use PPO for online RL.\end{tabular} & 2023 & \cite{2023glam} \\
\hline
\end{tabular}%
}
\vspace{-5mm}
\end{table*}

\subsection{Interactive Learning}
\label{sec:interactive_learning}
Despite advances in PEFT and memory mechanisms, AI agents on mobile/edge platforms still struggle with dynamic environments, partial observability, sparse feedback, and unpredictable dynamics, where offline training or static policies fall short.
\textit{Interactive learning} addresses this by refining agent behavior \textit{at test time} through continuous interaction with environments, humans, or other agents (\tabref{tab:interactive_learning}). 
By leveraging real-time feedback (\eg user corrections, failure signals, environmental changes), agents adapt efficiently without dense supervision or full retraining, enabling robust generalization in tasks like wearable user modeling, AR interaction, and navigation in novel environments.
Interactive learning methods can be grouped into four categories, \ie \textit{Human feedback-based learning}~\cite{2024simuser,2025r1_reward},
\textit{Imitation learning}~\cite{2024Geist,2024llm_personalize}, 
\textit{Observational learning}~\cite{2024velma,2024oda,2024copy}, and
\textit{Reinforcement learning}~\cite{shinn2023reflexion,2024agent_Q,2023glam}.

\begin{figure*}[t]
    \centering
    \subfloat[Human feedback-based learning.]{
        \includegraphics[height=0.24\textwidth]{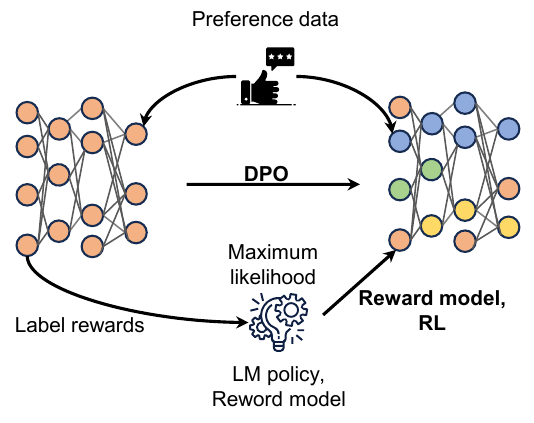}
        \label{fig:HF_learning}
    }
    \subfloat[Imitation and observational learning.]{
        \includegraphics[height=0.24\textwidth]{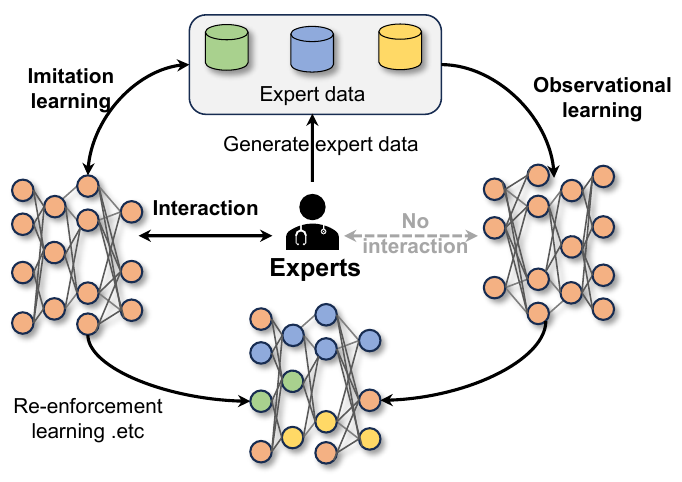}
        \label{fig:IL_Ol}
    }
    \subfloat[Reinforcement learning.]{
        \includegraphics[height=0.24\textwidth]{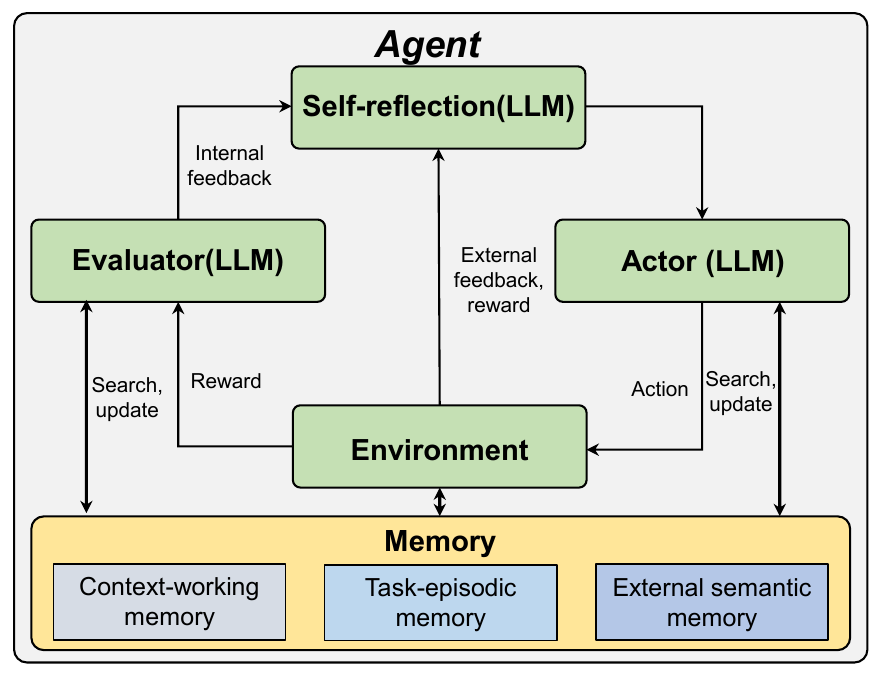}
        \label{fig:RL_LLM_learning}
    }
    \vspace{-2mm}
    \caption{Illustration of interactive learning.}
    \vspace{-6mm}
    \label{fig:memory_types}
\end{figure*}

\subsubsection{Human Feedback-based Learning}
\label{sec:HFL}
Human feedback-based learning refines agent policies using lightweight human input (\eg preferences, ratings, corrections), avoiding explicit reward design or full demonstrations.
It is particularly effective when objectives are ambiguous, demonstrations unavailable, or task demands evolve, enabling flexible test-time adaptation (\figref{fig:HF_learning}).
Feedback typically comes in three forms:
\textit{i) Preference feedback}, where humans select preferred behaviors (\eg SimUser~\cite{2024simuser} generates usability preferences via CoT reasoning).
\textit{ii) Scalar feedback}, where behaviors receive numerical ratings (\eg ULTRAFEEDBACK~\cite{2023ultrafeedback} provides fine-grained GPT-4 ratings for reward models).
\textit{iii) Corrective feedback}, where humans highlight and revise errors (\eg V-Droid~\cite{2025ExpeL} employs LLM verifiers for real-time GUI correction).
Three main training paradigms are used:
\textit{i) Reward modeling}, which learns explicit estimators from feedback (\eg R1-Reward~\cite{2025r1_reward} reformulates multimodal reward modeling with StableReinforce).
\textit{ii) Direct Preference Optimization (DPO)}~\cite{2023DPO}, which bypasses reward models by directly optimizing policies with binary preference probabilities.
\textit{iii) Reinforcement Learning with Human Feedback (RLHF)}, which integrates reward models into RL (\eg InstructGPT~\cite{2022InstructGPT}).

\subsubsection{Imitation Learning}
\label{sec:Imitation_Learning}
In agentic systems, designing explicit reward functions is often infeasible due to sparse feedback and limited resources.
Imitation learning (IL) bypasses this by learning policies directly from expert demonstrations, where state–action trajectories encode task objectives implicitly.
By mimicking expert behaviors, IL reduces reward-engineering overhead and accelerates deployment, making it effective for real-time agents such as drones, vehicles, and wearables (\figref{fig:IL_Ol}).
Recent advances adapt IL to LLM-based and multimodal agents:
Geist \etal~\cite{2024Geist} introduce occupancy-aligned distribution matching for improved LLM adaptation.
Zhang \etal~\cite{2024Zhang} propose a multimodal IL framework integrating visual–LiDAR fusion with reinforcement-guided prompt optimization.
EMMA~\cite{2024emma} aligns a VLM with an LLM expert via a DAgger-DPO algorithm to mitigate compounding errors.
LLM-Personalize~\cite{2024llm_personalize} bootstraps planning with demonstrations and iteratively refines policies through preference-aligned self-training. 

\subsubsection{Observational Learning}
\label{sec:Observational_Learning}
Unlike imitation learning, which directly maps expert state–action pairs, observational learning equips agents with behavioral competence by \textit{interpreting observations} rather than replicating actions.
This paradigm emphasizes building internal models of objectives, dynamics, and causal relations from multimodal inputs (videos, logs, text), enabling scalable self-supervised adaptation without explicit supervision.
It is particularly effective for semantic grounding, long-horizon planning, and cross-modal alignment, where inference—not replication—drives behavior.
Recent work illustrates diverse implementations:
VELMA~\cite{2024velma} verbalizes visual trajectories into textual prompts for in-context action prediction.
ODA~\cite{2024oda} applies an observe–act–reflect cycle with compact subgraphs for multi-hop reasoning and KG–FM synergy.
CORY~\cite{2024copy} coordinates pioneer and observer LLMs with shared rewards, enhancing robustness through co-adaptation.

\subsubsection{Reinforcement Learning}
\label{sec:RL}
In dynamic environments with long-horizon tasks and sparse rewards, static fine-tuning or fixed supervision often fail. 
Reinforcement learning (RL) offers an interaction-driven framework that optimizes policies via cumulative feedback, aligning actions with long-term outcomes without handcrafted rewards or expert demonstrations.
This makes RL well-suited for FM-based agents in open-ended tasks such as semantic navigation, tool use, and multi-step instruction following (\figref{fig:RL_LLM_learning}).
Recent work explores integrating RL with feedback and memory:
Reflexion~\cite{shinn2023reflexion} enables verbal RL via self-reflective feedback and layered memory.
REMEMBERER~\cite{2023REMEMBERER} augments Q-learning with episodic memory for experience reuse.
Agent Q~\cite{2024agent_Q} couples MCTS with preference-guided off-policy optimization for web reasoning.
FINCON~\cite{2024fincon} applies RL to financial decision-making through a Manager–Analyst multi-agent design.
GLAM~\cite{2023glam} grounds LLMs as policies in text environments via online PPO, improving efficiency and generalization.

\begin{table*}[t]
\centering
\caption{Summary of adaptive system scheduling for agentic FM retraining.}
\vspace{-2mm}
\tiny
\label{tab:retrain}
\renewcommand{\arraystretch}{1.05}
\setlength{\tabcolsep}{6pt}
\resizebox{\textwidth}{!}{%
\begin{tabular}{|c|c|c|c|c|c|}
\hline
\multicolumn{3}{|c|}{\textbf{Categories}} & 
\multicolumn{1}{c|}{\textbf{Technique highlight for improving}} & 
\multicolumn{1}{c|}{\textbf{Year}} & 
\multicolumn{1}{c|}{\textbf{Ref}} \\
\hline

\multirow{26}{*}[-0ex]{\textbf{\begin{tabular}{c}Mobile \\ LLM-adaptive \\system \\scheduling \\level\\~(\S\ref{sec:system_scheduling}) \end{tabular}}} 
& \multirow{10}{*}[-0ex]{\textbf{\begin{tabular}{c}Memory and \\ parameter \\management\\~(\S\ref{sec:memory_parameter_management}) \end{tabular}}} 
& \multirow{5}{*}[-0ex]{\textbf{\begin{tabular}{c}Memory \\allocation\\~(\S\ref{sec:memory_parameter_management}) \end{tabular}}} 
& \begin{tabular}[c]{@{}c@{}}Model scheduling, offloading, hierarchical tensor placement.\end{tabular} & 2024 & \cite{2024edge_LLM} \\
\cline{4-6}
& & & \begin{tabular}[c]{@{}c@{}}Token-wise recomputation, bi-level memory planning.\end{tabular} & 2025 & \cite{2025memo} \\
\cline{4-6}
& & & \begin{tabular}[c]{@{}c@{}}Polyhedral dependence graphs, memory reuse, dynamic mamory management.\end{tabular} & 2025 & \cite{2025silvestre} \\
\cline{4-6}
& & & \begin{tabular}[c]{@{}c@{}}Partition model states, offload to NVMe, memory-centric tiling.\end{tabular} & 2021 & \cite{rajbhandari2021zero-Infinity} \\
\cline{4-6}
& & & \begin{tabular}[c]{@{}c@{}}Sliding window eviction, tensor partitioning, in-place recomputation.\end{tabular} & 2023 & \cite{zhang2023coop} \\
\cline{4-6}
\cline{3-6}

& & \multirow{5}{*}[-0ex]{\textbf{\begin{tabular}{c}Memory \\swapping\\~(\S\ref{sec:memory_swapping}) \end{tabular}}} 
& \begin{tabular}[c]{@{}c@{}}Dynamically allocate tensors, optimize offloading, hierarchical placement.\end{tabular} & 2024 & \cite{2024edge_LLM} \\
\cline{4-6}
& & & \begin{tabular}[c]{@{}c@{}}Paged Optimizers, automatic page migration, gradient checkpointing.\end{tabular} & 2023 & \cite{dettmers2023qloraefficientfinetuningquantized} \\
\cline{4-6}
& & & \begin{tabular}[c]{@{}c@{}}Sparse gradient compression, parallel cross-layer swapping.\end{tabular} & 2025 & \cite{2025LSPOffload} \\
\cline{4-6}
\cline{4-6}
& & & \begin{tabular}[c]{@{}c@{}}Offload expert parameters, VM-like prefetching, LRU caching.\end{tabular} & 2024 & \cite{2024ES_MoE} \\
\cline{4-6}
\cline{4-6}
& & & \begin{tabular}[c]{@{}c@{}}Offload gradients and optimizer states, reduce bandwidth usage.\end{tabular} & 2024 & \cite{jang2024smart-Infinity} \\
\cline{4-6}
\cline{2-6}

& \multirow{16}{*}[-0ex]{\textbf{\begin{tabular}{c}Computation \\graph level\\~(\S\ref{sec:computation_graph_level}) \end{tabular}}} 
& \multirow{3}{*}[-0ex]{\textbf{\begin{tabular}{c}Activation \\recomputation\\~(\S\ref{sec:act_recomputation}) \end{tabular}}} 
& \begin{tabular}[c]{@{}c@{}}Recompute $QK^\top$ and softmax, minimize footprint,  tile-wise backward pass.\end{tabular} & 2022 & \cite{2022flashattention} \\
\cline{4-6}
& & & \begin{tabular}[c]{@{}c@{}}Discard intermediate masked weights, retain input activations.\end{tabular} & 2025 & \cite{2025lors} \\
\cline{4-6}
& & & \begin{tabular}[c]{@{}c@{}}Fuse recomputation into single kernel.\end{tabular} & 2023 & \cite{2023flashattention2} \\
\cline{4-6}
\cline{3-6}

& & \multirow{4}{*}[-0ex]{\textbf{\begin{tabular}{c}Activation \\compression\\~(\S\ref{sec:operator_fusion}) \end{tabular}}} 
& \begin{tabular}[c]{@{}c@{}}Project updates into sparse subspaces, enable fine-tuning on GPUs, combine with checkpointing.\end{tabular} & 2025 & \cite{2025LSPOffload} \\
\cline{4-6}
& & & \begin{tabular}[c]{@{}c@{}}Exploit token-level sparsity, skip redundant activations.\end{tabular} & 2025 & \cite{2025lemo} \\
\cline{4-6}
& & & \begin{tabular}[c]{@{}c@{}}Compile-time graph pruning, reorder scheduling, retain essential activations.\end{tabular} & 2023 & \cite{2023pockengine} \\
\cline{4-6}
& & & \begin{tabular}[c]{@{}c@{}}Fuse matrix multiplication, reorder execution.\end{tabular} & 2022 & \cite{2022flashattention} \\
\cline{3-6}

& & \multirow{4}{*}[-0ex]{\textbf{\begin{tabular}{c}Operator \\fusion\\~(\S\ref{sec:operator_fusion}) \end{tabular}}} 
& \begin{tabular}[c]{@{}c@{}}Replace split-K with split-Q, eliminate inter-warp communication.\end{tabular} & 2023 & ~\cite{2023flashattention2} \\
\cline{4-6}
& & & \begin{tabular}[c]{@{}c@{}}Fuse low-rank update and mask, recompute on backward pass.\end{tabular} & 2025 & \cite{2025lors} \\
\cline{4-6}
& & & \begin{tabular}[c]{@{}c@{}}Identify commutative operators, merge operators, reuse context.\end{tabular} & 2024 & \cite{2024data_juicer} \\
\cline{4-6}
& & & \begin{tabular}[c]{@{}c@{}}Fuse LayerNorm and BatchMatMul.\end{tabular} & 2024 & \cite{2023pockengine} \\
\cline{3-6}

& & \multirow{5}{*}[-0ex]{\textbf{\begin{tabular}{c}Operator \\reordering\\~(\S\ref{sec:oprator_reordering}) \end{tabular}}} 
& \begin{tabular}[c]{@{}c@{}}Load $K$/$V$ blocks, load $Q$ sequentially, avoid full attention matrix.\end{tabular} & 2022 & \cite{2022flashattention} \\
\cline{4-6}
& & & \begin{tabular}[c]{@{}c@{}}Apply straight-through estimator, transform backward computation.\end{tabular} & 2025 & \cite{2025lors} \\
\cline{4-6}
& & & \begin{tabular}[c]{@{}c@{}}Polyhedral Dependence Graphs, optimize execution and memory.\end{tabular} & 2025 & \cite{2025silvestre} \\
\cline{4-6}
& & & \begin{tabular}[c]{@{}c@{}}Reorder gradient computation, early tensor release, reduce memory.\end{tabular} & 2023 & \cite{2023pockengine} \\
\cline{4-6}
\cline{4-6}
& & & \begin{tabular}[c]{@{}c@{}}Asynchronous pipelining, break dependencies, improve throughput.\end{tabular} & 2024 & \cite{2024flashattention3} \\
\hline

\end{tabular}%
}
\vspace{-5mm}
\end{table*}

\subsection{Adaptive System Scheduling}
\label{sec:system_scheduling}
Beyond algorithm-level optimization, adaptive system scheduling targets the performance–efficiency trade-off in agentic FM retraining (\figref{fig:test_time_adaptation}). 
It dynamically reallocates memory and compute within the Transformer graph by managing activation tensors and operator paths in real time. By exploiting FM retraining characteristics, it maximizes hardware utilization, enabling scalable, efficient, and on-the-fly test-time adaptation (\tabref{tab:retrain}).

\begin{figure}[t]
    \centering
    \includegraphics[width=0.45\textwidth]{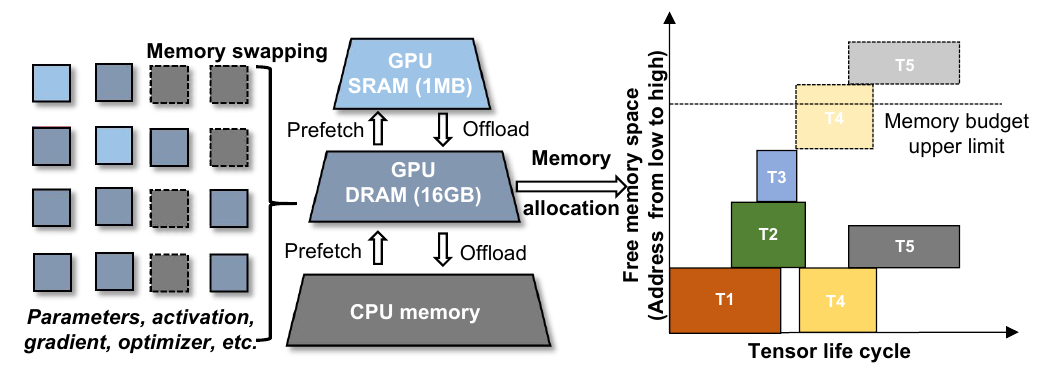}
    \caption{Illustration of memory and parameter management.}
    \vspace{-6mm}
    \label{fig:parameter_menagement}
\end{figure}

\subsubsection{Memory and Parameter Management}
\label{sec:memory_parameter_management}
Memory and parameter management complements algorithm-level PEFT (\secref{sec:peft}) by addressing system bottlenecks in FM adaptation. It emphasizes adaptive \textit{allocation} and \textit{swapping} of activations, gradients, optimizer states, and parameters to reduce fragmentation and peak usage, enabling scalable retraining under constrained hardware (\figref{fig:parameter_menagement}).

\textit{\textbf{a. Memory allocation.}  }
Static allocation fails under dynamic tensor shapes and irregular attention, leading to fragmentation. Recent work improves layout and reuse across layers and tiers: EDGE-LLM~\cite{2024edge_LLM} adaptively places tensors via graph traversal over SRAM–DRAM–SSD; Memo~\cite{2025memo} applies token-wise recomputation with bi-level planning; Silvestre \etal~\cite{2025silvestre} exploit polyhedral dependence graphs for KV-cache reuse under dynamic shapes; and ZeRO-Infinity~\cite{rajbhandari2021zero-Infinity} shards states and offloads activations to CPU/NVMe for trillion-scale fine-tuning.

\textit{\textbf{b. Memory swapping.}}
\label{sec:memory_swapping}
Since memory-heavy components are not always active, swapping adaptively offloads them to CPU/SSD tiers and reloads on demand, balancing compute–memory trade-offs at the cost of I/O. Efficiency is enhanced by scheduling and compression: QLoRA~\cite{dettmers2023qloraefficientfinetuningquantized} introduces paged optimizers with unified memory; Edge-LLM~\cite{2024edge_LLM} searches cost models for tensor offloading; ProTrain overlaps swapping with compute; ES-MoE~\cite{2024ES_MoE} pipelines expert-level caching; LSPOffload~\cite{2025LSPOffload} combines sparse compression with cross-layer bidirectional swapping; Elixir, PatrickStar~\cite{PatrickStar9940581}, and Smart-Infinity~\cite{jang2024smart-Infinity} improve utilization via profiling, dynamic redistribution, and near-storage computing; while ZeRO-Offload jointly offloads data and compute across GPU–CPU–NVMe for scalable training.

\begin{figure}[t]
    \centering
    \subfloat[activation recomputation.]{
        \includegraphics[width=0.24\textwidth]{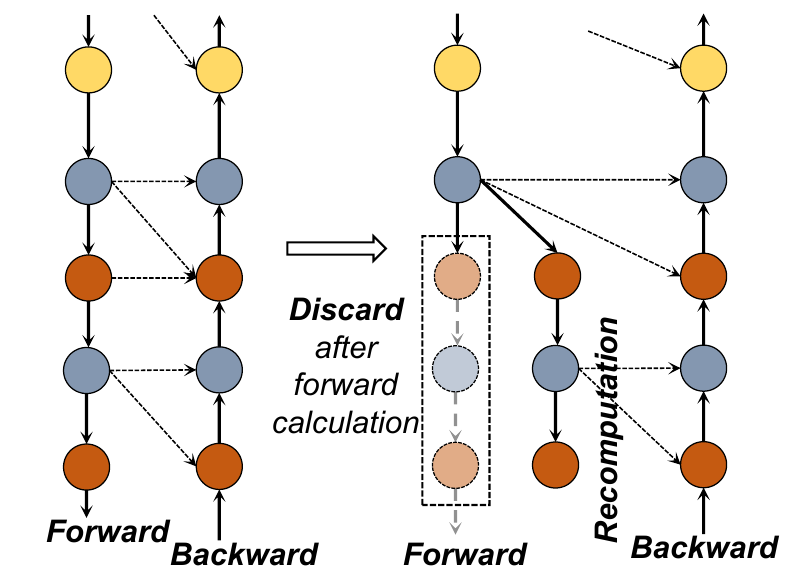}
        \label{fig:activation_recomputation}
    }
    \subfloat[activation compression.]{
        \includegraphics[width=0.20\textwidth]{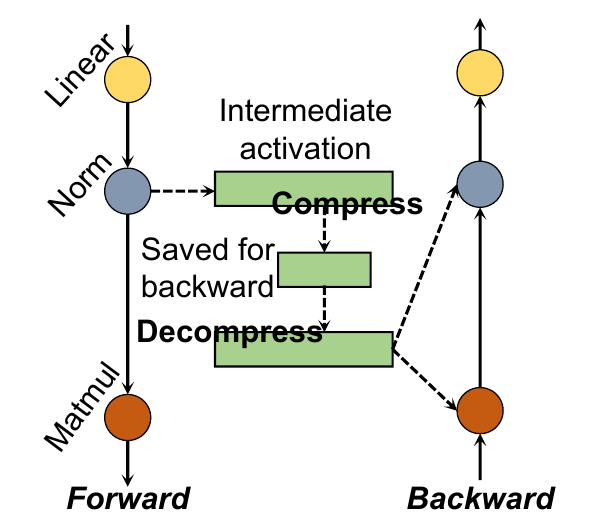}
        \label{fig:activation_compress}
    }
    \vspace{-2mm}
    \caption{Activation recomputation and compression.}
    \vspace{-6mm}
    \label{fig:memory_types}
\end{figure}

\subsubsection{Computation Graph Optimization}
\label{sec:computation_graph_level}
At the computation graph level, adaptive restructuring reduces FM adaptation overhead by pruning redundant states, refining execution order, and improving access patterns.

\textbf{\textit{a. Adaptive activation recomputation.}}
\label{sec:act_recomputation}
Recomputation lowers peak memory by discarding activations in forward and regenerating them in backward, trading compute for memory (\figref{fig:activation_recomputation}).
FlashAttention-1/2/3~\cite{2022flashattention,2023flashattention2,2024flashattention3} progressively integrate fusion and asynchronous pipelining to reduce recomputation overhead, while LoRS~\cite{2025lors} adaptively retains only input activations, recomputing masked weights to cut graph-tracking costs.

\textit{\textbf{b. Adaptive activation compression.}}  
\label{sec:activation_compression}
Compression directly shrinks activation storage via \textit{low-rank projection}, \textit{sparsity}, or \textit{token skipping} (\figref{fig:activation_compress}).
PockEngine~\cite{2023pockengine} prunes and reorders computation graphs at compile time; LSP-Offload~\cite{2025LSPOffload} projects gradients into sparse subspaces to complement checkpointing; LEMO~\cite{2025lemo} exploits token-level sparsity with fused operations, adaptively reducing memory to $1/N$ of baseline.

\begin{figure}[t]
    \centering
    \includegraphics[width=0.4\textwidth]{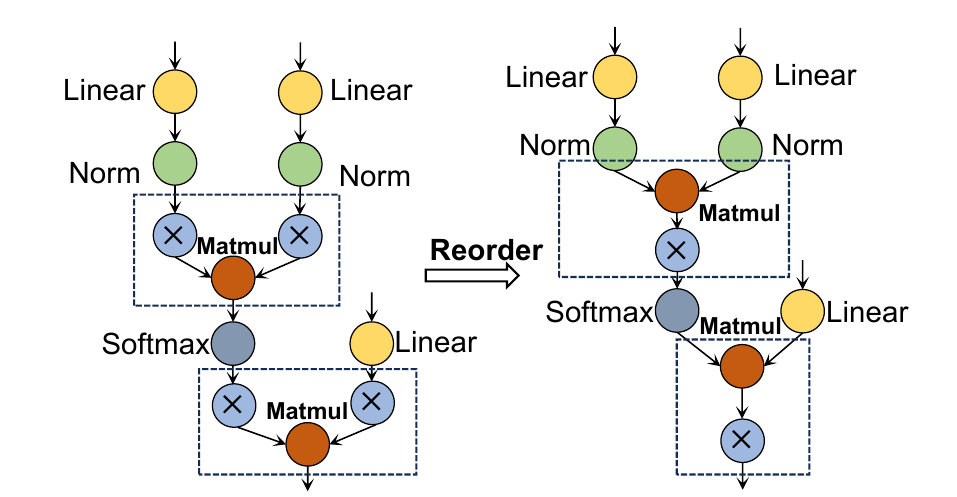}
    \caption{Illustration of operator reordering.}
    \vspace{-5mm}
    \label{fig:operator_reorder}
\end{figure}

\textit{\textbf{c. Adaptive operator fusion.}}  
\label{sec:operator_fusion}  
Operator fusion dynamically merges tightly coupled ops (\eg matmul, normalization, masking) into unified kernels, reducing memory traffic, shortening activation lifetimes, and alleviating I/O overhead in resource-constrained fine-tuning (\figref{fig:operator_reorder}).
FlashAttention~\cite{2022flashattention,2023flashattention2} fuses attention kernels to improve locality;
PockEngine~\cite{2023pockengine} compiles decomposed ops (\eg LayerNorm+MatMul) into single kernels;
Data-Juicer~\cite{2024data_juicer} applies context-aware, reordering-based fusion for pipeline efficiency;
LoRS~\cite{2025lors} fuses low-rank updates with masking into \texttt{$mask_{addmm}$}, adaptively discarding intermediates and recomputing during backprop.

\begin{figure}[t]
    \centering
    \includegraphics[width=0.33\textwidth]{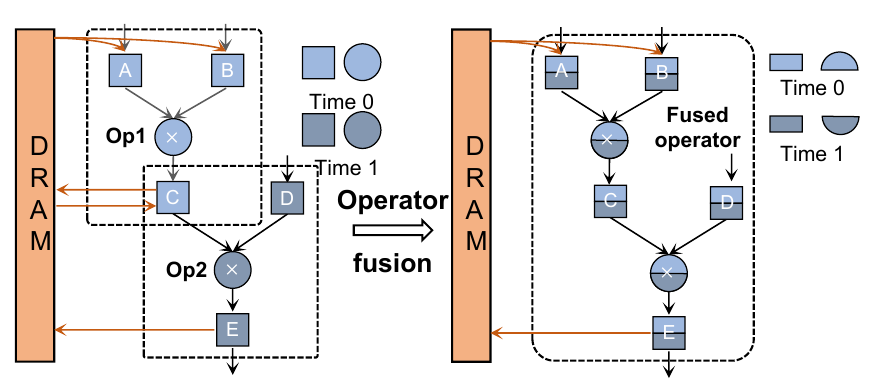}
    \caption{Illustration of operator reordering.}
    \vspace{-7mm}
    \label{fig:operator_fusion}
\end{figure}

\textit{\textbf{d. Adaptive operator reordering.}}  
\label{sec:oprator_reordering}  
Operator reordering dynamically restructures execution order to shorten dependency chains, improve locality, and enable early release of intermediates under memory-constrained fine-tuning (\figref{fig:operator_fusion}).
FlashAttention~\cite{2022flashattention,2023flashattention2,2024flashattention3} exemplifies progressive reordering: preloading $K,V$ to SRAM for on-chip aggregation (v1), deferring scaling with merged logsumexp (v2), and pipelined intra-/inter-warp scheduling (v3).
PockEngine~\cite{2023pockengine} compiles reordered gradient updates for in-place execution;
LoRS~\cite{2025lors} restructures backward passes via mask estimators and decomposed products;
Silvestre \etal~\cite{2025silvestre} use polyhedral dependence graphs to adaptively reorder across temporal/spatial dimensions, improving scheduling efficiency in RLFT workflows.

\begin{table*}[t]
\centering
\caption{Summary of distributed test-time FM retraining at agent networks.}
\vspace{-2mm}
\tiny
\renewcommand{\arraystretch}{1.05}
\label{tab:distributed}
\resizebox{\textwidth}{!}{%
\begin{tabular}{|c|c|c|c|c|c|}
\hline
\multicolumn{3}{|c|}{\textbf{Categories}} & 
\multicolumn{1}{c|}{\textbf{Technique highlight for improving}} & 
\multicolumn{1}{c|}{\textbf{Year}} & 
\multicolumn{1}{c|}{\textbf{Ref}} \\
\hline

\multirow{22}{*}[-0ex]{\textbf{\begin{tabular}{@{}c@{}} Distributed \\ test-time \\ adaptation \\ for LLMs\\~(\S\ref{sec:dis_test_adaptation}) \end{tabular}}} 
& \multirow{11}{*}[-0ex]{\textbf{\begin{tabular}{c} Federated \\ fine-tuning \\ ~(\S\ref{sec:FFT})\end{tabular}}} 
& \multirow{4}{*}[-0ex]{\textbf{\begin{tabular}{c} Communication \\efficiency \\ ~(\S\ref{sec:FFT})\end{tabular}}} 
& \begin{tabular}{c}Select critical layer prompts, skip momentum exchange, optimize dual-side updates.\end{tabular} & 2023 & \cite{2023FedPepTAO} \\
\cline{4-6}
& & & \begin{tabular}{c}Train adapters only, freeze backbone, transmit adapter configurations and parameters.\end{tabular} & 2023 & \cite{2023adafl} \\
\cline{4-6}
& & &  \begin{tabular}{c}Transmit lightweight PEFT modules, apply quantization and compression, unified communication.\end{tabular}& 2024 & \cite{2024FS_LLM} \\
\cline{4-6}
& & &  \begin{tabular}{c}Transmit minimal trainable components, model communication cost.\end{tabular}& 2024 & \cite{2024FedPEFT} \\
\cline{3-6}

& & \multirow{4}{*}[-0ex]{\textbf{\begin{tabular}{c} Memory \\optimization \\ ~(\S\ref{sec:FFT_memory})\end{tabular}}} 
& \begin{tabular}{c}Restrict prompt updates to low-dimensional latent space, avoid backpropagation. \end{tabular}& 2023 & \cite{2023fedbpt} \\
\cline{4-6}
& & &  \begin{tabular}{c}Deploy low-rank adapters, load truncated submatrices, rank self-pruning. \end{tabular}& 2024 & \cite{2024HETLORA} \\
\cline{4-6}
& & &  \begin{tabular}{c}Select trainable weights, configure low-rank adapters.\end{tabular}& 2024 & \cite{2024FedPipe} \\
\cline{4-6}
& & &  \begin{tabular}{c}Freeze backbone, update lightweight adapters and heads, store pretrained weights locally.\end{tabular}& 2023 & \cite{2023kim} \\
\cline{3-6}

& & \multirow{3}{*}[-0ex]{\textbf{\begin{tabular}{c} Computation \\efficiency\\ ~(\S\ref{sec:FFT_computing})\end{tabular}}} 
&  \begin{tabular}{c}Activate shallow adapters, reconfigure adapter structures, cache cross-round activations.\end{tabular}& 2023 & \cite{2023adafl} \\
\cline{4-6}
& & &  \begin{tabular}{c}Optimize prompts via CMA-ES, update prompts with forward inference.\end{tabular}& 2023 & \cite{2023fedbpt} \\
\cline{4-6}
& & &  \begin{tabular}{c}Integrate resource-efficient operators, offload to multi-GPU and CPU, communication compression.\end{tabular}& 2024 & \cite{2024FS_LLM} \\
\cline{2-6}

& \multirow{11}{*}[-0ex]{\textbf{\begin{tabular}{c} Distributed \\fine-tuning\\ ~(\S\ref{sec:DFT})\end{tabular}}} 
& \multirow{3}{*}[-0ex]{\textbf{\begin{tabular}{c} Data parallel \\adaptation \\~(\S\ref{sec:DFT_data_PA})\end{tabular}}} 
&  \begin{tabular}{c}Train LoRA locally, synchronize updates via peer-to-peer.\end{tabular}& 2025 & \cite{2025Dec_lora} \\
\cline{4-6}
& & &  \begin{tabular}{c}Drop transformer layers stochastically, tune PEFT modules, personalize layer sharing.\end{tabular}& 2025 & \cite{2025DropPEFT} \\
\cline{4-6}
& & &  \begin{tabular}{c}Deferred initialization, sharding strategies, Communication overlap optimization.\end{tabular}& 2023 & \cite{zhao2023pytorchFSDP} \\
\cline{3-6}

& & \multirow{5}{*}[-0ex]{\textbf{\begin{tabular}{c} Model parallel \\adaptation \\~(\S\ref{sec:DFT_model_PA})\end{tabular}}} 
&  \begin{tabular}{c}Partition layers, retain trainable components, exchange activations and gradients.\end{tabular}& 2022 & \cite{2022petals} \\
\cline{4-6}
& & &  \begin{tabular}{c}Partition layers, balance memory and throughput, synchronize intra-stage gradients.\end{tabular}& 2024 & \cite{2024pac} \\
\cline{4-6}
& & &  \begin{tabular}{c}Split client/server submodels via weight importance, exchange activations and gradients.\end{tabular}& 2025 & \cite{2025hsplitlora} \\
\cline{4-6}
\cline{4-6}
\cline{4-6}
& & &  \begin{tabular}{c} Hierarchical GPU-CPU workload, demand-priority scheduling.\end{tabular}& 2024 & \cite{2024aptmoe} \\
\cline{4-6}
& & &  \begin{tabular}{c}Adaptive pipelining reduces communication time and improves training throughput.\end{tabular}& 2023 & \cite{shi2023pipemoe} \\
\cline{3-6}

& & \multirow{3}{*}[-0ex]{\textbf{\begin{tabular}{c} Hybrid parallel \\adaptation \\~(\S\ref{sec:DFT_edge_adaptation})\end{tabular}}} 
&  \begin{tabular}{c}Partition model layers, select cut layer, allocate server resources.\end{tabular}& 2025 & \cite{2025li} \\
\cline{4-6}
& & &  \begin{tabular}{c}Offload  computation, retain lightweight adapters, decompose ranks adaptively.\end{tabular}& 2025 & \cite{2025hsplitlora} \\
\cline{4-6}
& & &  \begin{tabular}{c}Delegate forward/backward computation, update adapters, mitigate device constraints.\end{tabular}& 2025 & \cite{2025splitllm} \\
\hline

\end{tabular}%
}
\vspace{-5mm}
\end{table*}

\begin{figure*}[t]
    \centering
  \subfloat[Federated fine-tuning at agent networks.]{
    \includegraphics[height=0.24\textwidth]{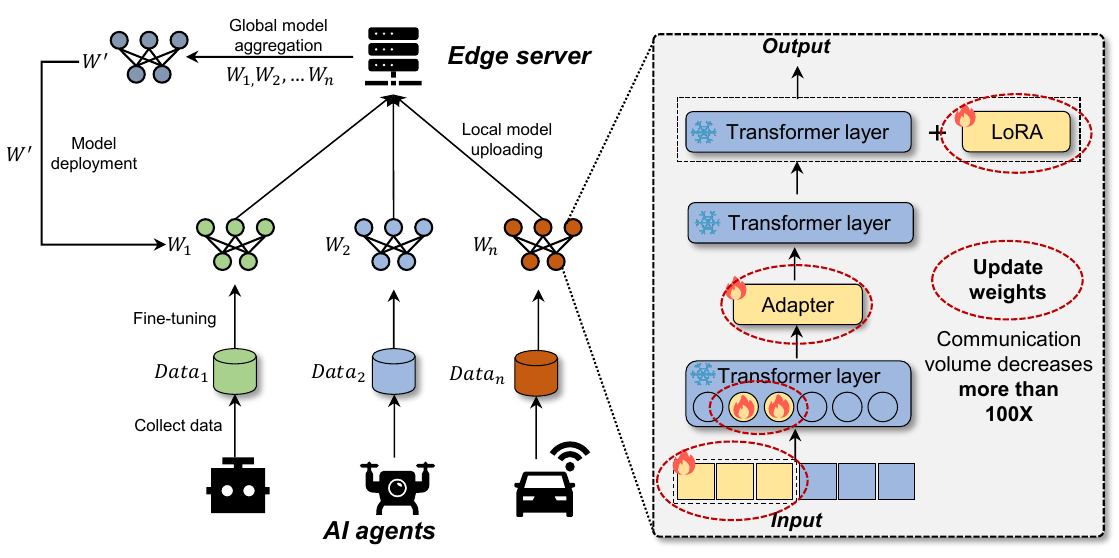}
    \label{fig:reparam}
    }
  \subfloat[Distributed fine-tuning at agent networks.]{
    \includegraphics[height=0.26\textwidth]{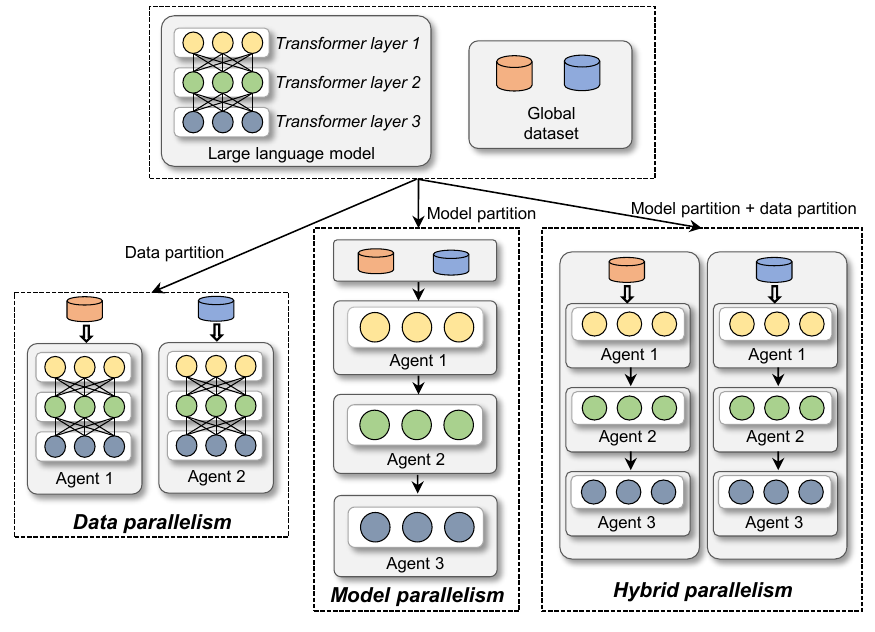}
    \label{fig:reparam}
    }
    \caption{Illustration of distributed test-time retraining of FMs at agent networks.}
    \vspace{-6mm}
    \label{fig:dis_fine_tuning}
\end{figure*}

\subsection{Distributed Test-time FM Retraining at Agent Networks}
\label{sec:dis_test_adaptation}
Federated fine-tuning adapts FMs without centralizing data, but massive parameters, resource constraints, and heterogeneous distributions make full-model training infeasible. Thus, recent work focuses on lightweight, adaptive strategies:

\subsubsection{Federated Fine-tuning}
\label{sec:FFT}
Federated fine-tuning adapts FMs without centralizing data, but massive parameters, resource constraints, and heterogeneous distributions make full-model training infeasible. Thus, recent work focuses on lightweight, adaptive strategies: 

\textit{\textbf{a. Communication efficiency.}}  
Transmit only small trainable modules (\eg adapters, prompts, biases), enhanced with quantization or compression.
FedPepTAO~\cite{2023FedPepTAO} selects critical layers adaptively;
AdaFL~\cite{2023adafl} freezes $>$99\% of weights, cutting transmission by $126\times$;
FS-LLM~\cite{2024FS_LLM} integrates PEFT with quantized streaming for $1000\times$ reduction;
FedPEFT~\cite{2024FedPEFT} formalizes adaptive cost-aware parameter exchange.

\textit{\textbf{b. Memory optimization.}}
\label{sec:FFT_memory}
Reduce local footprint by constraining updates to lightweight or quantized components.
FedBPT~\cite{2023fedbpt} eliminates backprop with latent prompt tuning, lowering client memory $3\times$;
HETLoRA~\cite{2024HETLORA} prunes heterogeneous low-rank modules, updating $<$5\% parameters;
FedPipe~\cite{2024FedPipe} combines selective LoRA ranks with 4/8-bit quantization, shrinking memory $<10\%$ of baseline.
Other adaptive strategies include gradient-based selective layer tuning and adapter-based updates~\cite{2023kim}.

\textit{\textbf{c. Computation efficiency.}}
\label{sec:FFT_computing}
Computation efficiency in federated FM fine-tuning alleviates local training load by \textit{limiting backpropagation}, \textit{dynamically adapting trainable modules}, and \textit{employing lightweight operators}.
AdaFL~\cite{2023adafl} progressively activates shallow adapters with dynamic reconfiguration and activation caching to avoid redundant passes;
FedBPT~\cite{2023fedbpt} eliminates gradients via forward-only black-box prompt tuning;
FS-LLM~\cite{2024FS_LLM} integrates mixed precision, gradient accumulation, and quantized communication, reusing frozen models across clients to reduce compute demand.

\subsubsection{Distributed FM Fine-tuning}
\label{sec:DFT}
Distributed FM fine-tuning extends adaptation across collaborative device networks, enabling scalable training under resource-constrained and heterogeneous environments. By coordinating computation and storage in \textit{peer-to-peer} or hierarchical \textit{mobile–edge–cloud} systems, it overcomes single-device limits. 
Approaches fall into three categories: \textit{data parallel}, \textit{model parallel}, and \textit{hybrid parallel adaptation}.

\textit{\textbf{a. Data parallel adaptation.}} 
\label{sec:DFT_data_PA}
Replicates model states across devices and synchronizes updates, with \textit{dynamic parameter-efficient modules} and dropout-based strategies reducing overhead.
Dec-LoRA~\cite{2025Dec_lora} trains LoRA locally and synchronizes updates peer-to-peer under non-IID data, DropPEFT~\cite{2025DropPEFT} applies stochastic layer dropout with adaptive configuration to cut compute and communication on edge devices.

\textit{\textbf{b. Model parallel adaptation.}}  
\label{sec:DFT_model_PA}  
Partitions FM layers/submodules across devices for pipeline-style computation, balancing memory and compute under edge constraints.
PETALS~\cite{2022petals} distributes Transformer layers while keeping PEFT modules local. Li \etal~\cite{2025li} apply split learning for LoRA, with CARD dynamically selecting cut layers to minimize latency/energy.
Other adaptive schedulers include PipeMoE~\cite{shi2023pipemoe}, which tunes pipeline degree and overlaps comm/compute, and APTMoE~\cite{2024aptmoe}, which allocates workloads across GPUs/CPUs by expert popularity with demand-priority scheduling.

\textit{\textbf{c. Hybrid parallel adaptation.}}
\label{sec:DFT_edge_adaptation}
Combines data, model, and offloading strategies to dynamically optimize memory, compute, and communication across heterogeneous tiers.
PAC~\cite{2024pac} adapts micro-batch scheduling with intra-stage AllReduce;
HSplitLoRA~\cite{2025hsplitlora} partitions backbones server-side while tuning LoRA edge-side with adaptive rank decomposition;
SplitLLM~\cite{2025splitllm} extends to multi-tier edge–cloud, splitting forward/backward paths while updating adapters locally.
\section{Dynamic Multi-modal FMs in Agentic AI Systems}
\label{sec_MLLM}

Building on dynamic inference and test-time adaptation, agentic AI systems on mobile/edge platforms face heightened challenges in \textit{multi-modal settings}. High-resolution vision, continuous speech, and heterogeneous sensor streams intensify redundancy and memory pressure, while complicating cross-modal alignment, consistency, and scalability.
To address this, dynamic multi-modal FMs integrate \textit{architectural-} and \textit{input-level adaptations}, including:
\textit{dynamic attention},
\textit{dynamic routing},
\textit{adaptive cross-modal alignment}, and
\textit{token compression/pruning}.
Together, these dynamic mechanisms balance efficiency and robustness for scalable multi-modal FM deployment under embedded resource constraints.  

\subsection{Dynamic Multi-modal FMs}
\label{Dynamic Multi-modal FMs}
\begin{table*}[t]
\centering
\caption{Summary of Dynamic Multi-modal FMs and Dynamic Cross-modal Alignment for Agentic AI Systems.}
\vspace{-2mm}
\tiny
\label{tab:dynamic_multimodal_alignment}
\renewcommand{\arraystretch}{1.0}
\setlength{\tabcolsep}{6pt}
\resizebox{\textwidth}{!}{%
\begin{tabular}{|c|c|c|c|c|}
\hline
\multicolumn{2}{|c|}{\textbf{Categories}} & 
\multicolumn{1}{c|}{\textbf{Technique highlight}} & 
\multicolumn{1}{c|}{\textbf{Year}} & 
\multicolumn{1}{c|}{\textbf{Ref}} \\
\hline

\multirow{8}{*}[-0ex]{\centering\textbf{\begin{tabular}{c} Dynamic \\multi-modal FMs \\ ~(\S\ref{Dynamic Multi-modal FMs})\end{tabular}}} 

& \multirow{4}{*}[-0ex]{\centering\textbf{\begin{tabular}{c} Dynamic \\attention \\ ~(\S\ref{Dynamic Attention})\end{tabular}}} 
& \begin{tabular}[c]{@{}c@{}}Modality-specific attention, dynamic visual KV cache, periodic token update.\end{tabular} & 2025 & \cite{zhang2025a-vl} \\
\cline{3-5}
& & \begin{tabular}[c]{@{}c@{}}Grid sparse attention, adaptive stride, modality boundary permutation, pre-filling.\end{tabular} & 2025 & \cite{limminference} \\
\cline{3-5}
& & \begin{tabular}[c]{@{}c@{}}Selective token compression, softmax skipping, HilbertCurve permutation, quantized kernels.\end{tabular} & 2025 & \cite{zhangspargeattention} \\
\cline{3-5}
& & \begin{tabular}[c]{@{}c@{}}Adaptive layer/head switching, latency scheduling.\end{tabular} & 2025 & \cite{xu2025learning} \\
\cline{2-5}

& \multirow{4}{*}[-0ex]{\centering\textbf{\begin{tabular}{c} Dynamic \\routing  \\ ~(\S\ref{Dynamic Routing})\end{tabular}}} 
& \begin{tabular}[c]{@{}c@{}}Sparse MoE routing, entropy regularization, priority routing.\end{tabular} & 2022 & \cite{mustafa2022limoe} \\
\cline{3-5}
& & \begin{tabular}[c]{@{}c@{}}Adaptive deformable transformation, instruction-aware gating, sparse adapters.\end{tabular} & 2024 & \cite{shen2024mome} \\
\cline{3-5}
& & \begin{tabular}[c]{@{}c@{}}Expert interaction, perturbation supervision, adaptive reweighting.\end{tabular} & 2025 & \cite{xin2025i2moe} \\
\cline{3-5}
\cline{3-5}
\cline{3-5}
\cline{3-5}
& & \begin{tabular}[c]{@{}c@{}}Sparse LoRA expert routing, domain conflict mitigation, auxiliary loss.\end{tabular} & 2024 & \cite{chen2024llava-mole} \\
\cline{3-5}
\hline

\multirow{4}{*}[-0ex]{\centering\textbf{\begin{tabular}{c} Dynamic \\multi-modal input \\adaptation \\ ~(\S\ref{sec:dynamic_inputs})\end{tabular}}} 

& \multirow{4}{*}[-0ex]{\centering\textbf{\begin{tabular}{c} Dynamic \\cross-modal \\alignment \\ ~(\S\ref{sec:dynamic_align})\end{tabular}}} 
& \begin{tabular}[c]{@{}c@{}}Trainable latent connections, adaptive block selection.\end{tabular} & 2025 & \cite{huang2025mpnp} \\
\cline{3-5}
& & \begin{tabular}[c]{@{}c@{}}Linear adaptors for alignment, meta-response generation, language-level I/O alignment.\end{tabular} & 2024 & \cite{wang2024modaverse} \\
\cline{3-5}
& & \begin{tabular}[c]{@{}c@{}}Cumulative model merging, realign modalities with replay-based connectors.\end{tabular} & 2025 & \cite{zhang2025merge} \\
\cline{3-5}
\cline{3-5}
& & \begin{tabular}[c]{@{}c@{}}Bind diverse modalities through language, contrastive alignment, unified semantic space.\end{tabular} & 2025 & \cite{batur2025sample} \\
\cline{3-5}
\cline{3-5}
\hline
\end{tabular}%
}
\vspace{-3mm}
\end{table*}

Dynamic multi-modal FMs adapt their architectures to optimize computation, memory, and modality alignment in resource-constrained environments (\tabref{tab:dynamic_multimodal_alignment}).  

\subsubsection{Dynamic Attention}
\label{Dynamic Attention}
Dynamic multi-modal FMs adapt architectures to optimize computation, memory, and modality alignment in resource-constrained environments (\tabref{tab:dynamic_multimodal_alignment}). Core strategies include \textit{dynamic attention}~\cite{zhang2025a-vl,xu2025learning}, which generalizes sparse and hierarchical mechanisms to cross-modal settings by adaptively activating tokens, heads, or blocks according to modality-specific patterns and latency budgets, thereby minimizing redundant computation and memory while preserving inference accuracy, and \textit{dynamic routing}, which selectively activates modality experts to balance efficiency–accuracy trade-offs. For example, A-VL~\cite{zhang2025a-vl} reduces cost by dynamically selecting visual KV caches and exploiting the decay of text attention, MMInference~\cite{limminference} applies permutation-based sparse attention with modality-aware stride and phase search to cut pre-fill complexity, SpargeAttention~\cite{zhangspargeattention} combines token compression with sparse block prediction to skip redundant multiplications, and AdaLLaVA~\cite{xu2025learning} employs a probabilistic scheduler to dynamically activate or skip Transformer blocks, heads, and neurons based on input content and resource budgets. 

\subsubsection{Dynamic Routing}
\label{Dynamic Routing}
Dynamic routing in multimodal FMs extends mixture-of-experts beyond unimodal efficiency optimization to address heterogeneous feature distributions, cross-modal alignment, and  resource constraints. 
It dynamically activates modality- or interaction-specific experts to mitigate interference, handle missing modalities, and scale efficiently under limited compute and memory, typically through modality-aware gating, adaptive expert specialization, and sparse activation. 
\textit{First}, PathWeave, FuseMoE, and LIMoE~\cite{mustafa2022limoe} employ \textit{adaptive gating} to align heterogeneous features while avoiding efficiency collapse; 
\textit{Second}, MoME~\cite{shen2024mome}, I$^2$MoE~\cite{xin2025i2moe}, and CL-MoE introduce specialized or interaction-driven experts with \textit{dynamic reweighting} to reduce interference and enhance continual learning; 
\textit{Third}, MoTE, EvoMoE~\cite{jing2025evomoe}, DeepSeek-VL2, LLaVA-MoLE~\cite{chen2024llava-mole}, and Uni-MoE adopt token-level or evolving sparse routing with \textit{adaptive load balancing} to optimize scalability and efficiency in resource-constrained deployments.

\subsection{Dynamic Multi-modal Input Adaptation}
\label{sec:dynamic_inputs}
Dynamic multi-modal input adaptation addresses the core challenge of processing heterogeneous and redundant inputs under resource constraints in embedded and edge-deployed agentic AI systems.  
It must handle high-resolution vision, continuous speech, and dense sensor streams, which impose heavy demands on memory, computation, and energy. 
Efficient adaptation therefore requires input-aware strategies that dynamically allocate resources and sustain cross-modal alignment during real-time inference and retraining.  

\subsubsection{Dynamic Cross-modal Alignment}
\label{sec:dynamic_align}
\begin{table*}[t]
\centering
\caption{Summary of multi-modal token compression for dynamic multi-modal FMs in agentic AI systems.}
\vspace{-2mm}
\tiny
\label{tab:token_compression}
\renewcommand{\arraystretch}{1.05}
\setlength{\tabcolsep}{15pt}
\resizebox{\textwidth}{!}{%
\begin{tabular}{|c|c|c|c|c|}
\hline
\multicolumn{2}{|c|}{\textbf{Categories}} & 
\multicolumn{1}{c|}{\textbf{Technique highlight}} & 
\multicolumn{1}{c|}{\textbf{Year}} & 
\multicolumn{1}{c|}{\textbf{Ref}} \\
\hline

\multirow{9}{*}[-0ex]{\centering\textbf{\begin{tabular}{c} Multi-modal \\token \\compression  \\ ~(\S\ref{token_compress})\end{tabular}}} 

& \multirow{3}{*}[-0ex]{\centering\textbf{\begin{tabular}{c} Token \\pruning  \\ ~(\S\ref{token_compress})\end{tabular}}} 
& \begin{tabular}[c]{@{}c@{}}Prune low-attention tokens, bypass deep layers.\end{tabular} & 2024 & \cite{2024tokenp} \\
\cline{3-5}
\cline{3-5}
& & \begin{tabular}[c]{@{}c@{}}Select salient tokens, prune redundancies, merge keys.\end{tabular} & 2024 & \cite{2024Llava-prumerge} \\
\cline{3-5}
\cline{3-5}
\cline{3-5}
& & \begin{tabular}[c]{@{}c@{}}Optimal transport for pruning, Sinkhorn estimation, prefilling pruning.\end{tabular} & 2025 & \cite{2025topv} \\
\cline{2-5}

& \multirow{3}{*}[-0ex]{\centering\textbf{\begin{tabular}{c} Token \\merging  \\ ~(\S\ref{token_compress})\end{tabular}}} 
& \begin{tabular}[c]{@{}c@{}}Fuse visual tokens, query-based merging.\end{tabular} & 2025 & \cite{2025llava_mini} \\
\cline{3-5}
& & \begin{tabular}[c]{@{}c@{}}Adaptive pooling for vision token compression, semantic abstraction.\end{tabular} & 2024 & \cite{2024deco} \\
\cline{3-5}
\cline{3-5}
& & \begin{tabular}[c]{@{}c@{}}Similarity-driven token merging, length reduction, threshold control.\end{tabular} & 2023 & \cite{2023accelerating} \\
\cline{2-5}

& \multirow{3}{*}[-0ex]{\centering\textbf{\begin{tabular}{c} Adaptive \\sampling  \\ ~(\S\ref{token_compress})\end{tabular}}} 
& \begin{tabular}[c]{@{}c@{}}Vision-language matching, recursive partitioning.\end{tabular} & 2025 & \cite{2025adaptiveSampling} \\
\cline{3-5}
& & \begin{tabular}[c]{@{}c@{}}AnyRes encoding, bilinear interpolation, adaptive resampling.\end{tabular} & 2024 & \cite{li2024llava} \\
\cline{3-5}
& & \begin{tabular}[c]{@{}c@{}}Block-wise streaming, TMRoPE, dynamic resampling.\end{tabular} & 2025 & \cite{bai2025qwen2} \\
\hline
\end{tabular}%
}
\vspace{-6mm}
\end{table*}

Dynamic cross-modal alignment~\cite{huang2025mpnp,wang2024modaverse,zhang2025merge} enables multimodal FMs on edge and mobile devices to adaptively include or exclude modalities based on system availability, while supporting intra-modality domain adaptation to ensure robustness under heterogeneous and resource-constrained sensing. 
Approaches span three directions: 
\textit{First}, for \textit{new modality adaptation}, MPnP~\cite{huang2025mpnp} introduces nonlinear key–value aligners with latent connections to dynamically control injection depth, and ModaVerse~\cite{wang2024modaverse} aligns language I/O via single-stage tuning to reduce projection–instruction mismatch; 
\textit{Second}, for \textit{incremental adaptation and forgetting mitigation}, MERA~\cite{zhang2025merge} combines cumulative average merging with selective replay, MoInCL~\cite{pian2024MoInCL} leverages instruction-guided pseudo-targets, BABEL applies sequential binary alignments with gradient-norm weighting, and SEMI~\cite{batur2025sample} generates LoRA adapters via a hypernetwork to enhance cross-modal consistency; 
\textit{Third}, for \textit{dynamic fusion}, PathWeave adopts an adapter-in-adapter MoE framework for adaptive modality coordination, while FuseMoE employs instance-level gating to dynamically fuse features, mitigating redundancy and semantic drift.

\subsubsection{Multi-modal Token Compression}
\label{token_compress}
Multi-modal token compression addresses the quadratic cost of self-attention in handling long sequences from high-resolution vision, continuous speech, and heterogeneous sensors, which is prohibitive for mobile and edge deployment. 
Recent approaches emphasize dynamic and adaptive compression by pruning, merging, or sampling tokens to jointly balance efficiency and accuracy. 
\textit{First}, \textit{token pruning} methods such as FastV~\cite{2024tokenp}, VTW~\cite{2025VTW}, and DivPrune dynamically discard redundant or low-importance tokens based on attention statistics, divergence, or diversity maximization, while TopV~\cite{2025topv} applies transport-based selection for latency reduction. 
\textit{Second, token merging} techniques including A-ToMe~\cite{2023accelerating}, LOOK-M, and LLaVA-Mini~\cite{2025llava_mini} compact semantically similar tokens across modalities into fewer representations, reducing sequence length while preserving contextual coherence. 
\textit{Third, adaptive sampling} strategies such as AKS~\cite{2025adaptiveSampling}, LongVU~\cite{shen2024longvu}, and Qwen2.5-VL~\cite{bai2025qwen2} further enhance efficiency by dynamically selecting task-relevant video frames, image patches, or spatiotemporal regions under varying resource budgets, while DeepSeek-VL2 combines global–local tiling for fine-grained yet efficient visual understanding. 
\section{Agentic AI Applications}
\label{Agentic AI Applications}
Agentic AI systems are rapidly extending into diverse domains, imposing various adaptation and efficiency challenges. 

\subsection{Representative Applications}
We first present four representative applications, illustrating the diversity of interaction mechanisms, from physical manipulation to digital automation and creative generation, highlighting adaptive and resource-efficient demands.

\textit{Embodied Agents}.
Embodied agents, including robots~\cite{duan2022survey,2024mengweisurvey}, drones~\cite{baytas2019design}, grippers~\cite{liu2020bioinspired}, and wearables~\cite{adilkhanov2022haptic}, perceive and act in physical environments under strict latency, energy, and compute constraints. Adaptation is critical for multimodal perception, embodied interaction, and sim-to-real transfer. PaLM-SSE~\cite{driess2023palm} integrates continuous sensor modalities into an LLM for robotic planning and manipulation, HuggingGPT~\cite{shen2023hugginggpt} orchestrates perception and action through specialized models, and ReCA~\cite{wan2025reca} co-designs algorithms, systems, and hardware for real-time multi-agent collaboration. 

\textit{GUI Agents}.
GUI agents perceive, reason about, and act upon graphical user interfaces, combining multimodal inputs (screenshots, text, speech, contextual cues) with actions via clicks, taps, or API calls. Their applications span workflow automation~\cite{li2024chatcite}, accessibility~\cite{chheang2024towards}, web/mobile navigation~\cite{wang2024mobile}, and human–computer interaction~\cite{kuuru2022embodied}. 
Adaptive fusion and reasoning are central to recent advances: MP-GUI~\cite{wang2025mp} employs modality-specific perceivers with adaptive fusion for robust GUI understanding; GUI-World~\cite{chen2024gui} benchmarks dynamic GUI interactions and shows persistent challenges for LLMs; InfiGUI-R1~\cite{liu2025infigui} transitions from reactive to reasoning-based agents with spatial distillation and RL for error recovery; and Mirage-1~\cite{xie2025mirage} develops hierarchical multimodal skills with adaptive Monte Carlo Tree Search for long-horizon GUI tasks.

\textit{Generative Agents}.
Generative agents produce new content across modalities, \eg text, image, audio, code, and media, while operating under latency and resource constraints. 
They power diverse applications including summarization~\cite{zhang2024comprehensive}, dialogue~\cite{ni2023recent}, translation~\cite{wu2024transagents}, creative tools~\cite{yao2025application}, code generation~\cite{jiang2024survey}, and multimodal synthesis~\cite{shivappa2023audiobook}. 
Beyond content creation, generative agents also model human behavior.
Generative simulations of 1,000 People~\cite{park2024generative} replicate personality traits and survey responses with 85\% accuracy, while AgentSociety~\cite{piao2025agentsociety} scales to 10,000 agents and 5M+ interactions to study social phenomena like polarization. 
Distributed deployments~\cite{adornetto2025generative} emphasize modularity, energy efficiency, and privacy in heterogeneous edge environments. 
To sustain output quality (coherence, factuality, creativity) under resource limits, these agents must dynamically adjust complexity, caching, or approximations (\eg quantization, lightweight modules). 
Yet continual adaptation in generative settings, especially for repeated user/environment interactions, remains underexplored.

\textit{Personal Assistive Agents}.
Personal assistive agents operate on mobile/edge devices to support health monitoring~\cite{yin2024wearable}, accessibility~\cite{li2024personal}, and productivity~\cite{he2025plan}. 
Unlike cloud-based assistants, they rely on local data and dynamically adapt to user-specific contexts, shifting environments, and resource variability under tight privacy and power constraints. 
Advances such as Pluto/Charon (PAC)~\cite{2024pac} enable lightweight on-device personalization, Privacyasst~\cite{zhang2024privacyasst} sanitize prompts to mitigate data leakage. 
They highlight the dual challenge of achieving adaptive personalization with scarce training data and sustaining resource-aware operation via quantization, pruning, and elastic inference. 

\begin{figure}[t]
    \centering
  \subfloat[Mobile personal agent.]{
    \includegraphics[height=0.13\textwidth]{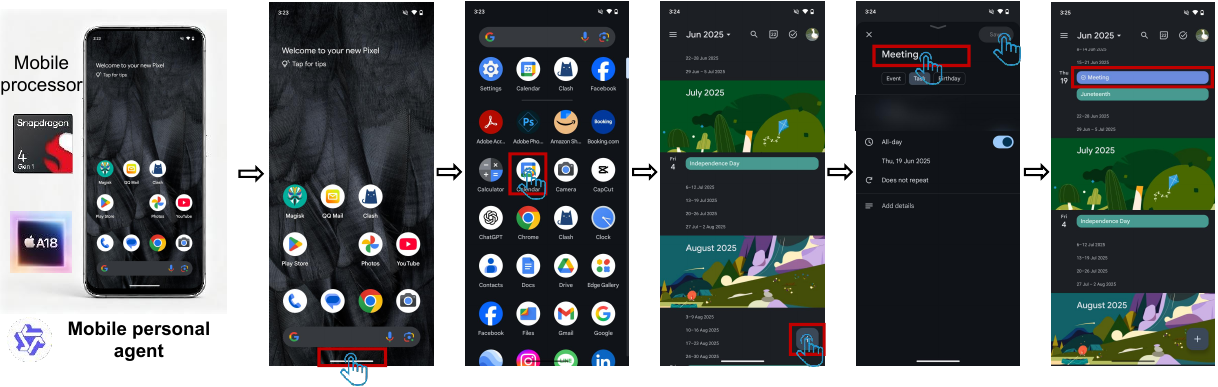}
    \label{fig:case_1}
    }
  \\
  \subfloat[Autonomous driving agent.]{
    \includegraphics[height=0.15\textwidth]{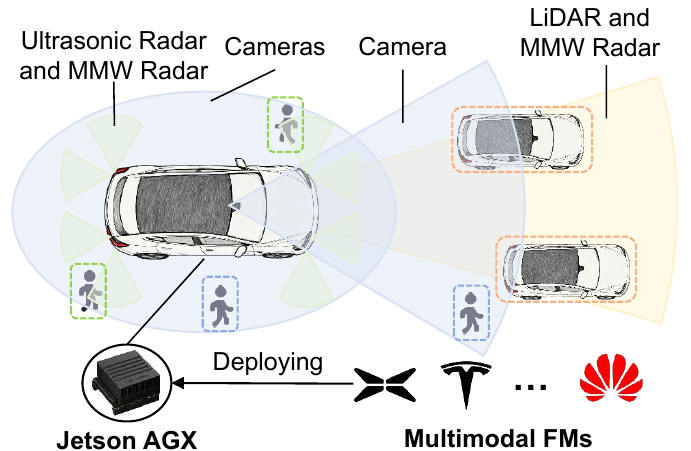}
    \label{fig:case_2}
    }
  \subfloat[Smart home assistant.]{
    \includegraphics[height=0.15\textwidth]{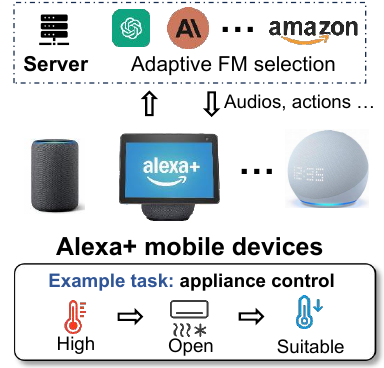}
    \label{fig:case_3}
    }
    \\
  \subfloat[Embodied agent~\cite{aloha2}.]{
    \includegraphics[height=0.15\textwidth]{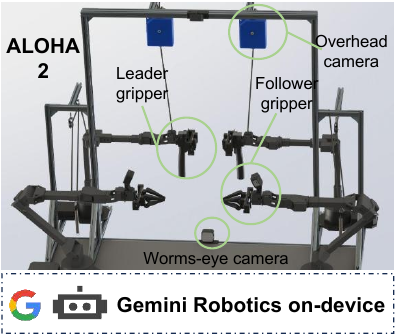}
    \label{fig:case_5}
    }
  \subfloat[Generative agent.]{
    \includegraphics[height=0.15\textwidth]{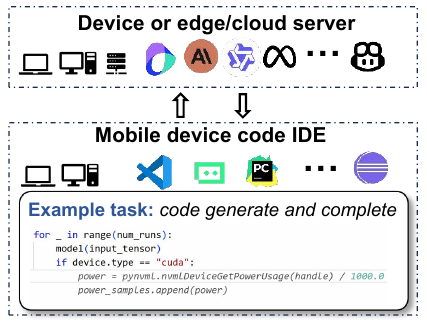}
    \label{fig:case_4}
    }
    \caption{Illustration of case study implementation.}
    \vspace{-6mm}
    \label{fig:case_imp}
\end{figure}

\subsection{Case Study Implementation}
To illustrate how agentic AI operates in practice, we show representative deployments across mobile, automotive, smart home, robotics, and software development (\figref{fig:case_imp}). 

\textit{Mobile Personal Agent}.
Alibaba’s GUI-Owl family exemplifies a cross-device personal assistant. The lightweight GUI-Owl-7B~\cite{ye2025mobile} runs locally on smartphones and PCs for shopping or scheduling, while the larger GUI-Owl-32B is deployed on home-edge hubs for complex reasoning. Built on Mobile-Agent-v3, it integrates \textit{manager}, \textit{worker}, \textit{reflector}, and \textit{notetaker} agents, enabling adaptive task coordination, real-time correction, and cross-device memory.

\textit{Autonomous Driving Agent}.
XPeng Motors~\cite{cvpr2025-wad} compresses a 72B-parameter World FM via RL and distillation to run on in-vehicle NVIDIA Orin-X units. With XNet for perception, XPlanner for trajectory prediction, and XBrain for reasoning, the system dynamically balances accuracy–latency trade-offs to support safety-critical functions such as adaptive braking and lane switching.

\textit{Smart Home Voice Assistant}.
Amazon’s Alexa+~\cite{amazon-alexa-plus} combines on-device wake-word detection with edge-hosted Nova FM and external LLMs (\eg Claude). This hybrid design dynamically allocates inference across devices, ensuring low-latency responses while scaling to complex orchestration tasks like API-based appliance control and personalized dialogue.

\textit{Embodied Agent}.
DeepMind’s Gemini Robotics~\cite{team2025gemini_robotics} runs multimodal reasoning on edge servers while deploying lightweight decoders on robots (\eg ALOHA 2~\cite{aloha2}). By adaptively fusing 3D perception, planning, and safety checks with low-latency motor control, it enables real-world dexterous manipulation from assembly to healthcare delivery.

\textit{Generative Agent}.
ByteDance’s Trae integrates Doubao-1.5-Pro~\cite{doubao2023} and DeepSeek-R1~\cite{2025deepseek-r1} into IDE-native workflows. Through SOLO and Chat modes, plus MCP-based automation, it adapts code generation, debugging, and pull request analysis to developer needs. The 6A workflow (Align–Architect–Atomize–Approve–Automate–Assess) improves individual productivity by 40\%, demonstrating adaptive generative agents for enterprise-scale collaboration.

\subsection{Models and Datasets}
Recent FMs have been increasingly explored for agentic deployment, with some explicitly designed for mobile/edge efficiency and others requiring post-hoc compression and dynamic adaptation. 
Representative efforts include LLaMA~\cite{touvron2023llama}, OPT~\cite{zhang2022opt}, and SAM~\cite{kirillov2023sam} from \textit{Meta}; PaLM~\cite{chowdhery2022palm}, Gemini~\cite{gemini2023}, and T5~\cite{raffel2020t5} from \textit{Google}; Phi~\cite{li2023phi}, Orca~\cite{mukherjee2023orca}, and Kosmos~\cite{kosmos2023} from \textit{Microsoft}; Doubao~\cite{doubao2023} from \textit{ByteDance}; and GPT~\cite{brown2020language,achiam2023gpt} and DALL·E~\cite{ramesh2021dalle} from \textit{OpenAI}. 
While FMs benefit from massive pretraining corpora (\eg GPT-4 with 13T tokens), domain-specific agentic applications remain constrained by fragmented data ecosystems, demanding adaptive and resource-efficient retraining strategies.

Evaluation datasets span multiple modalities. For \textit{language}, BIG-Bench~\cite{srivastava2022bigbench} and HELM~\cite{liang2022helm} benchmark reasoning, planning, and dialogue. For \textit{vision}, Open X-Embodiment~\cite{rtx2023openx} and ScanNet++~\cite{dai2023scannetpp} support robotic perception and manipulation. For \textit{multimodality}, ScienceQA~\cite{lu2022scienceqa}, MMBench~\cite{liu2023mmbench}, and SEED-Bench~\cite{li2023seed} test cross-modal reasoning and grounded interaction. For \textit{embodied tasks}, BEHAVIOR-1K~\cite{li2023behavior}, MineDojo~\cite{fan2022minedojo}, and CALVIN~\cite{mees2022calvin} capture long-horizon perception–action loops.

Despite this progress, current benchmarks insufficiently reflect the dynamic, resource-constrained conditions of mobile/edge deployment. Developing adaptive, domain-specific models and datasets that capture environmental variability, multimodal asynchrony, and real-time feedback remains a critical challenge for evaluating agentic AI in practice.

\subsection {Inference Engines}
To support the performance–portability demands of agentic AI, inference engines integrate resource-efficient optimizations to adapt FMs across heterogeneous hardware, from edge servers to highly constrained mobile/embedded devices~\cite{llama_cpp_github,kwon2023efficient,coreml2023,tflite2023,mnn2023,ncnn2023,mindsporelite2023}.
\textit{llama.cpp}~\cite{llama_cpp_github} demonstrates on-device deployment via quantization, SIMD kernels, heterogeneous backends, and memory mapping, while \textit{vLLM}~\cite{kwon2023efficient} targets edge/server environments with PagedAttention for KV-cache management and continuous batching, trading portability for large-scale throughput.
Cross-platform frameworks such as TensorRT-LLM and OpenVINO~\cite{openvino2023} accelerate inference on GPUs/CPUs, whereas Core ML~\cite{coreml2023} and TensorFlow Lite~\cite{tflite2023} specialize in mobile/wearables with quantization, pruning, and hardware-specific kernels. Lightweight engines like MNN~\cite{mnn2023}, NCNN~\cite{ncnn2023}, and MindSpore Lite~\cite{mindsporelite2023} focus on IoT and Android devices.

\section{Open Issues}
\label{sec_open issue}
This section outlines key challenges and potential directions.

\subsection{Elastic Inference for Perception–Action Loop} 
A central challenge for agentic AI lies in reconciling static foundation models (FMs) with the dynamic perception–action loops of real-world environments. Current FMs excel at symbol processing over text, vision, or multimodal data, yet lack elastic inference that adapts to real-time sensing, long-horizon control, and fluctuating hardware resources, leaving them misaligned with latency, efficiency, and adaptability demands in embedded or embodied deployments.
Recent explorations, such as VLA models for end-to-end perception–action coupling~\cite{team2025gemini_robotics,cvpr2025-wad}, hierarchical reasoning for modular control, and world models for causal dynamics~\cite{gao2024vista}, show promise but remain immature, failing to close the gap between massive FMs and asynchronous, resource-constrained settings.
Key \textit{gaps} include the absence of elastic architectural adaptation, where perception, reasoning, memory, and action submodules scale independently across heterogeneous devices, and the lack of joint resource–model co-optimization that aligns hardware scheduling with submodule reconfiguration. Future solutions may involve meta-controllers that dynamically assemble task- and domain-specific FMs, and modular operators that support composable, elastic scaling and distributed offloading across agents and edge nodes.

\subsection{Lightweight yet Generalizable Physical Intelligence}
Despite advances in text and multimodal reasoning, current FMs remain misaligned with the requirements of physical intelligence, where agents must sustain closed-loop perception–planning–control under real-world uncertainty~\cite{doubao2023,brown2020language}. Lightweight models lack the capacity to bridge abstract reasoning with fine-grained action, handle noisy multimodal feedback, or maintain long-horizon causal dependencies, while large embodied FMs (\eg 40$\sim$140B multimodal models) still suffer from scarce robotic data and persistent sim-to-real gaps. Pretrain–finetune pipelines further struggle in sequential, causal operations, limiting generalization.
The central challenge is delivering models that are both lightweight and generalizable: capable of fusing heterogeneous sensor streams, embedding contextual memory for adaptive planning, and producing executable action sequences under strict latency, memory, and energy budgets. 
Progress will require advances in multimodal fusion, domain generalization, and resource-aware adaptation, making lightweight yet generalizable physical intelligence a key frontier for agentic AI.

\subsection{Responsive Online FM Adaptation}
Agentic AI systems on autonomous platforms (\eg drones, underwater robots, quadrupeds) often lack stable cloud connectivity, requiring foundation models to adapt online under strict memory, compute, and energy limits. 
The challenge is sustaining performance when full retraining is infeasible, data streams are unlabeled, and source data is inaccessible, creating risks of error propagation and catastrophic forgetting. Balancing accuracy and efficiency hinges on selecting which data to use and which modules or neurons to update.
Promising directions include PEFT, prompt learning, interactive learning, and memory-augmented adaptation to shrink trainable parameters, combined with memory management, execution scheduling, and distributed retraining to align updates with hardware constraints. Yet integrating these techniques with label-free streaming, limited backpropagation, and missing source data remains unresolved. Moreover, current autoregressive architectures, rooted in probabilistic generation, are brittle for long-horizon reasoning, compounding errors over time. 
Addressing these gaps calls for uncertainty-aware data selection, real-time relearning, and joint algorithm–system co-design that couples adaptive parameter updates with device-aware scheduling. 

\subsection{Real-time Distributed Multi-modal Sensing}
Mobile and edge agents with heterogeneous sensors (\eg cameras, LiDAR, RF) enable multimodal FMs for autonomous driving and smart environments, yet real-time sensing is fundamentally constrained by \textit{asynchrony}, \textit{redundancy}, and \textit{communication overhead}. Sensor streams differ in rate and latency (\eg camera \textit{vs.} LiDAR under a 100 Mbps link show a 1:4 imbalance, introducing 40 ms delays). Waiting for slow modalities increases latency, while discarding them reduces accuracy, creating an inherent latency–accuracy–communication trade-off that current synchronous, full-modality MLLMs cannot resolve~\cite{bai2025qwen2,wu2024deepseek2vl}.
Emerging paradigms such as predict–verify~\cite{chen2024cascade} and early-exit~\cite{2022adainfer,schuster2022calm,2023FREE} partly mitigate cost, but lack principled ways to capture \textit{modality affinity}, a unified measure of cross-modal consistency, complementarity, and communication relevance across distributed agents and dynamic inputs. Key challenges remain: \textit{(i)} building generalizable models to evaluate and exploit modality affinity under asynchrony and bandwidth limits, \textit{(ii)} designing \textit{non-blocking imputation} and communication-efficient fusion for missing or delayed modalities, and \textit{(iii)} enabling adaptive predict–verify loops with rollback to balance timeliness, accuracy, and communication cost. Addressing these challenges demands holistic co-design of \textit{FM architectures}, \textit{runtime scheduling}, and \textit{adaptive network protocols} to jointly optimize sensing, computation, and communication for robust real-world deployment.

\subsection{Efficient Multi-agent Collaboration}
As FMs grow and edge devices remain resource-limited, multi-agent collaboration is increasingly necessary to overcome single-agent bottlenecks. Yet most existing methods rely on static compile-time partitioning, where model splitting dictates task dispatch. This rigid coupling forces frequent re-compilation under dynamic conditions, incurring high latency and overhead. 
The problem is worsened by Transformer-based FMs, whose intermediate features often exceed raw input size~\cite{2024mengweisurvey}, creating prohibitive transmission costs that hinder distributed deployment.
The core challenge is enabling \textit{elastic, runtime collaboration} that dynamically balances model partitioning and communication. This requires \textit{i)} rapid re-partitioning and adaptive scheduling across heterogeneous devices, \textit{ii)} reducing transmission overhead from large intermediate activations, and \textit{iii)} communication-efficient protocols that sustain responsiveness under bandwidth, latency, and energy limits. 
While recent advances in swarm robotics, federated perception~\cite{chiu2025v2v}, and structured coordination frameworks~\cite{2025multi-agent} highlight potential, achieving scalable, communication-aware, and resource-efficient collaboration in dynamic environments remains open issues.

\subsection{Interactive and Collaborative Human–AI Systems}
Human–AI collaboration requires agents that can perceive intent, interpret context, and generate adaptive responses through natural, multimodal interaction. 
The central challenge lies in sustaining \textit{real-time, resource-aware adaptation} on mobile and edge platforms while integrating continuous feedback from users and environments. Current FMs, which rely mainly on parameter updates, remain limited for dynamic interaction. 
Open problems include \textit{i)} enabling cognition updates through memory, external knowledge, and prompt-driven control; \textit{ii)} unifying perception–language–action pipelines under resource constraints; and \textit{iii)} balancing responsiveness, personalization, and energy efficiency. 
Humans must be treated not only as data annotators but also as interactive partners, providing feedback during inference to refine reasoning, guide actions, and establish trust in collaborative systems.

\section{Conclusion}
\label{sec_conclusuion}
Agentic AI marks a paradigm shift, embedding foundation models into mobile, embedded, and edge systems where adaptivity and resource efficiency are necessities rather than optimizations. 
Unlike cloud settings, real-world deployments face diverse tasks, dynamic environments, and constrained hardware, demanding elastic inference and online adaptation.
This survey clarifies key concepts, presents a taxonomy of enabling techniques, and highlights applications spanning embodied, GUI, generative, and personal agents. We also emphasize distributed coordination, where communication and scheduling shape multi-agent efficiency.
Looking forward, progress will depend on unified benchmarks, algorithm–system–hardware co-design, trustworthy adaptation pipelines, and robust low-latency communication. We aim to motivate future work toward scalable, reliable, and personalized agentic AI bridging powerful FMs with resource-limited real-world environments.

\section*{Acknowledgments}
This work was partially supported by the National Science Fund for Distinguished Young Scholars (No. 62025205) and the National Natural Science Foundation of China (No. 62522215, 62532009, 6247074224).

\bibliography{IEEEart}
\bibliographystyle{IEEEtran}

\end{document}